\documentclass{article}

\usepackage[preprint]{neurips_2025}

\usepackage[utf8]{inputenc} 
\usepackage[T1]{fontenc}    
\usepackage{hyperref}       
\usepackage{url}            
\usepackage{booktabs}       
\usepackage{amsfonts}       
\usepackage{nicefrac}       
\usepackage{microtype}      
\usepackage{xcolor}         

\usepackage{xspace}
\newcommand{\benchtitle}{$\mathcal{E}motion\mathcal{H}allucer$\xspace}
\usepackage{epigraph}
\usepackage{amssymb}
\usepackage{multirow}
\usepackage{amsmath}
\usepackage{bbding}
\definecolor{customBlue}{HTML}{2952FF}
\definecolor{customPurple}{HTML}{991456}
\definecolor{customGreen}{HTML}{5F6F52}

\definecolor{customFirst}{HTML}{09B2D9}
\definecolor{customSecond}{HTML}{FF3B30}

\usepackage{colortbl} 
\usepackage{graphicx} 
\usepackage{subcaption}
\usepackage{wrapfig}

\usepackage[most]{tcolorbox}
\usepackage{lipsum}
\usepackage{longtable}

\title{\benchtitle: Evaluating Emotion Hallucinations in Multimodal Large Language Models}

\author{
\textbf{Bohao Xing}$^{1}$ \quad
\textbf{Xin Liu}$^{1}$\thanks{Corresponding author (xin.liu@lut.fi).} \quad
\textbf{Guoying Zhao}$^{2}$ \quad
\textbf{Chengyu Liu}$^{3}$ \quad
\\
\textbf{Xiaolan Fu}$^{4}$ \quad
\textbf{Heikki Kälviäinen}$^{1,5}$ \\
$^{1}$Lappeenranta-Lahti University of Technology LUT \quad $^{2}$University of Oulu \\ 
$^{3}$Southeast University \quad $^{4}$Shanghai Jiao Tong University \quad $^{5}$Brno University of Technology \\
}

\begin{document}

\maketitle

\begin{abstract}
Emotion understanding is a critical yet challenging task. 
Recent advances in Multimodal Large Language Models (MLLMs) have significantly enhanced their capabilities in this area. 
However, MLLMs often suffer from ``hallucinations'', generating irrelevant or nonsensical content.
To the best of our knowledge, despite the importance of this issue, there has been no dedicated effort to evaluate emotion-related hallucinations in MLLMs.
In this work, we introduce \textbf{EmotionHallucer}, the first benchmark for detecting and analyzing emotion hallucinations in MLLMs. 
Unlike humans, whose emotion understanding stems from the interplay of biology and social learning, MLLMs rely solely on data-driven learning and lack innate emotional instincts. Fortunately, emotion psychology provides a solid foundation of knowledge about human emotions.
Building on this, we assess emotion hallucinations from two dimensions: emotion psychology knowledge and real-world multimodal perception. 
To support robust evaluation, we utilize an adversarial binary question–answer (QA) framework, which employs carefully crafted basic and hallucinated pairs to assess the emotion hallucination tendencies of MLLMs.
By evaluating 38 LLMs and MLLMs on EmotionHallucer, we reveal that:
i) most current models exhibit substantial issues with emotion hallucinations;
ii) closed-source models outperform open-source ones in detecting emotion hallucinations, and reasoning capability provides additional advantages;
iii) existing models perform better in emotion psychology knowledge than in multimodal emotion perception.
As a byproduct, these findings inspire us to propose the \textbf{PEP-MEK} framework, which yields an average improvement of 9.90\% in emotion hallucination detection across selected models.
Resources will be available at \url{https://github.com/xxtars/EmotionHallucer}.

\end{abstract}

\vspace{-1.5em}
\setlength{\epigraphwidth}{0.6\textwidth}  
\epigraph{\textit{Inevitably, emotions are inseparable from the idea of good and evil.}}{~Antonio Damasio, \textit{The Feeling of What Happens}.}
\vspace{-1.5em}

\section{Introduction}

Emotion understanding is one of the most fundamental yet challenging tasks in artificial intelligence~\cite{koelstra2011deap, hakak2017emotion}. Over the past decades, this field has attracted significant attention from the research community~\cite{nandwani2021review, li2020deep, el2011survey, ezzameli2023emotion, rahdari2019multimodal, xing2024emo}. 
Much of the existing work has focused on independent sub-tasks across different modalities: 
in the text modality, tasks such as sentiment analysis~\cite{wankhade2022survey} and emotion cause detection~\cite{lee2010text}; 
in the image modality, facial expression recognition~\cite{li2020deep} and affective scene analysis~\cite{zhao2021affective}; 
in the speech modality, speech emotion recognition~\cite{wani2021comprehensive}; 
and in the video modality, multimodal emotion recognition~\cite{abdullah2021multimodal}, dynamic facial expression recognition~\cite{zhao2021former}, and body gesture-based emotion recognition~\cite{liu2021imigue}, among others.

Recently, MLLMs have demonstrated remarkable capabilities in text and vision understanding~\cite{alayrac2022flamingo, li2023blip, achiam2023gpt}, and have begun to play an increasingly important role in emotion understanding~\cite{zhang2023dialoguellm, lian2023explainable, lian2025affectGPT, cheng2024emotion, xing2024emo}. 
These models offer the potential for unified, multimodal emotion understanding that transcends traditional unimodal and single-task approaches.

However, despite their strong capabilities, MLLMs often generate incorrect or ungrounded responses based on textual or visual inputs~\cite{li2023evaluating, tong2024eyes, petryk2024aloha}. 
This issue is commonly called ``hallucination''\cite{rohrbach2018object}. Hallucination is generally categorized into two types \cite{bai2024hallucination}:
(1) factuality hallucination, where outputs conflict with real-world facts;
(2) faithfulness hallucination, where outputs diverge from input instructions, provided context, or exhibit internal inconsistencies. 
In response to these challenges, growing attention has been paid to analyzing and mitigating hallucinations in MLLMs. 
Existing hallucination benchmarks, however, are primarily designed for general-purpose tasks~\cite{wang2024videohallucer}, leaving hallucinations in emotion understanding tasks largely unexplored. 
Unlike general tasks, emotion understanding involves subjective and psychological processes, often shaped by cognitive appraisal, context, and social cues, which makes emotion-related hallucinations harder to detect.

\begin{figure}[t]
    \centering
    \begin{subfigure}[b]{0.55\linewidth}
        \centering
        \includegraphics[width=\linewidth]{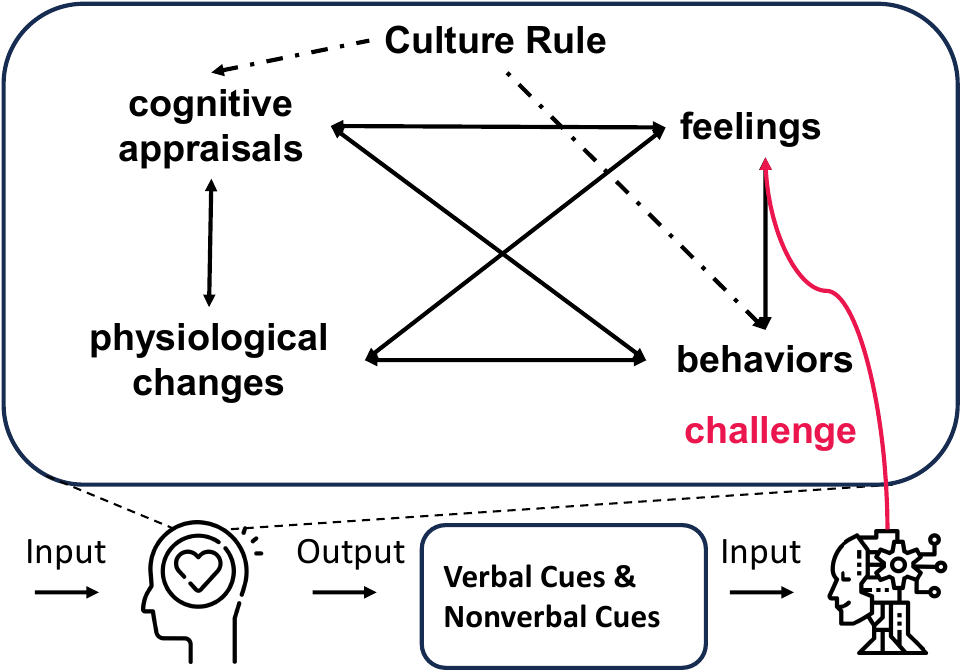}
        \caption{}
        \label{fig:challenage}
    \end{subfigure}
    \begin{subfigure}[b]{0.4\linewidth}
        \centering
        \includegraphics[width=\linewidth]{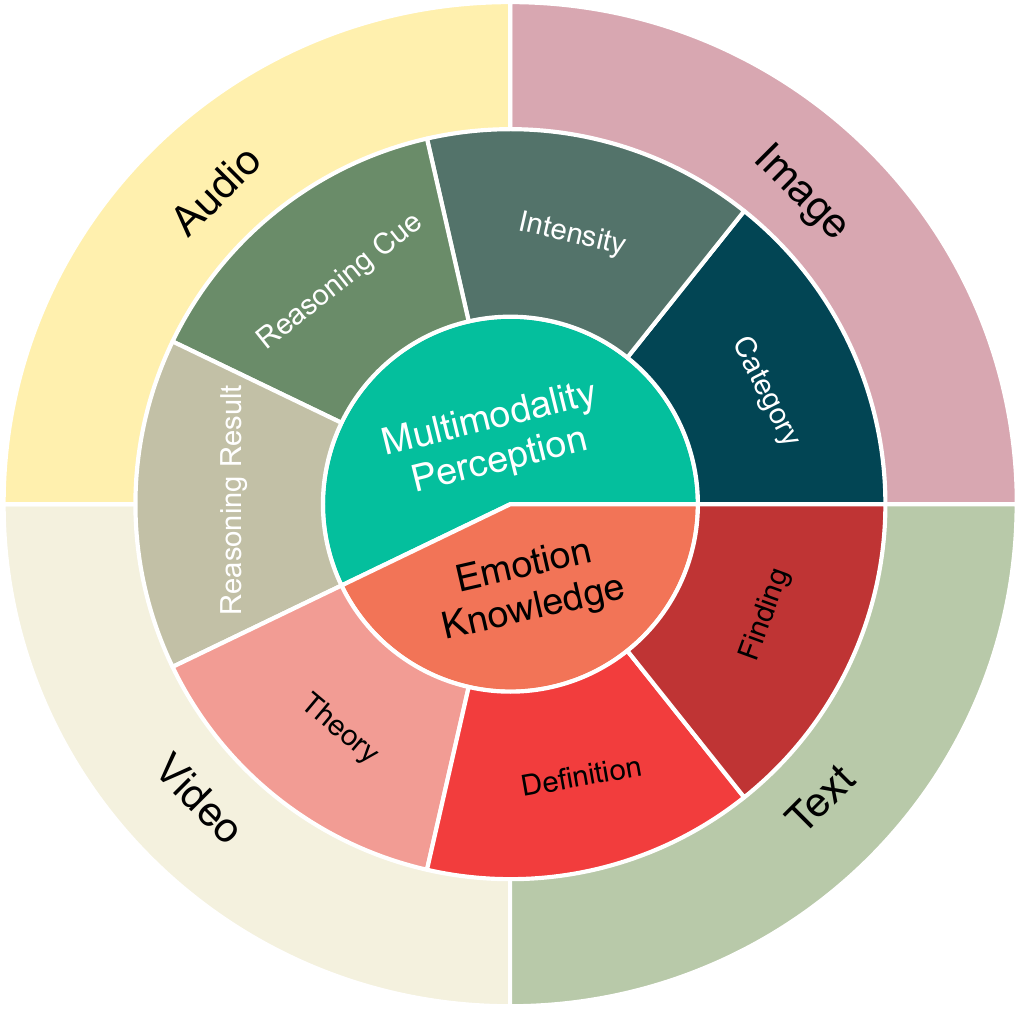}
        \caption{}
        \label{fig:category}
    \end{subfigure}
    \vspace{-0.5em}
    \caption{Emotion understanding differences and the EmotionHallucer. \textbf{(a)} The distinction between how humans and MLLMs understand emotions. Based on the component process model~\cite{scherer2009dynamic} and dynamic systems approach~\cite{lewis2005bridging}, human emotion understanding involves dynamic interactions among cognitive appraisals, physiological changes, feelings, and behaviors. 
    In contrast, MLLMs rely on data-driven learning from external behavioral cues, which limits their ability to accurately infer the underlying emotional states. 
    \textbf{(b)} EmotionHallucer is organized along two main dimensions, Emotion Knowledge and Multimodality Perception, and includes seven subcategories across four modalities.}
    \label{fig:motivation}
    \vspace{-1em}
\end{figure}

On the other hand, human emotion arises from a combination of innate biological mechanisms and lifelong social learning~\cite{zeidner2003development}, which makes it challenging to model, as shown in Figure~\ref{fig:challenage}. 
In contrast, MLLMs rely on data-driven learning, and lack the embodied and experiential grounding that humans use to interpret emotions naturally and intuitively. 
Fortunately, \textit{\textbf{learning how emotion develops may help us understand more about emotion itself}}~\cite{shiota2017emotion}. 
Decades of research in psychology have offered rich insights into how humans perceive, process, and reason about emotions~\cite{niedenthal2017psychology}, offering a valuable knowledge source to support more reliable emotion understanding in MLLMs.

Motivated by these observations, we first consider how emotion-related hallucinations should be defined and categorized. 
Unlike general hallucinations, emotion hallucinations tend to be more complex, as emotion understanding involves not only objective perception but also psychological and sociological reasoning~\cite{zaki2012neuroscience}.
Given this complexity, we propose \textbf{EmotionHallucer}, the first benchmark specifically designed to evaluate emotion hallucinations.
EmotionHallucer targets two key aspects: hallucinations related to emotion psychology knowledge (focusing on factuality hallucination) and hallucinations in real-world multimodal emotion understanding (focusing on faithfulness hallucination), as illustrated in Figure~\ref{fig:category}.
To ensure reliable evaluation and reduce confounding factors~\cite{li2023evaluating,zhang2023language}, we adopt a binary QA-based evaluation framework\cite{li2023evaluating,wang2024videohallucer}. 
Specifically, we construct adversarial QA pairs\cite{tong2024eyes}, where each pair consists of a basic and an intentionally hallucinated question to test models. To mitigate language bias, we balance ``yes'' and ``no'' answers, and provide concise explanations to reduce misinterpretation.

By evaluating 38 LLMs and MLLMs on EmotionHallucer, our analysis leads to three mian findings:
\textbf{First}, most current models exhibit substantial issues with emotion hallucinations.
\textbf{Second}, closed-source models outperform open-source ones in detecting emotion hallucinations, and reasoning capability provides additional advantages.
\textbf{Third}, existing models perform better in emotion psychology knowledge than in multimodal emotion perception.
Building on these findings, we propose \textbf{PEP-MEK}, a plug-and-play framework that incorporates both modality-specific and emotional knowledge to mitigate emotion hallucinations.
Experimental results show that applying PEP-MEK leads to a significant performance boost on EmotionHallucer, with an average improvement of 9.90\%.
We believe this framework can support future research and development in the detection and mitigation of emotion hallucinations in MLLMs.

\section{Related Work}

\subsection{Hallucination in Natural Language Processing}

Generative models in Natural Language Processing (NLP), particularly Large Language Models (LLMs), have demonstrated impressive performance across a wide range of language generation tasks. However, a major challenge remains: these models may occasionally produce text that is inaccurate, irrelevant, or illogical, a phenomenon commonly referred to as ``hallucination'' in NLP.
Hallucination typically refers to instances where the generated content is nonsensical or deviates from the intended meaning or source material~\cite{filippova2020controlled}. 
In the NLP community, this issue is empirically categorized into two types~\cite{bai2024hallucination}:
1) Factuality hallucination, which highlights inconsistencies between generated output and verifiable real-world facts, often manifesting as factual errors or fabrication;
2) Faithfulness hallucination, which refers to deviations from user instructions, input context, or internal consistency within the generated content.
Within specialized research domains, opinions diverge regarding the value of factuality hallucinations.
Some studies suggest that such hallucinations can be beneficial~\cite{maynez2020faithfulness,thomson2020gold}, arguing that the additional information they introduce may enhance the perceived informational value of the output.

\subsection{Hallucination in Computer Vision}

Recent advances in vision-language modeling have led to impressive performance across various generative tasks~\cite{alayrac2022flamingo,li2023blip,achiam2023gpt}.
Alongside these developments, increasing attention has been given to the issue of hallucination in this domain.
The concept of object hallucination in image captioning, along with the CHAIR metric, was first introduced by Rohrbach et al.\cite{rohrbach2018object}.
To provide a more robust evaluation framework, POPE\cite{li2023evaluating} proposed a binary VQA benchmark specifically aimed at detecting object hallucinations, offering greater reliability than CHAIR.
Subsequent studies have broadened the scope of hallucination research to include relationships, attributes, counting, OCR, and other visual phenomena~\cite{sun2023aligning,wang2023amber,guan2024hallusionbench,cui2023holistic,chen2024unified,liu2024phd}. 
More recently, research has extended to video-based hallucinations, reflecting the growing complexity of multimodal understanding~\cite{zhang2024eventhallusion,wang2024videohallucer}.
However, to the best of our knowledge, despite the emergence of general hallucination benchmarks, no benchmark has yet been developed to evaluate hallucinations related to emotion understanding.
To fill this gap, we introduce the first benchmark specifically designed to assess emotion hallucinations.

\subsection{Emotion MLLMs}

With the rapid advancement of LLMs, a growing body of research has begun to explore their potential for emotion understanding. These models facilitate the integration of multimodal information, making complex emotional reasoning increasingly feasible. Representative works in this direction include AffectGPT\cite{lian2025affectGPT}, EMER\cite{lian2023explainable}, Emotion-LLAMA\cite{cheng2024emotion}, and Omni-Emotion\cite{yang2025omni}, which investigate how MLLMs can be adapted for emotion recognition and reasoning.
In parallel, other studies have focused on more domain-specific emotion understanding tasks~\cite{li2024eald, xing2024emo, li2025deemo}, contributing to increasingly flexible and specialized frameworks.
However, despite these advancements, the critical issue of hallucination in emotion understanding remains largely underexplored. One key reason for this gap is the lack of a dedicated benchmark to assess hallucinations in emotion understanding.
In this work, we address this limitation by introducing the first benchmark specifically designed to evaluate hallucination in emotion understanding.

\section{The \benchtitle Benchmark}


\subsection{Benchmark Construction}\label{sub:construction}

\begin{figure}[t]
    \centering
    \includegraphics[width=\linewidth]{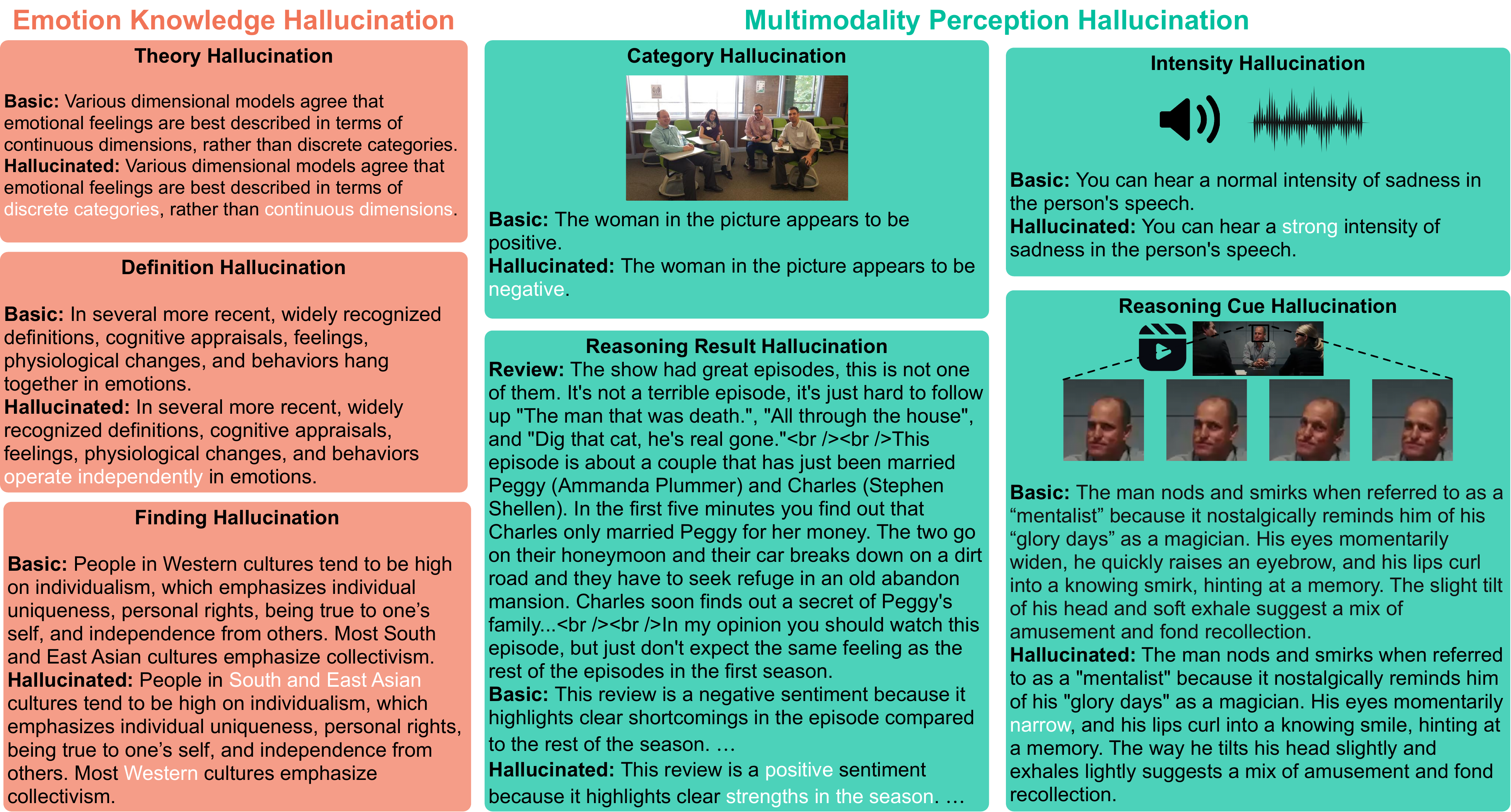}
    \caption{Example tasks in EmotionHallucer. Each pairs consists of a basic question, used to test the basic ability of MLLMs, and a hallucinated question, containing hallucinated content to evaluate the models' ability to detect hallucination. Emotion Knowledge Hallucination targets emotion psychology knowledge~\cite{scherer2009dynamic,lewis2005bridging,shiota2017emotion}, whereas Multimodality Perception Hallucination centers on real-world emotion understanding~\cite{deng2023soul,zhang2018adaptive,livingstone2018ryerson,lian2023mer,zadeh2019social,socialiq2github}.
    More details could be seen in \textbf{Appendix}~\ref{app:examples}.}
    \label{fig:example}
    \vspace{-1em}
\end{figure}

To evaluate hallucination in emotion understanding, we divide our benchmark into two primary categories: emotion psychology knowledge and real-world multimodal emotion perception.
This design yields seven specific evaluation settings spanning four modalities, as illustrated in Figure~\ref{fig:example}.
For the emotion knowledge dimension, we collect and curate factual statements from an authoritative textbook in emotion psychology~\cite{shiota2017emotion}.
For the multimodal perception dimension, we leverage several widely used datasets across different modalities: 
SOUL~\cite{deng2023soul} for text, 
Twitter15 and Twitter17 ~\cite{zhang2018adaptive} for image, 
RAVDESS~\cite{livingstone2018ryerson} for speech, 
and MER 2023~\cite{lian2023mer} and Social-IQ 2.0~\cite{zadeh2019social,socialiq2github} for video. 
These diverse sources allow us to construct hallucination instances that reflect both knowledge-based and perceptual challenges in emotion understanding.
The detailed annotation procedure is described in \textbf{Appendix}~\ref{app:anno}.

\subsubsection{Emotion Psychology Knowledge Hallucination}

To the best of our knowledge, this is the first hallucination benchmark that evaluates LLMs and MLLMs on their understanding of emotion psychology knowledge, with a focus on core theories from affective science~\cite{scherer2009dynamic,lewis2005bridging}.
We begin by selecting a set of well-established and unambiguous statements from an authoritative textbook in emotion psychology~\cite{shiota2017emotion}, which serve as ground truth.
Based on these statements, we construct hallucinated counterparts that intentionally contradict, distort, or misrepresent the original content.
This setting enables a examination of models’ susceptibility to hallucination within the domain of emotion knowledge, providing a framework to assess both knowledge grounding and hallucination behaviors in the context of human emotion theory.

\textbf{Theory.}
Our work introduces a Theory Hallucination setting in the context of emotion psychology theory~\cite{cannon1927james,scherer2009dynamic,lewis2005bridging}, as illustrated in Figure~\ref{fig:example}.
To support this setting, we collect a set of core statements derived from well-established theoretical frameworks in emotion psychology, covering a range of foundational perspectives~\cite{cannon1927james,scherer2009dynamic,lewis2005bridging,shiota2017emotion}.
Based on these statements, and through the annotation process, we construct 81 question-answer pairs that serve as test instances for evaluating hallucination in emotion theory knowledge.

\textbf{Definition.}
We also construct an emotion Psychology Definition Hallucination setting, which targets the definitions of widely accepted concepts in emotion psychology~\cite{berkowitz1999anger}, as shown in Figure~\ref{fig:example}.
To support this setting, we collect a set of definitions for key emotion-related terms from authoritative academic sources, ensuring that each definition reflects the consensus within the field.
This process results in 133 question-answer pairs, which serve as evaluation instances for assessing hallucinations related to definitional knowledge in affective science.

\textbf{Finding.}
We further construct an emotion Psychology Finding Hallucination setting, which focuses on empirical findings in emotion research that describe observed phenomena but have not yet been formalized into complete theoretical frameworks~\cite{kagitcibasi1997individualism}, as shown in Figure~\ref{fig:example}.
This category includes well-documented results such as cross-cultural differences in emotional expression~\cite{hareli2015cross}, developmental variations between infants and adults~\cite{best2013cost}, and other empirical observations reported in the literature.
This setting yields 178 question-answer pairs, which enable the evaluation of hallucination behaviors related to non-theoretical but empirically grounded knowledge in emotion psychology.

\subsubsection{Multimodality Perception Hallucination}

Beyond emotion psychology knowledge, we also construct a real-World Multimodal Emotion Perception Hallucination setting, which focuses on assessing hallucination risks in emotion understanding tasks involving multimodal inputs (e.g., text, audio, and visual signals), as shown in Figure~\ref{fig:example}.
To build this setting, we collect samples from widely used emotion understanding datasets~\cite{deng2023soul,zhang2018adaptive,livingstone2018ryerson,lian2023mer,zadeh2019social,socialiq2github}.
We then filter and refine these samples to construct instances in which hallucinated descriptions contradict or misinterpret the emotional content conveyed by the inputs.
This setting enables the evaluation of hallucinations in realistic, multimodal environments, where emotion understanding requires integrating ambiguous and context-dependent cues from multiple sources~\cite{zadeh2017tensor}.

\textbf{Category.}
Following prior work on object hallucination~\cite{rohrbach2018object}, and as shown in Figure~\ref{fig:example}, we introduce an emotion Category Hallucination setting, which refers to the incorrect generation or identification of emotion categories~\cite{dzedzickis2020human}.
Specifically, this type of hallucination refers to cases where models generate nonexistent or inappropriate emotion category. 
Additionally, the emotion categories in this setting are not limited to binary sentiment labels commonly used in text-based tasks (e.g., positive/negative), but also cover discrete basic emotion categories frequently adopted in visual and multimodal emotion recognition tasks (e.g., happiness, anger, sadness). 

\textbf{Intensity.}
We further introduce an emotion Intensity Hallucination setting, which concerns the misrepresentation of the strength of emotional expressions, as shown in Figure~\ref{fig:example}.
Unlike traditional approaches in affective computing that represent emotions using continuous valence-arousal dimensions~\cite{russell1980circumplex}, we adopt discrete intensity descriptors (e.g., mild/slightly, normal, strong) that are more aligned with the representational capabilities of LLMs.
In this setting, hallucination arises when the model exaggerates, downplays, or otherwise inaccurately describes the intensity of the emotional state conveyed by the input.

\textbf{Reasoning Result.}
Following previous works~\cite{lian2023mer, cheng2024emotion}, we introduce a Reasoning Result Hallucination setting. This phenomenon arises when the model accurately extracts emotional cues from multimodal inputs (e.g., facial expressions, vocal tone, or text), but still produces an incorrect emotional interpretation, as shown in Figure~\ref{fig:example}.
In this setting, the hallucination does not stem from the misidentification of input signals themselves, but from incorrect emotional reasoning. 
This design reflects the critical distinction between perception and reasoning in emotion understanding, highlighting that accurate signal recognition does not necessarily guarantee appropriate emotional conclusions.

\textbf{Reasoning Cue.}
Following previous works~\cite{lian2023mer, cheng2024emotion}, we introduce a Reasoning Cue Hallucination setting, as shown in Figure~\ref{fig:example}. 
In this setting, hallucination arises when the model overlooks, misinterprets, or fabricates key multimodal signals (e.g., missing the angry tone in speech, misreading a facial expression, or inferring unsupported emotional context from text), leading to unreliable reasoning processes. 
Crucially, Reasoning Cue Hallucination captures cases where the reasoning pathway itself is flawed due to incorrect cue selection or interpretation, independent of the correctness of the final emotion judgment.

\subsection{Benchmark Statistics}\label{sub:statistics}

\begin{figure}[t]
\hspace{-1em}
        \begin{minipage}{0.65\linewidth}
            \centering
            \includegraphics[width=\textwidth]{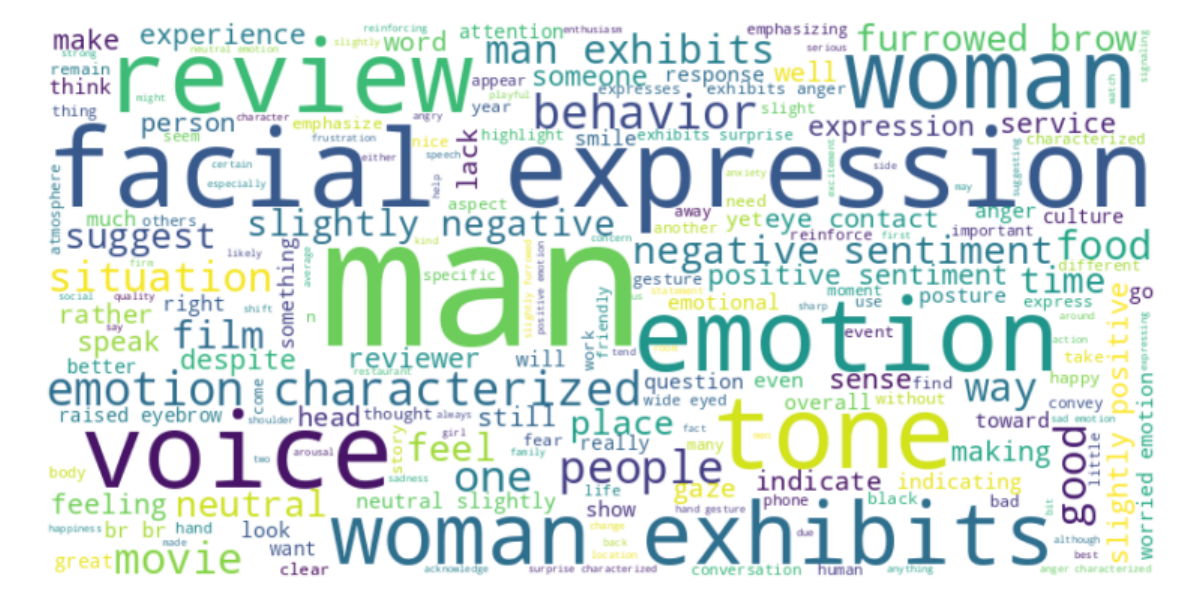}
            \caption{Word cloud of EmotionHallucer.}
            \label{fig:cloud}
        \end{minipage}
        \begin{minipage}{0.27\linewidth}
            \centering
            \tiny
            \caption{Dataset statistics of EmotionHallucer.}
            \label{tab:statistics}
            \begin{tabular}{lc}
                \toprule
                \textbf{Statistic} & \textbf{Count} \\
                \midrule
                Questions & 2,742 \\
                Images & 150 \\
                Audios & 368 \\
                Videos & 230 \\
                Avg Question len & 31.6 \\
                Avg Knowledge Text len & 19.9 \\
                Avg Review Text len & 108.7 \\
                Avg Image resolution & 579.5 $\times$ 466.5 \\
                Avg Audio len & 4.2 \\
                Avg Short Video len & 4.3 \\
                Avg Short Vieo resolution & 870.0 $\times$ 476.9 \\
                Avg Long Video len & 60.0 \\
                Avg Long Vieo resolution & 637.9 $\times$ 360.0 \\
                \bottomrule
            \end{tabular}
        \end{minipage}
        \vspace{-1.5em}
\end{figure}

\textbf{Quantitative Analysis.}
As shown in Table~\ref{tab:statistics}, we provide a quantitative analysis of EmotionHallucer. In total, the benchmark comprises 2,742 questions, with an average length of 31.6 words. EmotionHallucer covers four modalities: text, image, audio, and video.
We further categorize the data collected from MER 2023\cite{lian2023mer} and Social-IQ 2.0\cite{zadeh2019social,socialiq2github} into short video and long video subsets.
For the emotion knowledge text, lengths range from 4 to 56 words, with an average of 19.9 words.
In the case of real-world review texts\cite{deng2023soul} (see task details in \textbf{Appendix}~\ref{app:anno}), the length ranges from 12 to 199 words, averaging 108.7 words.
For real-world images~\cite{zhang2018adaptive}, resolutions range from 320 $\times$ 194 to 600 $\times$ 1024 pixels, with an average resolution of 579.5 $\times$ 466.5 pixels.
The real-world audio clips~\cite{livingstone2018ryerson} vary from 3.0 to 6.3 seconds in length, with an average duration of 4.2 seconds.
For real-world videos, short videos~\cite{lian2023mer} have an average resolution of 870.0 $\times$ 476.9 pixels, ranging from 576 $\times$ 256 to 1280 $\times$ 720 pixels. Their durations range from 0.52 to 38.29 seconds, with an average of 4.29 seconds. 
In contrast, long videos~\cite{zadeh2019social,socialiq2github} have a fixed duration of 60 seconds. Their resolution averages 637.9 $\times$ 360.0 pixels, ranging from 534 $\times$ 360 to 640 $\times$ 360 pixels.

\textbf{Qualitative Analysis.}
To provide a more intuitive understanding of VideoHallucer, we visualize key terms using a word cloud, as shown in Figure~\ref{fig:cloud}.
The visualization highlights frequently occurring concepts such as emotion, voice, facial expression, and tone, which correspond closely to the core modalities and reasoning components in our benchmark.
These results suggest that VideoHallucer effectively focuses on the multimodal cues essential for emotion hallucination.

\subsection{Evaluation Metric}\label{sub:evaluation}

\textbf{Hallucination Evaluation.} 
Following previous work~\cite{tong2024eyes,wang2024videohallucer}, we adopt a QA-based benchmark for the following reasons:
(1) Susceptibility to External Factors:
Caption-based evaluations, like BLEU~\cite{papineni2002bleu} and ROUGE~\cite{lin2004rouge}, are sensitive to external variables such as prompt design and caption length~\cite{li2023evaluating}, which can distort evaluation outcomes.
(2) Evaluation Complexity:
Existing approaches such as CHAIR~\cite{rohrbach2018object} rely on intricate, manually designed parsing rules, which complicate the evaluation process and hinder scalability.
(3) LLM Hallucination Bias:
Given the propensity of LLMs to produce hallucinated outputs, using their own generations for self-evaluation may compromise the reliability and objectivity of the results~\cite{wang2023amber}.

To reduce these evaluation biases, we propose EmotionHallucer with an adversarial evaluation framework inspired by~\cite{tong2024eyes,wang2024videohallucer}. 
For each evaluation instance, we construct a pair of complementary questions:
a basic question, which tests the model’s core perception and reasoning abilities, and a hallucinated question, which introduces intentionally fabricated content to test the model’s robustness against hallucination.
A response is considered correct only if the model answers both questions accurately as a pair.
This dual-question design enables a more rigorous assessment of whether a model can detect and resist hallucinations without compromising its performance on fundamental tasks.

\textbf{Bias Evaluation.}
In addition to the accuracy, we calculate the Yes Percentage Difference (Pct. Diff) and False Positive Ratio (FP Ratio)~\cite{guan2024hallusionbench,wang2024videohallucer} to reveal the bias of these MLLMs. Specifically, the Yes Percentage Difference is calculated as
\begin{equation}
    d_{y} = \frac{ \left| \{ Pred(m, q) = \text{``yes''} \}_{(m, q) \in V} \right| - \left| \{ GT(m, q) = \text{``yes''} \}_{(m, q) \in V} \right| }{ |V| },
\end{equation}
where $V$ is the set of question pairs. $m$ represents additional modality information such as image, audio, or video; for text-only questions, this component is absent. $q$ refers to the question itself. $GT(m, q)$ is the ground truth. A smaller $d_{y}$ indicates the number of ``yes'' responses from models is closer to the ground truth, revealing less language bias. the False Positive Ratio is calculated as
\begin{equation}
    r_{fp} = \frac{ \left| \{ Pred(m, q) = \text{``yes''} \}_{(m, q) \in W} \right| }{ |W| },
\end{equation}
where $W$ is the set of wrongly answered question pairs. $r_{fp}$ demonstrates the percentage of ``yes'' in all wrongly predicted answers. A value closer to 50\% indicates less bias from the models.

\section{Experiments}

In this section, we evaluate a range of widely used LLMs and MLLMs on EmotionHallucer.
The models are grouped according to the input modalities they support.
Implementation details and additional analyses are provided in \textbf{Appendix}~\ref{app:implementation}.

\begin{table}[t]
 \caption{Performance comparison on EmotionHallucer with additional ``Yes/No bias'' analysis.}
  \label{tab:sota-e}
  \centering
  \footnotesize
  \setlength{\tabcolsep}{5pt}
  \begin{tabular*}{\textwidth}{lcccccc}
    \toprule
    \multirow{2}{*}{Methods} & Model & \multicolumn{2}{c}{Yes/No Bias} & \multicolumn{3}{c}{Accuracy on EmotionHallucer}\\ 
    \cmidrule(rl){3-4} \cmidrule(rl){5-7}
    & Size & Pct. Diff ($\sim$0) & FP Ratio ($\sim$0.5)& Basic~$\uparrow$ & Hallucinated~$\uparrow$ & Overall~$\uparrow$ \\
    \midrule
    \multicolumn{7}{c}{\textit{Open-source}} \\
    Qwen2.5-Omni~\cite{xu2025qwen2} & 7B & -0.05 & 0.44 & 52.81 & 63.46 & 18.65 \\
    Emotion-LLaMA~\cite{cheng2024emotion} & 7B & 0.20 & 0.71 & 72.88 & 33.45 & 15.43 \\
    \midrule
    \multicolumn{7}{c}{\textit{Closed-source}} \\
    Gemini-2.5-Flash~\cite{gemini2025update} & - & 0.01 & 0.51 & 69.41 & 68.15 & 45.06 \\
    Gemini-2.5-Pro~\cite{gemini2025update} & - & 0.01 & 0.52 & 70.30 & 67.56 & 44.17 \\
    \bottomrule
  \end{tabular*}
  \vspace{-1em}
\end{table}

\subsection{Main Benchmark Results}

\textbf{Multimodality.}
The EmotionHallucer benchmark encompasses four modalities: text, image, audio, and video.
However, most existing MLLMs lack the ability to process all four modalities simultaneously, with many limited to two (e.g., text-image or text-video). 
To ensure a fair and consistent evaluation, we report both aggregate results for models capable of handling all four modalities and modality-specific results for models restricted to certain subsets. 
Additional analyses on model performance and modality-specific analysis are provided in \textbf{Appendix}~\ref{app:experiment}.

As shown in Table~\ref{tab:sota-e}, we compare models across all modalities available on EmotionHallucer. 
While the strongest closed-source Gemini models outperform their open-source counterparts, their overall performance remains suboptimal, reflecting the inherent challenges of emotional understanding and hallucination.
Notably, all open-source models fail to exceed the 25\% accuracy expected from random guessing, also underscoring the current limitations of MLLMs in handling emotional reasoning.
In the ``Yes/No Bias'' evaluation, general-purpose MLLMs demonstrate better neutrality, showing less tendency toward overconfident affirmations. In contrast, the emotion-specific model Emotion-LLaMA performs significantly worse in this regard.
This may be attributed to its fine-tuning focus on emotional content. It may also be due to the fact that the model is relatively outdated compared to recent MLLMs, potentially lacking the latest advances.

Given that most popular MLLMs are optimized for vision modalities and lack support for audio-only inputs, we present EmotionHallucer–NoAudio, a benchmark subset that excludes the audio modality (see Table~\ref{tab:sota-ewoa}).
Due to space constraints, we report results for 11 representative models.
Consistent with earlier findings, closed-source models outperform their open-source counterparts in both overall accuracy and the ``Yes/No Bias'' evaluation.
Gemini-2.5-Pro achieves the best performance, followed closely by Gemini-2.5-Flash.
Among open-source models, Qwen2.5 VL performs the best and notably exceeds the random-guessing baseline.

\begin{table}
 \caption{Performance comparison on EmotionHallucer-NoAudio with additional ``Yes/No bias'' analysis. More results could be seen in Table~\ref{tab:sota-ewoa-E} of the \textbf{Appendix}.}
   \label{tab:sota-ewoa}
  \centering
  \footnotesize
  \setlength{\tabcolsep}{4pt}
  \begin{tabular*}{\textwidth}{lcccccc}
    \toprule
    \multirow{2}{*}{Methods} & Model & \multicolumn{2}{c}{Yes/No Bias} & \multicolumn{3}{c}{Accuracy on EmotionHallucer–NoAudio}\\ 
    \cmidrule(rl){3-4} \cmidrule(rl){5-7}
    & Size & Pct. Diff ($\sim$0) & FP Ratio ($\sim$0.5)& Basic~$\uparrow$ & Hallucinated~$\uparrow$ & Overall~$\uparrow$ \\
    \midrule
    \multicolumn{7}{c}{\textit{Open-source}} \\
    LLaVA~\cite{liu2023visual} & 34B & -0.05 & 0.45 & 50.25 & 59.52 & 10.27 \\
    Llama3.2-vision~\cite{grattafiori2024llama} & 11B & 0.21 & 0.78 & 83.05 & 41.28 & 29.91 \\
    Video-ChatGPT~\cite{maaz2023video} & 7B & 0.09 & 0.59 & 61.91 & 44.67 & 18.44 \\
    Emotion-LLaMA~\cite{cheng2024emotion} & 7B & 0.12 & 0.63 & 66.55 & 42.43 & 18.86 \\
    Qwen2.5-VL~\cite{bai2025qwen2} & 72B & 0.08 & 0.63 & 78.08 & 62.15 & 43.02 \\
    Qwen2.5-Omni~\cite{xu2025qwen2} & 7B & 0.11 & 0.65 & 72.39 & 49.74 & 25.44 \\
    \midrule
    \multicolumn{7}{c}{\textit{Closed-source}} \\
    QvQ-Max~\cite{qwen2025QvQ72b} & - & 0.07 & 0.63 & 78.18 & 63.39 & 47.98 \\
    GPT-4o~\cite{hurst2024GPT} & - & -0.01 & 0.48 & 67.10 & 69.49 & 40.98 \\
    GPT-4.1~\cite{openai2024GPT4} & - & 0.05 & 0.58 & 74.58 & 64.71 & 44.47 \\
    Gemini-2.5-Flash~\cite{gemini2025update} & - & 0.06 & 0.61 & 78.55 & 66.80 & 50.56 \\
    Gemini-2.5-Pro~\cite{gemini2025update} & - & 0.07 & 0.64 & 81.31 & 67.01 & 51.58 \\
    \bottomrule
  \end{tabular*}
\end{table}

\begin{figure}
    \centering
    \vspace{-1em}
    \includegraphics[width=\linewidth]{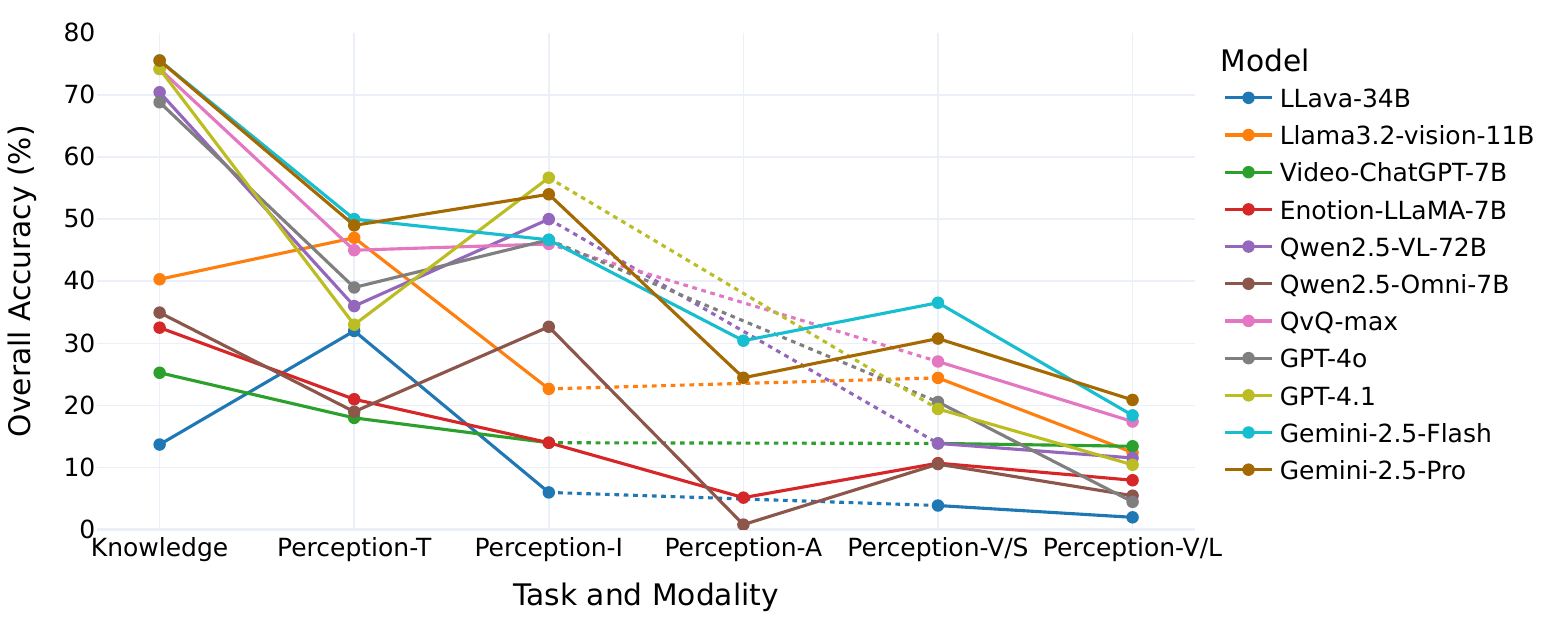}
    \caption{Unimodal performance of partial selected models. T, I, A, V/S, and V/L stand for Text, Image, Audio, Short Video, and Long Video, respectively.Additional models and implementation details are provided in the \textbf{Appendix}~\ref{app:experiment}.}
    \label{fig:unimodality}
    \vspace{-1.5em}
\end{figure}

\textbf{Unimodality.} 
To enable a more fine-grained understanding of MLLM behavior across tasks and modalities, we report performance on each individual modality.
As shown in Figure~\ref{fig:unimodality}, models perform best on emotion knowledge text, with accuracy decreasing steadily across perception-based tasks, from text to image, and further to audio and video.
This trend may be attributed to two main factors:
(1) current training pipelines are predominantly focused on text data, enhancing models’ performance on knowledge-oriented tasks, and
(2) a lack of high-quality emotional annotations in modalities, which limits the models’ ability to learn fine-grained emotion understanding.
These results underscore a critical future direction: improving cross-modal emotion understanding through better data quality and balanced modality training.
Additional detailed results and analysis are provided in Appendix~\ref{app:experiment}.

\section{Predict-Explain-Predict with Modality and Emotion Knowledge}

We observe that most models perform significantly worse under the multimodal perception hallucination setting compared to the emotion knowledge hallucination setting in EmotionHallucer.
Motivated by this observation, we propose a simple yet effective explanation-based method to mitigate hallucination issues arising from multimodal emotion perception errors.

\subsection{Emotion Knowledge vs. Multimodal Perception Hallucination Detection}

\begin{figure}[t]
    \centering
    \begin{subfigure}[b]{0.6\linewidth}
        \centering
        \includegraphics[width=\linewidth]{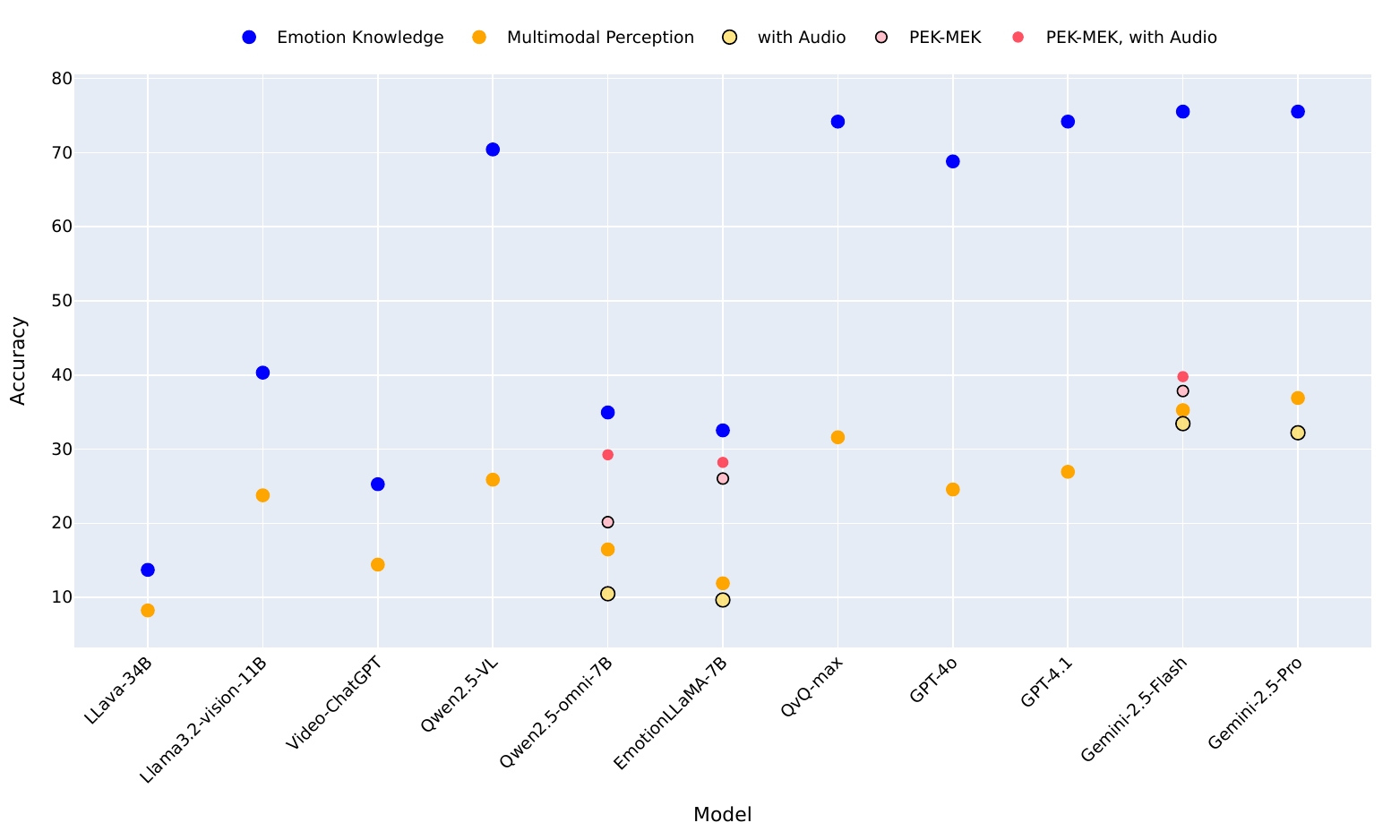}
        \caption{}
        \label{fig:cot_motivation}
    \end{subfigure}
    \begin{subfigure}[b]{0.33\linewidth}
        \centering
        \includegraphics[width=\linewidth]{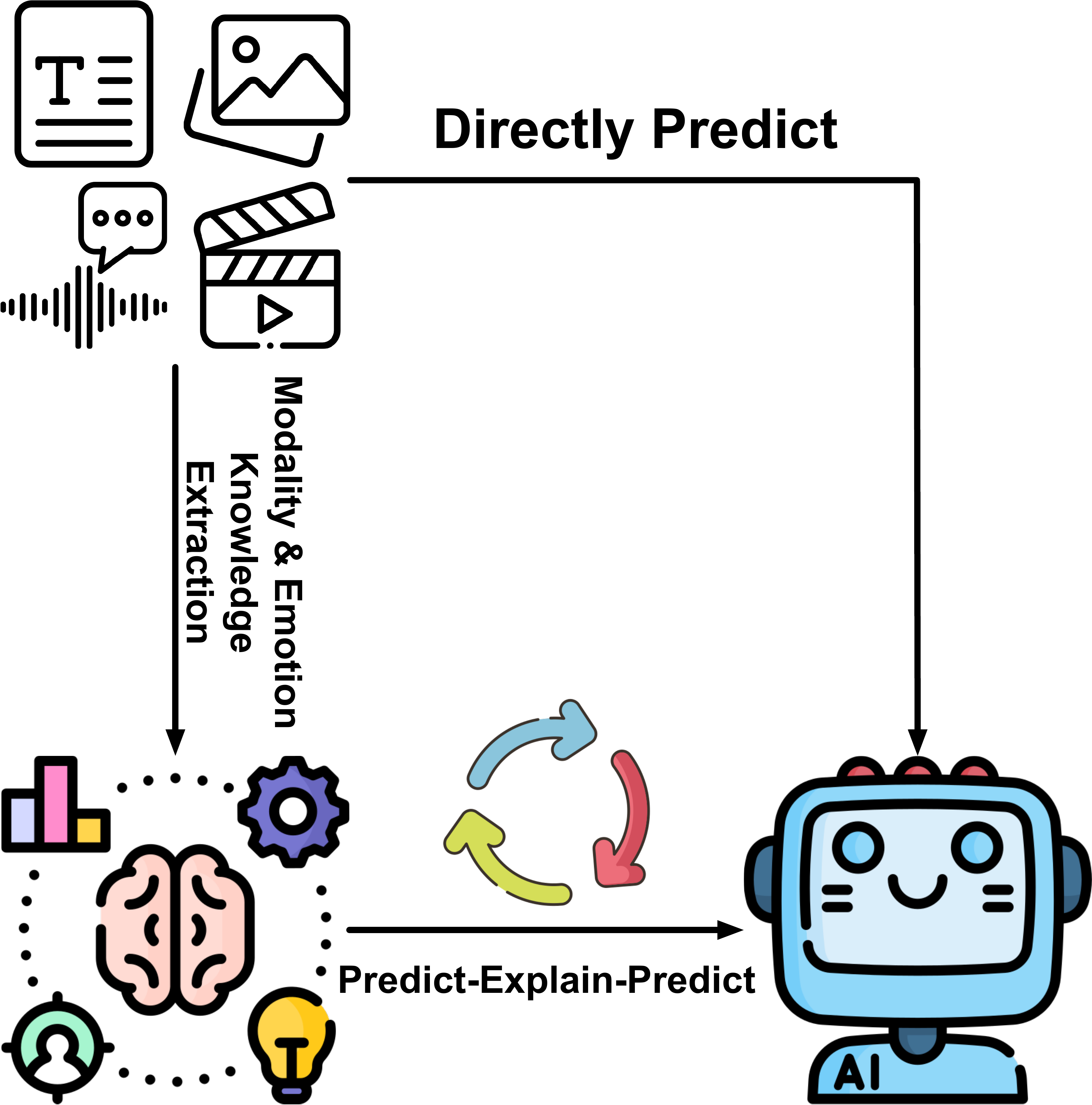}
        \caption{}
        \label{fig:PEP}
    \end{subfigure}
    \vspace{-1em}
    \caption{Emotion hallucination analysis and our proposed framework. \textbf{(a)} Comparison of hallucination detection in emotion knowledge and multimodal perception.
    \textbf{(b)} The PEP-MEK Framework.} 
    \vspace{-1em}
\end{figure}

\begin{table}[t]
 \caption{Results of PEP-MEK Framework.}
  \label{tab:PEP-MEK}
  \centering
  \footnotesize
  \setlength{\tabcolsep}{5pt}
  \begin{tabular*}{\textwidth}{lccccccc}
    \toprule
    \multirow{2}{*}{Methods} & Model & \multicolumn{2}{c}{Yes/No Bias} & \multicolumn{3}{c}{Accuracy on EmotionHallucer-P}\\ 
    \cmidrule(rl){3-4} \cmidrule(rl){5-7}
    & Size & Pct. Diff ($\sim$0) & FP Ratio ($\sim$0.5)& Basic~$\uparrow$ & Hallucinated~$\uparrow$ & Overall~$\uparrow$ \\
    \midrule
    \multicolumn{7}{c}{\textit{Open-source}} \\
    Qwen2.5-Omni~\cite{xu2025qwen2} & 7B & -0.18 & 0.30 & 35.51 & 71.96 & 10.49 \\
    \rowcolor{gray!20}
    \multicolumn{1}{r}{\tiny \textit{+PEP-MEK}} & & -0.19 & 0.29 & 37.49 & 74.87 & 20.15 \\
    Emotion-LLaMA~\cite{xu2025qwen2} & 7B & 0.28 & 0.77 & 77.34 & 22.02 & 9.65 \\
    \rowcolor{gray!20}
    \multicolumn{1}{r}{\tiny \textit{+PEP-MEK}} & & -0.08 & 0.41 & 46.35 & 62.86 & 26.03 \\
    \midrule
    \multicolumn{7}{c}{\textit{Closed-source}} \\
    Gemini-2.5-Flash~\cite{gemini2025update} & - & 0.00 & 0.50 & 61.54 & 62.05 & 33.44 \\
    \rowcolor{gray!20}
    \multicolumn{1}{r}{\tiny \textit{+PEP-MEK}} & & -0.07 & 0.40 & 56.66 & 71.17 & 37.84 \\
    \bottomrule
  \end{tabular*}
  \vspace{-1em}
\end{table}

We present model performance on emotion knowledge and multimodal perception tasks, as shown in Figure~\ref{fig:cot_motivation}.
A key observation is that most models achieve significantly lower accuracy on multimodal perception, highlighting a substantial gap compared to their performance on structured emotion knowledge tasks.
This suggests that while current models handle structured knowledge reasonably well, they struggle with real-world emotion understanding.
Accordingly, this section focuses on enhancing model robustness against multimodal perception hallucinations.
Among the evaluated models, QvQ-Max outperforms Qwen2.5-VL, which we attribute to differences in their underlying reasoning paradigms.
Given that LLMs are generally more adept at fact detection than hallucination detection~\cite{ji2023survey}, we hypothesize that incorporating a more structured and targeted reasoning framework, enriched with emotion-specific knowledge, may further reduce emotion hallucination in MLLMs.

\subsection{PEP-MEK Framework}

Building on these findings, we propose a novel framework, Predict-Explain-Predict with Modality and Emotion Knowledge (PEP-MEK), designed to enhance MLLMs’ performance in detecting multimodal emotion hallucinations.

Figure~\ref{fig:PEP} illustrates the overall architecture of the PEP-MEK framework.
Rather than relying on direct predictions, PEP-MEK introduces an intermediate reasoning stage to improve model transparency and decision reliability in emotion understanding.
First, PEP-MEK uses prompt to guide the model in autonomously extracting modality-specific and emotional knowledge from the input (text, image, audio, video).
This extracted information is then combined with the original input for an initial prediction.
Next, the model generates an explanation for its prediction using both the input and the extracted knowledge.
Then, the explanation is incorporated to refine the prediction.
This process enables the model to reason more effectively and reduce overconfident hallucinations.

As shown in Table~\ref{tab:PEP-MEK}, we evaluate the effectiveness of PEP-MEK on three representative MLLMs: a general-purpose model (Qwen2.5-Omni), an emotion-specific model (Emotion-LLaMA), and a closed-source model (Gemini-2.5-Flash).
The results demonstrate that PEP-MEK consistently improves performance, yielding an average accuracy gain of 9.9\% across models.
Notably, Emotion-LLaMA benefits the most, achieving a 16.38\%t increase in overall accuracy and a substantial reduction in bias-related metrics.
Full implementation details and qualitative examples are provided in Appendix~\ref{app:PEPMEK}.

\section{Conclusion}

In this work, we introduce EmotionHallucer, the first benchmark designed to detect emotion hallucinations in MLLMs.
Our adversarial evaluation strategy provides a rigorous assessment of both emotion understanding and hallucination susceptibility.
By evaluating hallucinations from two complementary perspectives, emotion psychology knowledge and multimodal emotion perception, EmotionHallucer enables fine-grained analysis of large models.
Through testing of 38 LLMs and MLLMs, we uncover the pervasiveness of emotion hallucinations, especially in tasks requiring perception across modalities.
To address these challenges, we propose PEP-MEK, a framework that shows strong potential in reducing emotion hallucinations and improving model robustness.
These findings highlight critical limitations in current emotion MLLMs and point to promising directions for advancing emotion-aware, multimodal reasoning in future research.

\bibliographystyle{plain} 
\bibliography{references} 

\begin{thebibliography}{10}

\bibitem{abdin2024phi}
Marah Abdin, Jyoti Aneja, Harkirat Behl, S{\'e}bastien Bubeck, Ronen Eldan, Suriya Gunasekar, Michael Harrison, Russell~J Hewett, Mojan Javaheripi, Piero Kauffmann, et~al.
\newblock Phi-4 technical report.
\newblock {\em arXiv preprint arXiv:2412.08905}, 2024.

\bibitem{abdullah2021multimodal}
Sharmeen M Saleem~Abdullah Abdullah, Siddeeq Y~Ameen Ameen, Mohammed~AM Sadeeq, and Subhi Zeebaree.
\newblock Multimodal emotion recognition using deep learning.
\newblock {\em Journal of Applied Science and Technology Trends}, 2(01):73--79, 2021.

\bibitem{achiam2023gpt}
Josh Achiam, Steven Adler, Sandhini Agarwal, Lama Ahmad, Ilge Akkaya, Florencia~Leoni Aleman, Diogo Almeida, Janko Altenschmidt, Sam Altman, Shyamal Anadkat, et~al.
\newblock Gpt-4 technical report.
\newblock {\em arXiv preprint arXiv:2303.08774}, 2023.

\bibitem{alayrac2022flamingo}
Jean-Baptiste Alayrac, Jeff Donahue, Pauline Luc, Antoine Miech, Iain Barr, Yana Hasson, Karel Lenc, Arthur Mensch, Katherine Millican, Malcolm Reynolds, et~al.
\newblock Flamingo: a visual language model for few-shot learning.
\newblock {\em Advances in Neural Information Processing Systems}, 35:23716--23736, 2022.

\bibitem{bai2023qwen}
Jinze Bai, Shuai Bai, Yunfei Chu, Zeyu Cui, Kai Dang, Xiaodong Deng, Yang Fan, Wenbin Ge, Yu~Han, Fei Huang, et~al.
\newblock Qwen technical report.
\newblock {\em arXiv preprint arXiv:2309.16609}, 2023.

\bibitem{bai2025qwen2}
Shuai Bai, Keqin Chen, Xuejing Liu, Jialin Wang, Wenbin Ge, Sibo Song, Kai Dang, Peng Wang, Shijie Wang, Jun Tang, et~al.
\newblock Qwen2. 5-vl technical report.
\newblock {\em arXiv preprint arXiv:2502.13923}, 2025.

\bibitem{bai2024hallucination}
Zechen Bai, Pichao Wang, Tianjun Xiao, Tong He, Zongbo Han, Zheng Zhang, and Mike~Zheng Shou.
\newblock Hallucination of multimodal large language models: A survey.
\newblock {\em arXiv preprint arXiv:2404.18930}, 2024.

\bibitem{berkowitz1999anger}
Leonard Berkowitz.
\newblock Anger.
\newblock 1999.

\bibitem{best2013cost}
Catherine~A Best, Hyungwook Yim, and Vladimir~M Sloutsky.
\newblock The cost of selective attention in category learning: Developmental differences between adults and infants.
\newblock {\em Journal of Experimental Child Psychology}, 116(2):105--119, 2013.

\bibitem{cannon1927james}
Walter~B Cannon.
\newblock The james-lange theory of emotions: A critical examination and an alternative theory.
\newblock {\em The American Journal of Psychology}, 39(1/4):106--124, 1927.

\bibitem{chen2024unified}
Xiang Chen, Chenxi Wang, Yida Xue, Ningyu Zhang, Xiaoyan Yang, Qiang Li, Yue Shen, Lei Liang, Jinjie Gu, and Huajun Chen.
\newblock Unified hallucination detection for multimodal large language models.
\newblock {\em arXiv preprint arXiv:2402.03190}, 2024.

\bibitem{cheng2024emotion}
Zebang Cheng, Zhi-Qi Cheng, Jun-Yan He, Kai Wang, Yuxiang Lin, Zheng Lian, Xiaojiang Peng, and Alexander Hauptmann.
\newblock Emotion-llama: Multimodal emotion recognition and reasoning with instruction tuning.
\newblock {\em Advances in Neural Information Processing Systems}, 37:110805--110853, 2024.

\bibitem{chu2024qwen2}
Yunfei Chu, Jin Xu, Qian Yang, Haojie Wei, Xipin Wei, Zhifang Guo, Yichong Leng, Yuanjun Lv, Jinzheng He, Junyang Lin, et~al.
\newblock Qwen2-audio technical report.
\newblock {\em arXiv preprint arXiv:2407.10759}, 2024.

\bibitem{cui2023holistic}
Chenhang Cui, Yiyang Zhou, Xinyu Yang, Shirley Wu, Linjun Zhang, James Zou, and Huaxiu Yao.
\newblock Holistic analysis of hallucination in gpt-4v (ision): Bias and interference challenges.
\newblock {\em arXiv preprint arXiv:2311.03287}, 2023.

\bibitem{deng2023soul}
Yue Deng, Wenxuan Zhang, Sinno~Jialin Pan, and Lidong Bing.
\newblock Soul: Towards sentiment and opinion understanding of language.
\newblock {\em arXiv preprint arXiv:2310.17924}, 2023.

\bibitem{ding2025kimi}
Ding Ding, Zeqian Ju, Yichong Leng, Songxiang Liu, Tong Liu, Zeyu Shang, Kai Shen, Wei Song, Xu~Tan, Heyi Tang, et~al.
\newblock Kimi-audio technical report.
\newblock {\em arXiv preprint arXiv:2504.18425}, 2025.

\bibitem{dzedzickis2020human}
Andrius Dzedzickis, Art{\=u}ras Kaklauskas, and Vytautas Bucinskas.
\newblock Human emotion recognition: Review of sensors and methods.
\newblock {\em Sensors}, 20(3):592, 2020.

\bibitem{el2011survey}
Moataz El~Ayadi, Mohamed~S Kamel, and Fakhri Karray.
\newblock Survey on speech emotion recognition: Features, classification schemes, and databases.
\newblock {\em Pattern Recognition}, 44(3):572--587, 2011.

\bibitem{ezzameli2023emotion}
K~Ezzameli and H~Mahersia.
\newblock Emotion recognition from unimodal to multimodal analysis: A review.
\newblock {\em Information Fusion}, page 101847, 2023.

\bibitem{filippova2020controlled}
Katja Filippova.
\newblock Controlled hallucinations: Learning to generate faithfully from noisy data.
\newblock {\em arXiv preprint arXiv:2010.05873}, 2020.

\bibitem{gemini2025update}
{Google DeepMind}.
\newblock Gemini: Our most capable model, getting better with thinking and coding.
\newblock \url{https://blog.google/technology/google-deepmind/gemini-model-thinking-updates-march-2025/}, March 2025.

\bibitem{grattafiori2024llama}
Aaron Grattafiori, Abhimanyu Dubey, Abhinav Jauhri, Abhinav Pandey, Abhishek Kadian, Ahmad Al-Dahle, Aiesha Letman, Akhil Mathur, Alan Schelten, Alex Vaughan, et~al.
\newblock The llama 3 herd of models.
\newblock {\em arXiv preprint arXiv:2407.21783}, 2024.

\bibitem{guan2024hallusionbench}
Tianrui Guan, Fuxiao Liu, Xiyang Wu, Ruiqi Xian, Zongxia Li, Xiaoyu Liu, Xijun Wang, Lichang Chen, Furong Huang, Yaser Yacoob, et~al.
\newblock Hallusionbench: an advanced diagnostic suite for entangled language hallucination and visual illusion in large vision-language models.
\newblock In {\em Proceedings of the IEEE/CVF Conference on Computer Vision and Pattern Recognition}, pages 14375--14385, 2024.

\bibitem{guo2025deepseek}
Daya Guo, Dejian Yang, Haowei Zhang, Junxiao Song, Ruoyu Zhang, Runxin Xu, Qihao Zhu, Shirong Ma, Peiyi Wang, Xiao Bi, et~al.
\newblock Deepseek-r1: Incentivizing reasoning capability in llms via reinforcement learning.
\newblock {\em arXiv preprint arXiv:2501.12948}, 2025.

\bibitem{hakak2017emotion}
Nida~Manzoor Hakak, Mohsin Mohd, Mahira Kirmani, and Mudasir Mohd.
\newblock Emotion analysis: A survey.
\newblock In {\em International Conference on Computer, Communications and Electronics (COMPTELIX)}, pages 397--402. IEEE, 2017.

\bibitem{han2024onellm}
Jiaming Han, Kaixiong Gong, Yiyuan Zhang, Jiaqi Wang, Kaipeng Zhang, Dahua Lin, Yu~Qiao, Peng Gao, and Xiangyu Yue.
\newblock Onellm: One framework to align all modalities with language.
\newblock In {\em Proceedings of the IEEE/CVF Conference on Computer Vision and Pattern Recognition}, pages 26584--26595, 2024.

\bibitem{hareli2015cross}
Shlomo Hareli, Konstantinos Kafetsios, and Ursula Hess.
\newblock A cross-cultural study on emotion expression and the learning of social norms.
\newblock {\em Frontiers in Psychology}, 6:1501, 2015.

\bibitem{hurst2024GPT}
Aaron Hurst, Adam Lerer, Adam~P Goucher, Adam Perelman, Aditya Ramesh, Aidan Clark, AJ~Ostrow, Akila Welihinda, Alan Hayes, Alec Radford, et~al.
\newblock Gpt-4o system card.
\newblock {\em arXiv preprint arXiv:2410.21276}, 2024.

\bibitem{ji2023survey}
Ziwei Ji, Nayeon Lee, Rita Frieske, Tiezheng Yu, Dan Su, Yan Xu, Etsuko Ishii, Ye~Jin Bang, Andrea Madotto, and Pascale Fung.
\newblock Survey of hallucination in natural language generation.
\newblock {\em ACM Computing Surveys}, 55(12):1--38, 2023.

\bibitem{jiang2023mistral7b}
Albert~Q. Jiang, Alexandre Sablayrolles, Arthur Mensch, Chris Bamford, Devendra~Singh Chaplot, Diego de~las Casas, Florian Bressand, Gianna Lengyel, Guillaume Lample, Lucile Saulnier, Lélio~Renard Lavaud, Marie-Anne Lachaux, Pierre Stock, Teven~Le Scao, Thibaut Lavril, Thomas Wang, Timothée Lacroix, and William~El Sayed.
\newblock Mistral 7b.
\newblock 2023.

\bibitem{jiang2024mixtral}
Albert~Q Jiang, Alexandre Sablayrolles, Antoine Roux, Arthur Mensch, Blanche Savary, Chris Bamford, Devendra~Singh Chaplot, Diego de~las Casas, Emma~Bou Hanna, Florian Bressand, et~al.
\newblock Mixtral of experts.
\newblock {\em arXiv preprint arXiv:2401.04088}, 2024.

\bibitem{jin2024chat}
Peng Jin, Ryuichi Takanobu, Wancai Zhang, Xiaochun Cao, and Li~Yuan.
\newblock Chat-univi: Unified visual representation empowers large language models with image and video understanding.
\newblock In {\em Proceedings of the IEEE/CVF Conference on Computer Vision and Pattern Recognition}, pages 13700--13710, 2024.

\bibitem{kagitcibasi1997individualism}
Cigdem Kagitcibasi.
\newblock Individualism and collectivism.
\newblock {\em Handbook of Cross-cultural Psychology}, 3:1--49, 1997.

\bibitem{koelstra2011deap}
Sander Koelstra, Christian Muhl, Mohammad Soleymani, Jong-Seok Lee, Ashkan Yazdani, Touradj Ebrahimi, Thierry Pun, Anton Nijholt, and Ioannis Patras.
\newblock Deap: A database for emotion analysis; using physiological signals.
\newblock {\em IEEE Transactions on Affective Computing}, 3(1):18--31, 2011.

\bibitem{lee2010text}
Sophia Yat~Mei Lee, Ying Chen, and Chu-Ren Huang.
\newblock A text-driven rule-based system for emotion cause detection.
\newblock In {\em Proceedings of the NAACL HLT Workshop on Computational Approaches to Analysis and Generation of Emotion in Text}, pages 45--53, 2010.

\bibitem{lewis2005bridging}
Marc~D Lewis.
\newblock Bridging emotion theory and neurobiology through dynamic systems modeling.
\newblock {\em Behavioral and Brain Sciences}, 28(2):169--194, 2005.

\bibitem{li2024eald}
Deng Li, Xin Liu, Bohao Xing, Baiqiang Xia, Yuan Zong, Bihan Wen, and Heikki K{\"a}lvi{\"a}inen.
\newblock Eald-mllm: Emotion analysis in long-sequential and de-identity videos with multi-modal large language model.
\newblock {\em arXiv preprint arXiv:2405.00574}, 2024.

\bibitem{li2025deemo}
Deng Li, Bohao Xing, Xin Liu, Baiqiang Xia, Bihan Wen, and Heikki K{\"a}lvi{\"a}inen.
\newblock Deemo: De-identity multimodal emotion recognition and reasoning.
\newblock {\em arXiv preprint arXiv:2504.19549}, 2025.

\bibitem{li2023blip}
Junnan Li, Dongxu Li, Silvio Savarese, and Steven Hoi.
\newblock Blip-2: Bootstrapping language-image pre-training with frozen image encoders and large language models.
\newblock In {\em International Conference on Machine Learning}, pages 19730--19742. PMLR, 2023.

\bibitem{li2020deep}
Shan Li and Weihong Deng.
\newblock Deep facial expression recognition: A survey.
\newblock {\em IEEE Transactions on Affective Computing}, 13(3):1195--1215, 2020.

\bibitem{li2024llama}
Yanwei Li, Chengyao Wang, and Jiaya Jia.
\newblock Llama-vid: An image is worth 2 tokens in large language models.
\newblock In {\em European Conference on Computer Vision}, pages 323--340. Springer, 2024.

\bibitem{li2023evaluating}
Yifan Li, Yifan Du, Kun Zhou, Jinpeng Wang, Wayne~Xin Zhao, and Ji-Rong Wen.
\newblock Evaluating object hallucination in large vision-language models.
\newblock {\em arXiv preprint arXiv:2305.10355}, 2023.

\bibitem{lian2025affectGPT}
Zheng Lian, Haoyu Chen, Lan Chen, Haiyang Sun, Licai Sun, Yong Ren, Zebang Cheng, Bin Liu, Rui Liu, Xiaojiang Peng, et~al.
\newblock Affectgpt: A new dataset, model, and benchmark for emotion understanding with multimodal large language models.
\newblock {\em arXiv preprint arXiv:2501.16566}, 2025.

\bibitem{lian2023mer}
Zheng Lian, Haiyang Sun, Licai Sun, Kang Chen, Mngyu Xu, Kexin Wang, Ke~Xu, Yu~He, Ying Li, Jinming Zhao, et~al.
\newblock Mer 2023: Multi-label learning, modality robustness, and semi-supervised learning.
\newblock In {\em Proceedings of the ACM International Conference on Multimedia}, pages 9610--9614, 2023.

\bibitem{lian2024open}
Zheng Lian, Haiyang Sun, Licai Sun, Lan Chen, Haoyu Chen, Hao Gu, Zhuofan Wen, Shun Chen, Siyuan Zhang, Hailiang Yao, et~al.
\newblock Open-vocabulary multimodal emotion recognition: Dataset, metric, and benchmark.
\newblock {\em arXiv preprint arXiv:2410.01495}, 2024.

\bibitem{lian2023explainable}
Zheng Lian, Haiyang Sun, Licai Sun, Hao Gu, Zhuofan Wen, Siyuan Zhang, Shun Chen, Mingyu Xu, Ke~Xu, Kang Chen, et~al.
\newblock Explainable multimodal emotion recognition.
\newblock {\em arXiv preprint arXiv:2306.15401}, 2023.

\bibitem{lin2023video}
Bin Lin, Yang Ye, Bin Zhu, Jiaxi Cui, Munan Ning, Peng Jin, and Li~Yuan.
\newblock Video-llava: Learning united visual representation by alignment before projection.
\newblock {\em arXiv preprint arXiv:2311.10122}, 2023.

\bibitem{lin2004rouge}
Chin-Yew Lin.
\newblock Rouge: A package for automatic evaluation of summaries.
\newblock In {\em Text Summarization Branches out}, pages 74--81, 2004.

\bibitem{liu2024deepseek}
Aixin Liu, Bei Feng, Bing Xue, Bingxuan Wang, Bochao Wu, Chengda Lu, Chenggang Zhao, Chengqi Deng, Chenyu Zhang, Chong Ruan, et~al.
\newblock Deepseek-v3 technical report.
\newblock {\em arXiv preprint arXiv:2412.19437}, 2024.

\bibitem{liu2023visual}
Haotian Liu, Chunyuan Li, Qingyang Wu, and Yong~Jae Lee.
\newblock Visual instruction tuning.
\newblock {\em Advances in Neural Information Processing Systems}, 36:34892--34916, 2023.

\bibitem{liu2024phd}
Jiazhen Liu, Yuhan Fu, Ruobing Xie, Runquan Xie, Xingwu Sun, Fengzong Lian, Zhanhui Kang, and Xirong Li.
\newblock Phd: A prompted visual hallucination evaluation dataset.
\newblock {\em arXiv e-prints}, pages arXiv--2403, 2024.

\bibitem{liu2021imigue}
Xin Liu, Henglin Shi, Haoyu Chen, Zitong Yu, Xiaobai Li, and Guoying Zhao.
\newblock imigue: An identity-free video dataset for micro-gesture understanding and emotion analysis.
\newblock In {\em Proceedings of the IEEE/CVF Conference on Computer Vision and Pattern Recognition}, pages 10631--10642, 2021.

\bibitem{livingstone2018ryerson}
Steven~R Livingstone and Frank~A Russo.
\newblock The ryerson audio-visual database of emotional speech and song (ravdess): A dynamic, multimodal set of facial and vocal expressions in north american english.
\newblock {\em PloS One}, 13(5):e0196391, 2018.

\bibitem{maaz2023video}
Muhammad Maaz, Hanoona Rasheed, Salman Khan, and Fahad~Shahbaz Khan.
\newblock Video-chatgpt: Towards detailed video understanding via large vision and language models.
\newblock {\em arXiv preprint arXiv:2306.05424}, 2023.

\bibitem{maynez2020faithfulness}
Joshua Maynez, Shashi Narayan, Bernd Bohnet, and Ryan McDonald.
\newblock On faithfulness and factuality in abstractive summarization.
\newblock {\em arXiv preprint arXiv:2005.00661}, 2020.

\bibitem{mistral2024small31}
{Mistral AI}.
\newblock Introducing mistral-3.1: our new small model.
\newblock \url{https://mistral.ai/news/mistral-small-3-1}, 2024.

\bibitem{nandwani2021review}
Pansy Nandwani and Rupali Verma.
\newblock A review on sentiment analysis and emotion detection from text.
\newblock {\em Social Network Analysis and Mining}, 11(1):81, 2021.

\bibitem{niedenthal2017psychology}
Paula~M Niedenthal and Fran{\c{c}}ois Ric.
\newblock {\em Psychology of Emotion}.
\newblock Psychology Press, 2017.

\bibitem{openai2024GPT4}
{OpenAI}.
\newblock Gpt-4 turbo and the next frontier of intelligence.
\newblock \url{https://openai.com/index/gpt-4-1/}, November 2024.
\newblock Accessed: 2025-04-30.

\bibitem{papineni2002bleu}
Kishore Papineni, Salim Roukos, Todd Ward, and Wei-Jing Zhu.
\newblock Bleu: a method for automatic evaluation of machine translation.
\newblock In {\em Proceedings of the Annual Meeting of the Association for Computational Linguistics}, pages 311--318, 2002.

\bibitem{perez2022red}
Ethan Perez, Saffron Huang, Francis Song, Trevor Cai, Roman Ring, John Aslanides, Amelia Glaese, Nat McAleese, and Geoffrey Irving.
\newblock Red teaming language models with language models.
\newblock {\em arXiv preprint arXiv:2202.03286}, 2022.

\bibitem{petryk2024aloha}
Suzanne Petryk, David~M Chan, Anish Kachinthaya, Haodi Zou, John Canny, Joseph~E Gonzalez, and Trevor Darrell.
\newblock Aloha: A new measure for hallucination in captioning models.
\newblock {\em arXiv preprint arXiv:2404.02904}, 2024.

\bibitem{qwen2024qwen3}
{Qwen Team}.
\newblock Qwen 3: A family of scalable language models.
\newblock \url{https://qwenlm.github.io/blog/qwen3/}, 2024.

\bibitem{qwen2025QvQ72b}
{Qwen Team}.
\newblock Preview of qwen-vl-72b: Advancing multimodal understanding.
\newblock \url{https://qwenlm.github.io/blog/qvq-72b-preview/}, 2025.
\newblock Accessed: 2025-04-30.

\bibitem{rahdari2019multimodal}
Farhad Rahdari, Esmat Rashedi, and Mahdi Eftekhari.
\newblock A multimodal emotion recognition system using facial landmark analysis.
\newblock {\em Iranian Journal of Science and Technology, Transactions of Electrical Engineering}, 43:171--189, 2019.

\bibitem{rohrbach2018object}
Anna Rohrbach, Lisa~Anne Hendricks, Kaylee Burns, Trevor Darrell, and Kate Saenko.
\newblock Object hallucination in image captioning.
\newblock {\em arXiv preprint arXiv:1809.02156}, 2018.

\bibitem{russell1980circumplex}
James~A Russell.
\newblock A circumplex model of affect.
\newblock {\em Journal of Personality and Social Psychology}, 39(6):1161, 1980.

\bibitem{scherer2009dynamic}
Klaus~R Scherer.
\newblock The dynamic architecture of emotion: Evidence for the component process model.
\newblock {\em Cognition and Emotion}, 23(7):1307--1351, 2009.

\bibitem{shiota2017emotion}
Michelle~N. Shiota and James~W. Kalat.
\newblock {\em Emotion}.
\newblock Oxford University Press, New York, NY, 3 edition, 2017.

\bibitem{smith1989dimensions}
Craig~A Smith.
\newblock Dimensions of appraisal and physiological response in emotion.
\newblock {\em Journal of personality and social psychology}, 56(3):339, 1989.

\bibitem{sun2023aligning}
Zhiqing Sun, Sheng Shen, Shengcao Cao, Haotian Liu, Chunyuan Li, Yikang Shen, Chuang Gan, Liang-Yan Gui, Yu-Xiong Wang, Yiming Yang, et~al.
\newblock Aligning large multimodal models with factually augmented rlhf.
\newblock {\em arXiv preprint arXiv:2309.14525}, 2023.

\bibitem{team2025gemma}
Gemma Team, Aishwarya Kamath, Johan Ferret, Shreya Pathak, Nino Vieillard, Ramona Merhej, Sarah Perrin, Tatiana Matejovicova, Alexandre Ram{\'e}, Morgane Rivi{\`e}re, et~al.
\newblock Gemma 3 technical report.
\newblock {\em arXiv preprint arXiv:2503.19786}, 2025.

\bibitem{QwQ32b}
Qwen Team.
\newblock Qwq-32b: Embracing the power of reinforcement learning, March 2025.

\bibitem{thomson2020gold}
Craig Thomson and Ehud Reiter.
\newblock A gold standard methodology for evaluating accuracy in data-to-text systems.
\newblock {\em arXiv preprint arXiv:2011.03992}, 2020.

\bibitem{tong2024eyes}
Shengbang Tong, Zhuang Liu, Yuexiang Zhai, Yi~Ma, Yann LeCun, and Saining Xie.
\newblock Eyes wide shut? exploring the visual shortcomings of multimodal llms.
\newblock In {\em Proceedings of the IEEE/CVF Conference on Computer Vision and Pattern Recognition}, pages 9568--9578, 2024.

\bibitem{wang2023amber}
Junyang Wang, Yuhang Wang, Guohai Xu, Jing Zhang, Yukai Gu, Haitao Jia, Jiaqi Wang, Haiyang Xu, Ming Yan, Ji~Zhang, et~al.
\newblock Amber: An llm-free multi-dimensional benchmark for mllms hallucination evaluation.
\newblock {\em arXiv preprint arXiv:2311.07397}, 2023.

\bibitem{wang2024videohallucer}
Yuxuan Wang, Yueqian Wang, Dongyan Zhao, Cihang Xie, and Zilong Zheng.
\newblock Videohallucer: Evaluating intrinsic and extrinsic hallucinations in large video-language models.
\newblock {\em arXiv preprint arXiv:2406.16338}, 2024.

\bibitem{wani2021comprehensive}
Taiba~Majid Wani, Teddy~Surya Gunawan, Syed Asif~Ahmad Qadri, Mira Kartiwi, and Eliathamby Ambikairajah.
\newblock A comprehensive review of speech emotion recognition systems.
\newblock {\em IEEE Access}, 9:47795--47814, 2021.

\bibitem{wankhade2022survey}
Mayur Wankhade, Annavarapu Chandra~Sekhara Rao, and Chaitanya Kulkarni.
\newblock A survey on sentiment analysis methods, applications, and challenges.
\newblock {\em Artificial Intelligence Review}, 55(7):5731--5780, 2022.

\bibitem{socialiq2github}
Adam~B. Wilf and collaborators.
\newblock Social-iq 2.0 challenge.
\newblock \url{https://github.com/abwilf/Social-IQ-2.0-Challenge}, 2023.

\bibitem{xing2024emo}
Bohao Xing, Zitong Yu, Xin Liu, Kaishen Yuan, Qilang Ye, Weicheng Xie, Huanjing Yue, Jingyu Yang, and Heikki K{\"a}lvi{\"a}inen.
\newblock Emo-llama: Enhancing facial emotion understanding with instruction tuning.
\newblock {\em arXiv preprint arXiv:2408.11424}, 2024.

\bibitem{xu2025qwen2}
Jin Xu, Zhifang Guo, Jinzheng He, Hangrui Hu, Ting He, Shuai Bai, Keqin Chen, Jialin Wang, Yang Fan, Kai Dang, et~al.
\newblock Qwen2. 5-omni technical report.
\newblock {\em arXiv preprint arXiv:2503.20215}, 2025.

\bibitem{yang2024qwen2technicalreport}
An~Yang, Baosong Yang, Binyuan Hui, Bo~Zheng, Bowen Yu, Chang Zhou, Chengpeng Li, Chengyuan Li, Dayiheng Liu, Fei Huang, Guanting Dong, Haoran Wei, Huan Lin, Jialong Tang, Jialin Wang, Jian Yang, Jianhong Tu, Jianwei Zhang, Jianxin Ma, Jianxin Yang, Jin Xu, Jingren Zhou, Jinze Bai, Jinzheng He, Junyang Lin, Kai Dang, Keming Lu, Keqin Chen, Kexin Yang, Mei Li, Mingfeng Xue, Na~Ni, Pei Zhang, Peng Wang, Ru~Peng, Rui Men, Ruize Gao, Runji Lin, Shijie Wang, Shuai Bai, Sinan Tan, Tianhang Zhu, Tianhao Li, Tianyu Liu, Wenbin Ge, Xiaodong Deng, Xiaohuan Zhou, Xingzhang Ren, Xinyu Zhang, Xipin Wei, Xuancheng Ren, Xuejing Liu, Yang Fan, Yang Yao, Yichang Zhang, Yu~Wan, Yunfei Chu, Yuqiong Liu, Zeyu Cui, Zhenru Zhang, Zhifang Guo, and Zhihao Fan.
\newblock Qwen2 technical report.
\newblock 2024.

\bibitem{yang2024qwen2}
An~Yang, Baosong Yang, Beichen Zhang, Binyuan Hui, Bo~Zheng, Bowen Yu, Chengyuan Li, Dayiheng Liu, Fei Huang, Haoran Wei, et~al.
\newblock Qwen2. 5 technical report.
\newblock {\em arXiv preprint arXiv:2412.15115}, 2024.

\bibitem{yang2025omni}
Qize Yang, Detao Bai, Yi-Xing Peng, and Xihan Wei.
\newblock Omni-emotion: Extending video mllm with detailed face and audio modeling for multimodal emotion analysis.
\newblock {\em arXiv preprint arXiv:2501.09502}, 2025.

\bibitem{yin2024woodpecker}
Shukang Yin, Chaoyou Fu, Sirui Zhao, Tong Xu, Hao Wang, Dianbo Sui, Yunhang Shen, Ke~Li, Xing Sun, and Enhong Chen.
\newblock Woodpecker: Hallucination correction for multimodal large language models.
\newblock {\em Science China Information Sciences}, 67(12):220105, 2024.

\bibitem{zadeh2019social}
Amir Zadeh, Michael Chan, Paul~Pu Liang, Edmund Tong, and Louis-Philippe Morency.
\newblock Social-iq: A question answering benchmark for artificial social intelligence.
\newblock In {\em Proceedings of the IEEE/CVF Conference on Computer Vision and Pattern Recognition}, pages 8807--8817, 2019.

\bibitem{zadeh2017tensor}
Amir Zadeh, Minghai Chen, Soujanya Poria, Erik Cambria, and Louis-Philippe Morency.
\newblock Tensor fusion network for multimodal sentiment analysis.
\newblock {\em arXiv preprint arXiv:1707.07250}, 2017.

\bibitem{zaki2012neuroscience}
Jamil Zaki and Kevin~N Ochsner.
\newblock The neuroscience of empathy: progress, pitfalls and promise.
\newblock {\em Nature Neuroscience}, 15(5):675--680, 2012.

\bibitem{zeidner2003development}
Moshe Zeidner, Gerald Matthews, Richard~D Roberts, and Carolyn MacCann.
\newblock Development of emotional intelligence: Towards a multi-level investment model.
\newblock {\em Human Development}, 46(2-3):69--96, 2003.

\bibitem{zhang2024eventhallusion}
Jiacheng Zhang, Yang Jiao, Shaoxiang Chen, Na~Zhao, Zhiyu Tan, Hao Li, and Jingjing Chen.
\newblock Eventhallusion: Diagnosing event hallucinations in video llms.
\newblock {\em arXiv preprint arXiv:2409.16597}, 2024.

\bibitem{zhang2023language}
Muru Zhang, Ofir Press, William Merrill, Alisa Liu, and Noah~A Smith.
\newblock How language model hallucinations can snowball.
\newblock {\em arXiv preprint arXiv:2305.13534}, 2023.

\bibitem{zhang2018adaptive}
Qi~Zhang, Jinlan Fu, Xiaoyu Liu, and Xuanjing Huang.
\newblock Adaptive co-attention network for named entity recognition in tweets.
\newblock In {\em Proceedings of the AAAI Conference on Artificial Intelligence}, volume~32, 2018.

\bibitem{zhang2023dialoguellm}
Yazhou Zhang, Mengyao Wang, Youxi Wu, Prayag Tiwari, Qiuchi Li, Benyou Wang, and Jing Qin.
\newblock Dialoguellm: Context and emotion knowledge-tuned large language models for emotion recognition in conversations.
\newblock {\em arXiv preprint arXiv:2310.11374}, 2023.

\bibitem{zhao2021affective}
Sicheng Zhao, Xingxu Yao, Jufeng Yang, Guoli Jia, Guiguang Ding, Tat-Seng Chua, Bjoern~W Schuller, and Kurt Keutzer.
\newblock Affective image content analysis: Two decades review and new perspectives.
\newblock {\em IEEE Transactions on Pattern Analysis and Machine Intelligence}, 44(10):6729--6751, 2021.

\bibitem{zhao2021former}
Zengqun Zhao and Qingshan Liu.
\newblock Former-dfer: Dynamic facial expression recognition transformer.
\newblock In {\em Proceedings of the ACM International Conference on Multimedia}, pages 1553--1561, 2021.

\end{thebibliography}


\newpage
\appendix

\section*{APPENDIX}  

To facilitate deeper understanding and reproducibility, we organize the appendix as follows:
\begin{itemize}
\item \textbf{Appendix}~\ref{app:anno} presents additional information on benchmark construction and annotation procedures.
\item \textbf{Appendix}~\ref{app:implementation} details the implementation of our evaluation setup and the PEP-MEK.
\item \textbf{Appendix}~\ref{app:experiment} includes additional experimental results along with in-depth analysis.
\item \textbf{Appendix}~\ref{app:examples} provides more examples.
\item \textbf{Appendix}~\ref{app:limitation} discusses the limitations of our work and directions for future improvement.
\end{itemize}

\section{Collection and Annotation Details}\label{app:anno}
\begin{figure}[h]
    \centering
    \includegraphics[width=\linewidth]{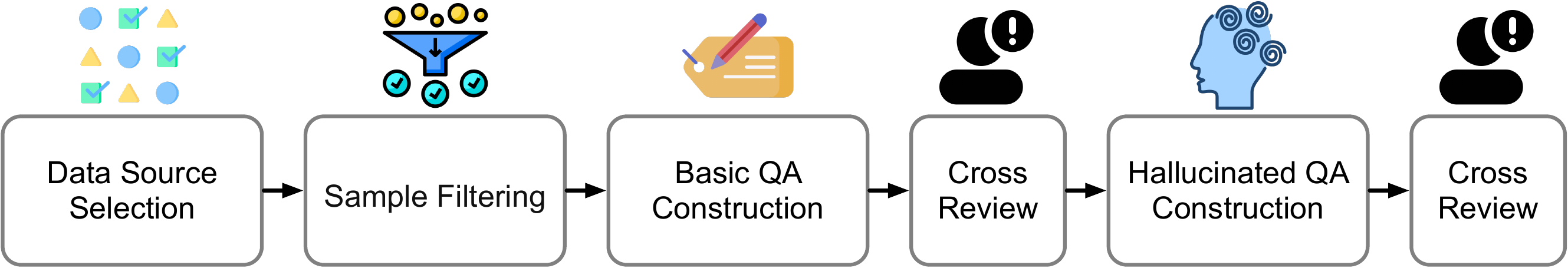}
    \caption{Overview of the annotation workflow.
The process consists of four key stages: selecting relevant data sources, filtering for appropriate samples, constructing basic QA pairs, and generating hallucinated QA variants by applying diverse hallucination strategies. Each QA construction step is followed by a cross-review phase to ensure quality and consistency across annotators. This unified pipeline is applied across all modalities, including text, image, audio, and video.}
    \label{fig:anno_pipeline}
\end{figure}

Here, we provide annotation details to better illustrate our data sources as well as each modality and category. 
The process begins with the collection of knowledge from emotion psychology and existing emotion understanding datasets. 
This foundational step ensures that our annotation is grounded in well-established emotional concepts and tasks.
Next, we engage a group of trained annotators to carefully filter and review the initial QA. Their task is to ensure that each QA pair is accurate, coherent, and well-grounded in the corresponding source content.
Following this, for each basic question, annotators are instructed to generate a corresponding hallucinated question.
To ensure high annotation quality, we adopt a cross-review verification mechanism. Each question (basic and hallucinated) is independently reviewed by a second annotator. Only when both annotators reach consensus is the item included in the final benchmark. 
The overall annotation workflow is shown in Figure~\ref{fig:anno_pipeline}.

\subsection{Collection and Annotation Across Modalities}
We first present our data collection and annotation procedures from a modality perspective, to provide a clearer understanding of our data sources and the original objectives of each dataset.

\subsubsection{Text} 
To ground our annotation process in established affective science, we selected a leading textbook in the field of emotion psychology: Emotion (3rd Edition) by Michelle N. Shiota and James W. Kalat~\cite{shiota2017emotion}. 
This book offers a comprehensive overview of key emotional theories~\cite{cannon1927james,scherer2009dynamic,lewis2005bridging}, physiological and cognitive underpinnings of emotions~\cite{berkowitz1999anger}, as well as cross-cultural perspectives~\cite{kagitcibasi1997individualism,hareli2015cross,best2013cost}. 
It is widely used in academic settings and is recognized for its balanced integration of classical theories (e.g., James-Lange~\cite{cannon1927james}, Schachter-Singer) and contemporary research findings. 
The text also emphasizes real-world applications and experimental paradigms~\cite{scherer2009dynamic,lewis2005bridging}, making it a suitable foundation for constructing psychologically grounded emotion-related tasks. 
We collected a subset of statements from the book that conveyed clear emotion psychology viewpoints and insights. These statements were carefully selected through manual review to ensure their clarity and relevance. For each selected statement, the basic question was taken directly from the original text, preserving the emotional concept or psychological mechanism as presented in the source. 
After the basic QA pair was established, annotators were instructed to generate a hallucinated version. 
Common strategies included addition, deletion, or distortion of key concepts or the intension and extension of a given inference, while ensuring the question remained grammatically fluent and superficially plausible. 
As a result, we get 784 basic and hallucination QA pairs.

In the real-world text modality emotion understanding, we incorporate existing datasets to better examine emotion perception hallucinations in real-world textual contexts. 
One such dataset is SOUL~\cite{deng2023soul}, which is specifically designed to evaluate fine-grained understanding of sentiment, stance, and opinion in natural language. 
In its original task format, SOUL presents a review text alongside multiple statements that reflect potential interpretations of the opinion expressed. Each statement is annotated as either correct or incorrect, depending on its emotional alignment with respect to the original review. This setup goes beyond traditional sentiment analysis by emphasizing deeper dimensions of subjective understanding, such as emotional appropriateness, implicit attitude, and contextual nuance. 
In our annotation process, for each review text, we collected all associated statements and manually refined them into a single, well-grounded summary statement. This statement was used as the basic question in our QA format. 
This approach allowed our QA pairs to inherit SOUL’s emphasis on deeper dimensions of subjective understanding, thus enabling more fine-grained evaluation of MLLMs’ capabilities in emotion understanding and hallucination detection.
As a result, we get 200 basic and hallucination QA pairs.

\subsubsection{Image}
We utilize the Twitter15 and Twitter17 datasets~\cite{zhang2018adaptive}, originally developed for named entity recognition and sentiment analysis in social media contexts. 
These datasets consist of tweets accompanied by images and structured annotations, including entity labels and overall tweet-level sentiment polarity (positive, negative, neutral). The data reflects real-world, noisy, and multimodal content, making it a valuable resource for emotion perception tasks.

In our annotation process, we focused on images. Notably, we observed that many images contain multiple people, often displaying divergent emotional expressions within the same instance, as shown in Figure~\ref{fig:PEP_e1}. 
This introduces a unique challenge for MLLMs, as it requires fine grained emotion understanding and the ability to disambiguate multi-actor sentiment based on visual cues. To reflect this, we reformulated each image into a QA format, selecting or synthesizing emotionally consistent description as basic questions, and constructing hallucinated variants by introducing emotionally incongruent elements, such as misattributing emotions to the wrong person or exaggerating emotional intensity.
As a result, we get 300 basic and hallucination QA pairs and 150 images.

\subsubsection{Audio}

We incorporate the RAVDESS dataset~\cite{livingstone2018ryerson}, which provides a multimodal collection of emotional expressions through both facial and vocal modalities. 
The dataset consists of high-quality recordings in North American English, featuring actors performing scripted speech and song with controlled emotional expressions. Emotions span a wide range (including calm, happy, sad, angry, fearful) and more, and are further labeled with varying levels of intensity, making RAVDESS a rich resource for studying graded emotional cues.

In our annotation process, we used only the audio modality as input, focusing on how emotion is conveyed through vocal tone and prosody rather than linguistic content. 
For each selected audio clip, annotators generated a basic question that captured the emotional tone perceived from the speech alone, based on the original emotion labels provided in the dataset. Hallucinated variants were then constructed by subtly misrepresenting the emotional valence or intensity, such as exaggerating, downplaying, or inverting the expressed emotion, allowing us to assess the sensitivity of MLLMs to audio signals without relying on semantic content.
As a result, we get 736 basic and hallucination QA pairs and 368 audios.

\subsubsection{Video}
Videos contain both visual and auditory information, they represent the most natural and comprehensive modality for capturing human emotional experiences. 
Motivated by this, we selected two data sources, MER 2023\cite{lian2023mer} and Social-IQ 2.0\cite{zadeh2019social,socialiq2github}, to evaluate emotion understanding hallucination in realistic, multimodal contexts. 
Both datasets involve video-based social situations, where emotion understanding relies not only on speech and text, but also on facial expressions, gestures, and interpersonal dynamics.

MER 2023 focuses on multimodal emotion recognition on sentence-length inputs. Each instance consists of a short video clip accompanied by aligned textual transcripts, from which the model is expected to identify the expressed emotion based on the combined audiovisual and textual cues. It emphasizes how emotions are conveyed through brief, but richly multimodal expressions.

Social-IQ 2.0, in contrast, is designed to assess social intelligence in multimodal contexts, requiring models to understand and reason about complex human interactions. 
Each video clip in the dataset portrays a real-world social scenario sourced from platforms like YouTube, often involving interpersonal communication, emotional exchanges, or goal-directed behavior. Importantly, each video is paired with multiple questions, each targeting a different aspect of the scene, such as the emotional state of a person, their underlying intentions, the appropriateness of their response. It transcends traditional discrete emotion concepts and aligns more closely with the diverse emotional experiences encountered in real-world contexts.
Each question is accompanied by several candidate answers, only one of which is correct, thereby framing the task as a multiple-choice question-answering problem.

In our annotation process, for MER, we first expanded the original emotion label into a basic reasoning process that incorporates both verbal and non-verbal cues. 
This reasoning was then formulated into a basic question that captures the emotion and its underlying justifications. 
Each annotator was subsequently tasked with generating a hallucinated version of the question by subtly altering key aspects of the emotional reasoning, for example, by misattributing the cause of the emotion, exaggerating its intensity, or neglecting relevant multimodal cues.
As a result, we get 360 basic and hallucination QA pairs and 180 videos.

For Social-IQ 2.0, the annotation process was more complex. We preserved the original multiple-question-per-video structure to maintain the richness of the reasoning context. 
First, we filtered out ambiguous or poorly defined questions. 
For the remaining high-quality questions, we refined and expanded each one, along with its correct answer, into a more detailed version, which served as the basic question in our QA format. 
Annotators then created hallucinated variants by introducing subtle distortions in social or emotion understanding, such as misinterpreting intentions, assigning incorrect emotional reactions. This enabled us to evaluate MLLMs’ ability to handle hallucinations in multimodal contexts.
As a result, we get 402 basic and hallucination QA pairs and 50 videos.

\subsection{Annotation Across Categories}

\subsubsection{Theory}

Throughout the development of emotion psychology, various emotional theories have emerged, each reflecting different perspectives on the nature, origin, and structure of emotions~\cite{cannon1927james,scherer2009dynamic,lewis2005bridging,shiota2017emotion}. 
In this setting, we focus on evaluating MLLMs’ capability to detect hallucinations in these theoretical frameworks, rather than judging the correctness or scientific validity of the theories themselves.

To clearly signal the basis of each item, we typically preface the question with phrases such as ``According to [Theory Name] …'' or ``[Theory Name] argues that …'', followed by a concise statement derived from the theory. 
When constructing hallucinated questions, we deliberately manipulate key aspects of the original statement, such as reversing causal relationships, disrupting conceptual sequences, or modifying the intension or extension of key emotional constructs. 
This strategy allows us to assess whether the model can factually understand the theoretical content.

\subsubsection{Definition}

In this setting, we focus on evaluating MLLMs’ capability to detecting hallucinations in commonly accepted definitions from emotion psychology~\cite{berkowitz1999anger}, which can help models better align with human emotion understanding. 
To probe their sensitivity, we typically preface the question with phrases such as ``[Concept] is …'', followed by a definition in psychology.

When constructing hallucinated questions, we deliberately manipulate critical aspects of the original definitions, introducing subtle distortions in the intension (i.e., the core conceptual meaning) or the extension (i.e., the range of phenomena the concept applies to) of emotional constructs.

For example, a widely accepted definition of anxiety might state:
``Anxiety is a general expectation that something bad might happen, without identifying any particular danger.''
To construct a hallucinated version, we might alter it to:
``Anxiety is a general expectation that something good might happen, without identifying any particular danger.''
This version subtly but significantly distorts the core meaning of anxiety by replacing negative anticipation with positive anticipation, which undermines the emotional construct’s essential nature.

Such hallucinated definitions are designed to appear structurally and linguistically plausible, yet semantically incorrect. They serve as a diagnostic tool to assess whether MLLMs can distinguish between valid and flawed emotional concept definitions.

\subsubsection{Finding}

In this setting, we focus on evaluating MLLMs’ capability to detect hallucinations in a set of empirical findings about emotion. Unlike formalized theories, these findings refer to observed regularities or patterns in emotional expression, perception, and regulation, often cross-cultural or physiological in nature, without necessarily forming a comprehensive theoretical framework~\cite{kagitcibasi1997individualism,hareli2015cross,best2013cost,smith1989dimensions}.

When constructing hallucinated questions, we deliberately alter the scope or the core semantics of the original statement. These manipulations may involve reversing cultural norms, misattributing physiological phenomena, or distorting the contextual boundaries of emotion.

For example, consider the empirical statement:
``Japanese consider it inappropriate to show negative emotion to a high-status person.''
A hallucinated version would state:
``Japanese consider it appropriate to show negative emotion to a high-status person.''
While structurally similar, this altered version directly contradicts established cross-cultural observations, thereby testing whether the model can detect such culturally incongruent assertions.

This type of adversarial probing helps reveal how well MLLMs understand fine-grained emotion, particularly in contexts involving cultural norms, physiological responses, or socio-emotional regulation. 
Such questions may be challenging even for humans unfamiliar with the specific context, yet the patterns they rely on often stem from deeply ingrained biological or sociocultural tendencies. 
Therefore, the ability of a model to distinguish between accurate and distorted emotional knowledge can serve as a strong indicator of its deeper emotional competence and alignment with human psychological realities.

\subsubsection{Category}

In this setting, we focus on real-world emotion understanding tasks, among which emotion recognition serves as a fundamental component~\cite{dzedzickis2020human}. 
In current standard setups, this typically includes text-based sentiment analysis tasks, which often rely on coarse sentiment labels such as positive or negative. 
However, we also consider discrete basic emotion categories, such as happiness, anger, and sadness, which are widely adopted in visual and multimodal emotion recognition tasks.
Beyond these categories, we incorporate insights from recent work such as OV-MER~\cite{lian2024open}, which proposes a more fine-grained and naturalistic taxonomy of emotions. 
This framework extends beyond basic emotions by incorporating more nuanced, everyday emotional expressions (e.g., relieved, anxious, content) to better capture the richness of human affective states.

In this context, we construct hallucinated examples by intentionally distorting the original emotion label annotations. Specifically, we substitute the original emotion with a semantically related or contrasting category. These substitutions can be adjacent in meaning (e.g., replacing angry with serious) or diametrically opposed (e.g., replacing angry with happy), depending on the intended degree of challenge.

Such manipulations allow us to probe whether MLLMs can detect inconsistencies in emotional labeling, and whether they possess a sufficiently fine-grained understanding of affective semantics to distinguish between closely related or contrasting emotions.

\subsubsection{Intensity}

In our setting, intensity refers to the strength or magnitude of an emotional state as expressed in natural language. Rather than relying on continuous valence-arousal scales commonly used in traditional affective computing~\cite{russell1980circumplex}, we adopt discrete verbal descriptors (e.g., mild, normal, strong), which align more closely with everyday language and the reasoning capabilities of LLMs. 

In this context, hallucinations occur when the model misjudges emotional intensity, either by exaggerating, underestimating, or otherwise misrepresenting the degree of emotion expressed in the input. 
For instance, changing ``He seemed slightly annoyed by the interruption'' to ``He seemed very annoyed by the interruption.''
This form of distortion challenges the model’s ability to capture nuanced emotional gradients and detect subtle affective cues, which is essential for applications requiring fine-grained emotional understanding, such as empathetic communication, psychological assessment, and emotion-aware content generation.

\subsubsection{Reasoning Result}

In the framework, reasoning result refers to the final emotional interpretation that a model generates after perceiving and integrating multimodal emotional cues such as facial expressions, vocal tone, and textual context. 
Unlike perception-level errors, which involve failures in detecting or extracting relevant signals, reasoning result hallucinations arise when the model correctly identifies input cues but still draws an incorrect or implausible emotional conclusion based on faulty reasoning or inference.

For example, given a scenario where a person has a furrowed brow, a tense voice, and says ``I can’t believe this happened again,'' a correct interpretation might be frustration. A hallucinated reasoning result would be labeling this as excitement or anger, despite all emotional cues being correctly perceived. 
In this case, the failure lies not in perception but in misinterpreting the emotional meaning of those cues.

This distinction emphasizes that accurate recognition of emotional signals is a necessary but insufficient condition for successful emotion understanding. Proper emotional inference requires coherent reasoning over the perceived inputs, making reasoning result hallucinations a key diagnostic for evaluating models’ deeper emotional competence.

\subsubsection{Reasoning Cue}

In addition, we define a reasoning cue hallucination, which targets failures in identifying, attending to, or correctly interpreting the emotional cues necessary for sound emotional reasoning—regardless of whether the final emotion classification is ultimately correct or not. 
In this setting, hallucinations occur when the model overlooks, misinterprets, or fabricates critical multimodal signals, such as failing to detect an angry tone in speech, misreading a sad facial expression as neutral, or drawing unsupported emotional implications from text.

We consider two primary forms of cue hallucination: (1) cases where a specific cue is modified while the final emotional result remains unchanged, thereby testing the model’s robustness to misleading or missing signals; and (2) cases where multiple key cues are altered simultaneously, often accompanied by a shift in the final emotion prediction, which challenges the model’s entire reasoning chain, from perception to conclusion.

This setting highlights the importance of the reasoning pathway itself, rather than just the final output. A model may occasionally reach the correct emotion label by chance or superficial pattern matching, but reasoning cue hallucination reveals whether the underlying process is semantically justified and grounded in valid emotional evidence. As such, this setting is crucial for assessing models’ interpretability and trustworthiness in real-world emotion understanding scenarios.

\section{Implementation and Experiment Details}\label{app:implementation}

\subsection{Setups for Baselines}

In our experiment, we selected several well-known LLMs and MLLMs for comparison, as shown in Table~\ref{tab:mllm_comparison}. 
We provide a list of model names, model sizes, supported input and output modalities, all organized by release date to enable a more detailed analysis. Additionally, we annotate whether each model is a reasoning model like DeepSeek-R1~\cite{guo2025deepseek}.
To ensure a fair comparison, we adopt the default hyper-parameters of these models. All models with fewer than 235B parameters were run locally on either a single NVIDIA A100 or four NVIDIA A100 GPUs. The remaining models were accessed via APIs provided by their developers.

\begin{table}[t]
\footnotesize
\centering
\caption{Comparison of representative MLLMs. For closed-source models, we report the API version date. I, V, A, and T represent Image, Video, Audio, and Text, respectively.}
\label{tab:mllm_comparison}
\begin{tabular}{lcccccc}
\toprule
\textbf{Model} & \textbf{Date} & \textbf{Model Size (B)} & \textbf{Input} & \textbf{Output} & \textbf{Reasoning} \\
\hline
\multicolumn{6}{c}{\textit{open-source}} \\
LLaVA~\cite{liu2023visual} & 04.2023 & 7\&13\&34 & I + T & T & No \\
Video-ChatGPT~\cite{maaz2023video} & 06.2023 & 7 & V + T & T & No \\
Mistral~\cite{jiang2023mistral7b} & 08.2023 & 7  & T & T & No \\
Qwen~\cite{bai2023qwen} & 09.2023 & 7\&14 & T & T & No \\
Chat-UniVi~\cite{jin2024chat} & 11.2023 & 7 & I + V + T & T & No \\
LLaMA-VID~\cite{li2024llama} & 11.2023 & 7 & I + V + T & T & No \\
Video-LLaVA~\cite{lin2023video} & 11.2023 & 7 & I + V + T & T & No \\
Onellm~\cite{han2024onellm} & 12.2023 & 7 & I + A + V + T & T & No \\
Mixtral~\cite{jiang2024mixtral} & 01.2024 & 8$\times$7 & T & T & No \\
Llama3~\cite{grattafiori2024llama} & 04.2024 & 8 & T & T & No \\
Emotion-LLaMA~\cite{cheng2024emotion} & 06.2024 & 7 & V + T & T & No \\
Llama3.1~\cite{grattafiori2024llama}   & 07.2024 & 8\&70  & T & T & No \\
Qwen2~\cite{yang2024qwen2technicalreport} & 07.2024 & 7 & Text& T & No \\
Llama3.2~\cite{grattafiori2024llama}   & 09.2024 & 3  & T & T & No \\
Llama3.2-vision~\cite{grattafiori2024llama} & 09.2024 & 11 & I + T & T & No \\
Llama3.3~\cite{grattafiori2024llama}   & 12.2024 & 70 & T & T & No \\
Phi4~\cite{abdin2024phi} & 12.2024 & 14 & T & T & No \\
Qwen2.5~\cite{yang2024qwen2} & 12.2024 & 3\&7\&14\&32\&72 & T & T & No \\
DeepSeek-V3~\cite{liu2024deepseek} & 12.2024 & - & T & T & No & \\
DeepSeek-R1~\cite{guo2025deepseek} & 01.2025 & 7\&8\&14\&32\&70\&671 & T & T & Yes\\
Qwen2.5-VL~\cite{bai2025qwen2} & 02.2025& 32\&72 & I + V + T & T & Yes\\
QwQ~\cite{QwQ32b} & 03.2025 & 32 & T & T & Yes \\
Gemma3~\cite{team2025gemma} & 03.2025 & 4\&12\&27 & I + T & T & No \\
Mistral-small3.1~\cite{mistral2024small31} & 03.2025 & 24 & I + T & T & No \\
Qwen2.5-Omni~\cite{xu2025qwen2} & 03.2025 & 7 & I+ V+ A + T & A + T & No\\
Qwen3~\cite{qwen2024qwen3} & 04.2025 & 4\&8\&14\&30\&32\&235 & T & T & Yes \\
Kimi-Audio~\cite{ding2025kimi} & 04.2025 & 7 & A + T & A + T & No \\
\hline
\multicolumn{6}{c}{\textit{closed-source}} \\
GPT-4o~\cite{hurst2024GPT} & 08.2024 & - & I + T & T & No \\
Qwen-Audio-Turbo~\cite{chu2024qwen2} & 12.2024 & - & A & T & No \\
Qwen-Plus~\cite{yang2024qwen2} & 01.2025 & - & T & T & No \\
Qwen-Max~\cite{yang2024qwen2} & 01.2025 & - & T & T & No \\
Qwen-VL-Plus~\cite{bai2025qwen2} & 01.2025 & - & T+I+V & T & No \\
QwQ-Plus~\cite{QwQ32b} & 03.2025 & - & T & T & Yes \\
QvQ-Max~\cite{qwen2025QvQ72b} & 03.2025 & - & I + T & T & Yes\\
Gemini-2.5-Pro~\cite{gemini2025update} & 03.2025 & - & I+ V+ A + T & T & Yes \\
Qwen-VL-Max~\cite{bai2025qwen2} & 04.2025 & - & T+I+V & T & No \\
Gemini-2.5-Flash~\cite{gemini2025update} & 04.2025 & - & I+ V+ A + T & T & Yes \\
GPT-4.1~\cite{openai2024GPT4} & 04.2025 & - & I + T & T & No \\
\bottomrule
\end{tabular}
\end{table}

\subsection{Implementation Detail of PEP-MEK}\label{app:PEPMEK}

As illustrated in Figure~\ref{fig:PEP}, there are two main components of the PEP-MEK: modality and emotion knowledge extraction and predict-explain-predict.

For the modality and emotion knowledge extraction, we ask the model to extract the modality~\cite{yin2024woodpecker} and emotion knowledge. 
Figure~\ref{fig:me_extraction} shows the prompt we designed to guide modality and emotion knowledge extraction. 
It considers both modality-specific knowledge (e.g., visual cues such as what can be seen in an image, or auditory cues such as what can be heard in speech) and emotion-specific knowledge (e.g., facial expressions, body posture, vocal prosody, emotion types and definitions, overall emotional atmosphere, and inferred causes). 
By structuring the prompt, we aim to ensure that the model captures contextually grounded and psychologically meaningful emotional signals across modalities.

Next, we prompt the model to generate an initial answer by combining the extracted modality and emotion knowledge with the target question. 
As shown in Figure~\ref{fig:predicr_first}, the prompt instructs the model to base its judgment on the input content (e.g., video, image, audio, or text), and, if needed, to incorporate the accompanying structured knowledge to refine its answer. 
The model is explicitly asked to respond with only one word, YES or NO, to enforce clarity in evaluation, without generating additional explanations.

We then prompt the MLLM to provide an explanation for its initial answer by referencing the extracted modality and emotion knowledge, the target question, and the initial answer. 
As shown in Figure~\ref{fig:predicr_second}, the model is asked to first generate a detailed explanation of its reasoning, then assess the factual accuracy and logical soundness of its own explanation, and finally re-affirm its conclusion with a concise binary answer (YES or NO). 
This structured response format allows us to evaluate not only the model’s decision, but also the justification process and self-consistency behind it.

\begin{figure}

\begin{tcolorbox}[colback=gray!5!white, colframe=black, fonttitle=\bfseries]
Please extract useful modality and emotional knowledge from the given text/image/audio/video. 
Your goal is to gather interpretable features that can support accurate emotion prediction, explanation, and refinement. Structure your response according to the following components:
            
\textbf{1. Overall Scene Mood and Context}
\begin{itemize}
\item Describe the general emotional atmosphere (e.g., joyful, tense, melancholic, peaceful).

\item Identify contextual elements in the environment (e.g., indoor/outdoor, weather, time of day, symbolic objects like candles, rain, broken glass).

\item What kind of situation or event might the scene represent (e.g., birthday party, farewell, conflict)?

\item For audio/text: consider background sounds (e.g., music, ambient noise) or narrative tone (e.g., hopeful, sarcastic, ominous).
\end{itemize}

\textbf{2. Human Presence and Character Analysis}
\begin{itemize}
  \item How many people are in the scene (visually or described in text/audio)?
  \item For each individual:
  \begin{itemize}
    \item Physical characteristics (e.g., age, gender, clothing style, if visible or described)
    \item Facial expression (e.g., happy, sad, fearful, angry, neutral, from video/image)
    \item Head pose and gaze direction (e.g., direct gaze, looking away, tilted head)
    \item Body posture and hand gesture (e.g., open arms, clenched fists, self-touch, leaning in or away)
    \item Voice tone and prosody (e.g., trembling, rising pitch, flat tone, fast or slow pace)
    \item Verbal emotional cues (e.g., exclamations, emotional word choice, metaphorical language)
    \item Speech content (e.g., direct expression of emotion: ``I'm scared,'' or indirect hints: ``I don't know what to do anymore.'')
  \end{itemize}
\end{itemize}

\textbf{3. Social and Emotional Interactions}
\begin{itemize}
  \item Describe the relationships or interactions between people:
  \begin{itemize}
    \item Are they physically close or distant?
    \item Is there eye contact, mirroring expressions, or body synchronization?
    \item Is any touch present (e.g., hugging, pushing, hand-holding)?
    \item Do they seem emotionally aligned or in conflict?
    \item In audio/text: Are their voices overlapping? Do they interrupt, agree, or show empathy?
  \end{itemize}
\end{itemize}

\textbf{4. Emotion Type, Intensity, and Diversity}
\begin{itemize}
  \item For each person, specify the likely dominant emotion and estimate its intensity (e.g., mild sadness vs. intense grief).
  \item Are there multiple emotions present in the video/audio/text, possibly conflicting ones?
\end{itemize}

\textbf{5. Emotion Knowledge and Reasoning Support}
\begin{itemize}
  \item For each identified emotion:
  \begin{itemize}
    \item Provide a short definition and its typical visual, auditory, or linguistic cues. Example: ``Fear is a response to perceived threat, often expressed by widened eyes, raised eyebrows, tense voice, and avoidance language.''
    \item If possible, suggest potential causes or triggers of these emotions based on the multimodal context.
  \end{itemize}
\end{itemize}

\end{tcolorbox}
\caption{Prompt for modality and emotion knowledge extraction.}
\label{fig:me_extraction}
\end{figure}

\begin{figure}
\begin{tcolorbox}[colback=gray!5!white, colframe=black, fonttitle=\bfseries]
Please provide a clear response to the question below by watching the video (reading the text, watching the image, listening the audio). 

If necessary, you can also use the accompanying modality and emotion knowledge to help refine your answer.

Modality and emotion knowledge:
\{\textcolor{blue}{knowledge}\}
            
Question:
\{\textcolor{blue}{question}\}

Please answer with ONLY ONE WORD: YES or NO. Do not provide any explanation or additional output.
\end{tcolorbox}
\caption{Prompt for modality and emotion knowledge extraction.}
\label{fig:predicr_first}
\end{figure}

\begin{figure}
\begin{tcolorbox}[colback=gray!5!white, colframe=black, fonttitle=\bfseries]
First, please provide a detailed explanation for your initial answer to the question. 
Then, verify both the factual accuracy of your explanation and the logic behind your answer. 
Finally, give a concise response to the question by answering with 'YES' or 'NO'.
            
Modality and emotion knowledge:
\{\textcolor{blue}{knowledge}\}
            
Question:
\{\textcolor{blue}{question}\}

Initial Answer: 
\{\textcolor{blue}{initial\_answer}\}

Answer Format:

1. [Explanation]

2. [Verification]

3. [Final Answer]
\end{tcolorbox}
\caption{Prompt for modality and emotion knowledge extraction.}
\label{fig:predicr_second}
\end{figure}

\subsubsection{Qualitative Analysis of PEP-MEK}
\begin{figure}
    \centering
    \includegraphics[width=\linewidth]{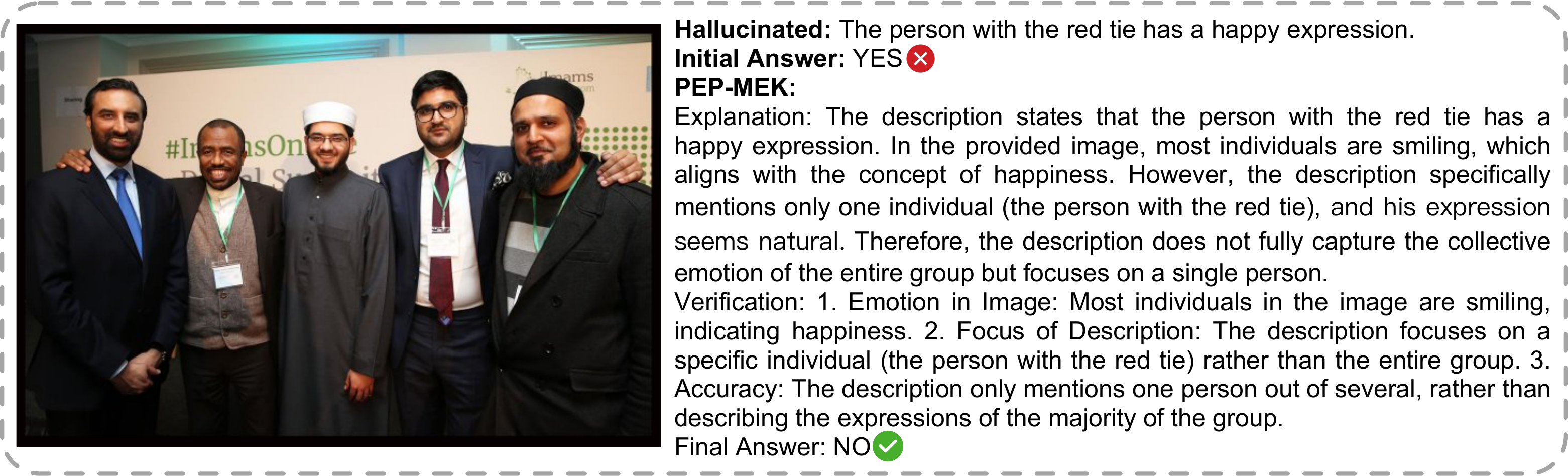}
    \caption{Qualitative analysis of PEP-MEK. In the observed image, everyone except the person wearing the red tie is smiling, displaying a happy expression. PEP-MEK noticed this detail during the explanation and ultimately provided the correct answer.}
    \label{fig:PEP_e1}
\end{figure}

As shown in Figure~\ref{fig:PEP_e1}, we present an example that benefits from the PEP-MEK approach, where the generated content originates from the Qwem2.5-Omni model. 
It is evident that in the initial response, the model exhibited a hallucination issue, failing to accurately interpret the facial expressions of individuals in the image. 
However, with the integration of PEP-MEK, the model was guided to attend to more discriminative visual details, for instance, recognizing that most individuals are smiling, except for the person wearing a red tie, who has a neutral or serious expression. 
This targeted guidance significantly improved the model’s emotion understanding of the image. Ultimately, the model identified the key feature and produced the correct answer, demonstrating the effectiveness of PEP-MEK in mitigating hallucinations and enhancing vision and enotion reasoning accuracy.

\section{More Experiment Details}\label{app:experiment}

\subsection{Result of Unimodality}

We present a more fine-grained evaluation and analysis of the model’s performance for each individual modality.

\subsubsection{Emotion Knowledge}

\begin{figure}
    \centering
    \includegraphics[width=\linewidth]{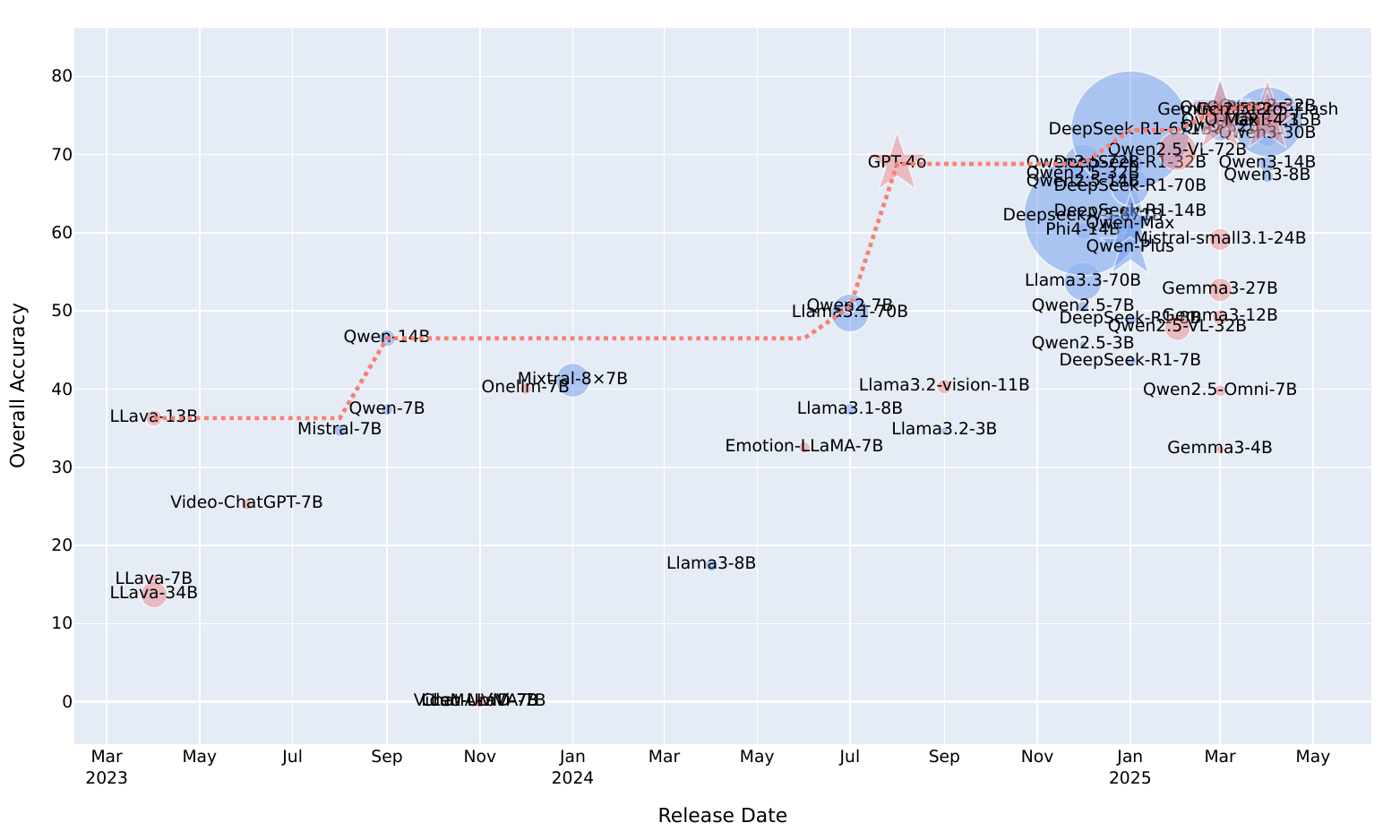}
    \caption{Performance comparison on EmotionHallucer-EK. Blue denotes LLMs, red denotes MLLMs. Circles indicate known parameter sizes, while stars represent unknown parameter sizes. Red dashed line denotes the top-performing model of the current month.}
    \label{fig:modelsize_theory}
\end{figure}

As shown in Table~\ref{tab:EmotionHallucer-EK}, we report the performance of various models on the EmotionHallucer-EK, which focuses on emotion knowledge hallucination. 
These results are further visualized in Figure~\ref{fig:modelsize_theory}. 
From the visualization, several key trends emerge: 
(1) model performance has steadily improved over time; 
(2) closed-source models generally outperform open-source models; 
(3) the incorporation of reasoning capabilities in open-source models significantly narrows the gap with closed-source models; 
(4) MLLMs generally perform worse than pure LLMs on this task; 
and (5) within the same model family, larger parameter sizes tend to correlate with better performance.
Overall, we observe a clear upward trend in model performance on emotion knowledge hallucination tasks, with many models achieving relatively high accuracy. 
This highlights the growing potential of leveraging LLMs for applications in emotion psychology, and points to future directions for further enhancing open models, particularly in multimodal emotion reasoning.

\begin{table}
 \caption{Performance comparison on EmotionHallucer-EK (Emotion Konwledge).}
  \label{tab:EmotionHallucer-EK}
  \centering
  \footnotesize
  \setlength{\tabcolsep}{5pt}
  \begin{tabular*}{\textwidth}{lcccccc}
    \toprule
    \multirow{2}{*}{Methods} & Model& \multicolumn{2}{c}{Yes/No Bias} & \multicolumn{3}{c}{Accuracy on EmotionHallucer-EK}\\ 
    \cmidrule(rl){3-4} \cmidrule(rl){5-7}
    & Size & Pct. Diff ($\sim$0) & FP Ratio ($\sim$0.5)& Basic~$\uparrow$ & Hallucinated~$\uparrow$ & Overall~$\uparrow$ \\
    \midrule
    
    \multicolumn{7}{c}{\textit{Open-source}} \\
    Mistral~\cite{jiang2023mistral7b} & 7B & 0.26 & 0.89 & 92.47 & 39.52 & 34.68 \\

    Qwen~\cite{bai2023qwen} & 7B & 0.21 & 0.81 & 87.63 & 45.70 & 37.37 \\
    Qwen~\cite{bai2023qwen} & 14B & 0.11 & 0.69 & 82.26 & 60.48 & 46.51 \\

    Mixtral~\cite{jiang2024mixtral} & 8x7B & 0.05 & 0.42 & 62.10 & 72.04 & 41.13 \\
    
    Llama3~\cite{grattafiori2024llama} & 8B & 0.40 & 0.98 & 98.39 & 18.55 & 17.47 \\
    
    Llama3.1~\cite{grattafiori2024llama} & 8B & 0.26 & 0.90 & 93.28 & 41.14 & 37.37 \\
    Llama3.1~\cite{grattafiori2024llama} & 70B & 0.22 & 0.94 & 96.77 & 51.88 & 49.73 \\

    Qwen2~\cite{yang2024qwen2technicalreport} & 7B & 0.18 & 0.86 & 93.01 & 56.18 & 50.54 \\
    
    Llama3.2~\cite{grattafiori2024llama} & 3B & 0.19 & 0.78 & 84.95 & 46.24 & 34.68 \\

    Llama3.3~\cite{grattafiori2024llama} & 70B & 0.20 & 0.92 & 95.97 & 56.18 & 53.76 \\

    Phi4~\cite{abdin2024phi} & 14B & 0.12 & 0.79 & 91.40 & 66.67 & 60.22 \\
 
    Qwen2.5~\cite{yang2024qwen2} & 3B & -0.13 & 0.27 & 58.87 & 84.95 & 45.70 \\
    Qwen2.5~\cite{yang2024qwen2} & 7B & 0.17 & 0.82 & 90.32 & 57.26 & 50.54 \\
    Qwen2.5~\cite{yang2024qwen2} & 14B & -0.01 & 0.46 & 80.91 & 83.60 & 66.40 \\
    Qwen2.5~\cite{yang2024qwen2} & 32B & 0.09 & 0.75 & 91.13 & 73.92 & 67.47 \\
    Qwen2.5~\cite{yang2024qwen2} & 72B & 0.11 & 0.83 & 94.35 & 72.31 & 68.82 \\
    
    DeepSeek-V3~\cite{liu2024deepseek} & 671B & 0.10 & 0.76 & 90.59 & 69.89 & 62.10 \\

    DeepSeek-R1~\cite{guo2025deepseek} & 7B & 0.19 & 0.80 & 87.90 & 50.27 & 43.55 \\
    DeepSeek-R1~\cite{guo2025deepseek} & 8B & 0.15 & 0.78 & 87.63 & 56.99 & 48.92 \\
    DeepSeek-R1~\cite{guo2025deepseek} & 14B & 0.08 & 0.71 & 88.17 & 71.24 & 62.63 \\
    DeepSeek-R1~\cite{guo2025deepseek} & 32B & 0.08 & 0.75 & 91.67 & 75.00 & 68.82 \\
    DeepSeek-R1~\cite{guo2025deepseek} & 70B & 0.09 & 0.76 & 91.40 & 72.58 & 65.86 \\
    DeepSeek-R1~\cite{guo2025deepseek} & 671B & 0.08 & 0.78 & 93.82 & 77.96 & 73.12 \\

    QwQ~\cite{QwQ32b} & 32B & 0.07 & 0.75 & 92.74 & 73.66 & 73.39 \\

    Qwen3~\cite{qwen2024qwen3} & 4B & 0.05 & 0.63 & 86.02 & 76.61 & 65.32 \\
    Qwen3~\cite{qwen2024qwen3} & 8B & 0.04 & 0.63 & 86.56 & 78.23 & 67.20 \\
    Qwen3~\cite{qwen2024qwen3} & 14B & 0.04 & 0.61 & 86.83 & 70.30 & 68.82 \\
    Qwen3~\cite{qwen2024qwen3} & 30B & 0.03 & 0.60 & 88.17 & 81.99 & 72.58 \\
    Qwen3~\cite{qwen2024qwen3} & 32B & 0.06 & 0.73 & 93.01 & 80.91 & \cellcolor{customFirst!45}76.08 \\
    Qwen3~\cite{qwen2024qwen3} & 235B & 0.05 & 0.67 & 91.13 & 81.72 & 74.18 \\
    
    \midrule
    LLaVA~\cite{liu2023visual} & 7B & 0.41 & 0.97 & 97.85 & 16.40 & 15.59 \\
    LLaVA~\cite{liu2023visual} & 13B & 0.23 & 0.85 & 90.32 & 43.82 & 36.29 \\
    LLaVA~\cite{liu2023visual} & 34B & 0.43 & 1.00 & 99.73 & 13.71 & 13.71 \\
    Video-ChatGPT~\cite{maaz2023video} & 7B & 0.26 & 0.83 & 86.56 & 34.95 & 25.27 \\

    Chat-UniVi~\cite{jin2024chat} & 7B & 0.50 & 1.00 & 100.00 & 0.00 & 0.00 \\
    LLaMA-VID~\cite{li2024llama} & 7B & 0.50 & 1.00 & 100.00 & 0.00 & 0.00 \\
    Video-LLaVA~\cite{lin2023video} & 7B & 0.50 & 1.00 & 100.00 & 0.00 & 0.00 \\
    Onellm~\cite{han2024onellm} & 7B & \cellcolor{customFirst!45}-0.01 & \cellcolor{customFirst!45}0.48 & 66.40 & 68.82 & 40.05 \\

    Emotion-LLaMA~\cite{cheng2024emotion} & 7B & -0.04 & 0.45 & 59.68 & 67.20 & 32.53 \\

    Mistral-small3.1~\cite{mistral2024small31} & 24B & 0.12 & 0.77 & 90.05 & 66.94 & 59.14 \\

    Llama3.2-vision~\cite{grattafiori2024llama} &  11B & 0.24 & 0.88 & 92.20 & 45.16 & 40.32 \\

    Qwen2.5-VL~\cite{bai2025qwen2} & 32B & 0.25 & 0.97 & 98.39 & 48.39 & 47.85 \\
    Qwen2.5-VL~\cite{bai2025qwen2} & 72B & 0.09 & 0.79 & 93.55 & 75.54 & 70.43 \\
    
    Gemma3~\cite{team2025gemma} & 4B & 0.28 & 0.90 & 93.01 & 36.56 & 32.26 \\
    Gemma3~\cite{team2025gemma} & 12B & 0.19 & 0.85 & 91.94 & 54.30 & 49.19 \\
    Gemma3~\cite{team2025gemma} & 27B & 0.20 & 0.92 & 96.24 & 55.38 & 52.69 \\

    Qwen2.5-Omni~\cite{xu2025qwen2} & 7B & 0.28 & 0.96 & 97.58 & 41.40 & 39.78 \\
    
    \midrule
    \multicolumn{7}{c}{\textit{Closed-source}} \\
    GPT-4o~\cite{hurst2024GPT} & - & 0.10 & 0.78 & 92.74 & 73.66 & 68.82 \\
    Qwen-Plus~\cite{yang2024qwen2} & - & 0.17 & 0.89 & 94.89 & 60.48 & 58.06 \\
    Qwen-Max~\cite{yang2024qwen2} &- & 0.14 & 0.84 & 93.28 & 66.59 & 61.02 \\
    QwQ-Plus~\cite{QwQ32b} & - & 0.07 & 0.76 & 93.55 & 79.30 & \cellcolor{customSecond!45}75.81 \\
    QvQ-Max~\cite{qwen2025QvQ72b} & - & 0.05 & 0.67 & 90.59 & 81.18 & 74.19 \\
    Gemini-2.5-Pro~\cite{gemini2025update} & - & \cellcolor{customSecond!45}0.02 & \cellcolor{customSecond!45}0.58 & 88.98 & 84.68 & 75.54 \\
    Gemini-2.5-Flash~\cite{gemini2025update} & - & 0.03 & 0.61 & 90.05 & 84.14 & 75.54 \\
    GPT-4.1~\cite{openai2024GPT4} & - & 0.08 & 0.81 & 94.62 & 77.69 & 74.19 \\
    \bottomrule
    \end{tabular*}
\end{table}

\subsubsection{Multimodality Perception: Text}

\begin{figure}
    \centering
    \includegraphics[width=\linewidth]{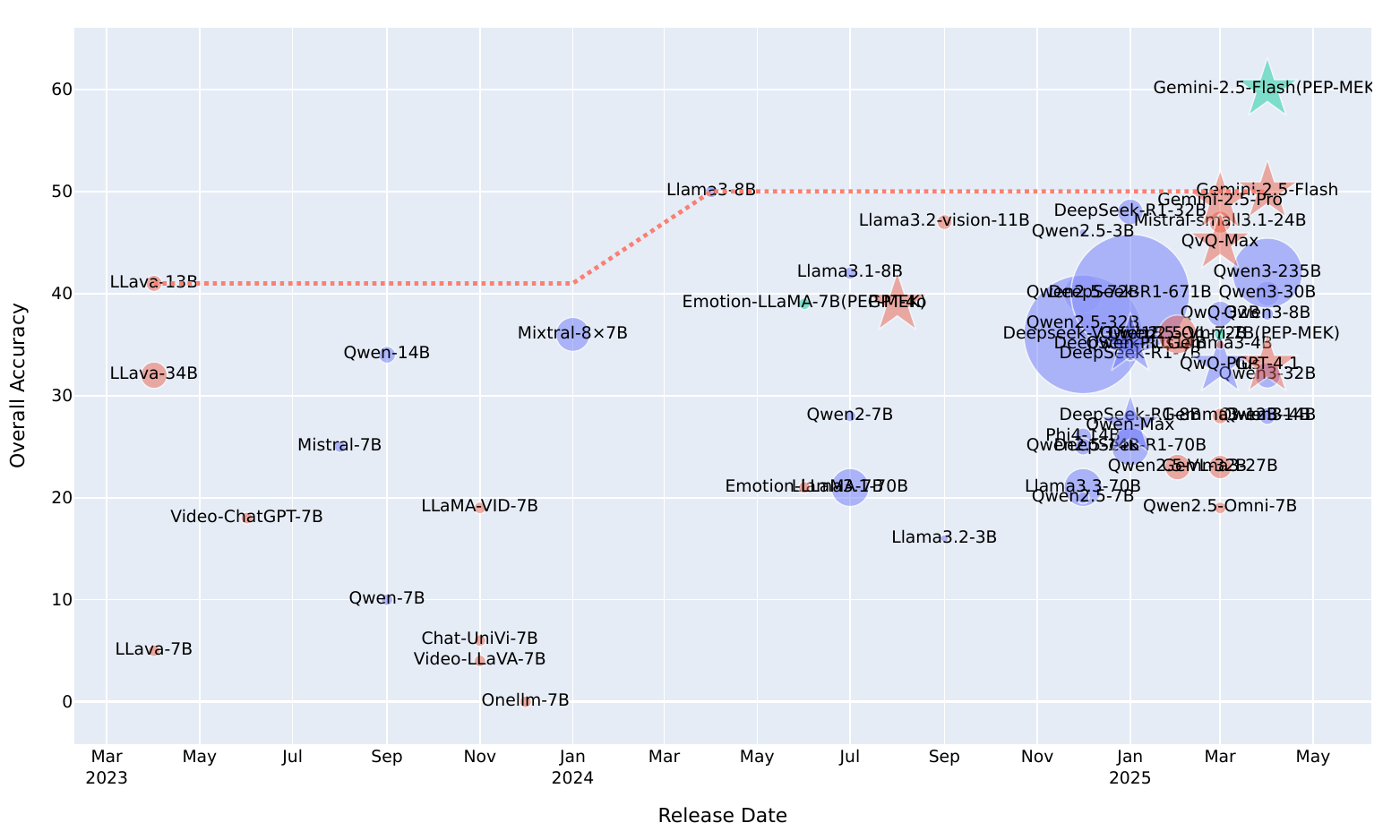}
    \caption{Performance comparison on EmotionHallucer-PT. Blue denotes LLMs, red denotes MLLMs. Circles indicate known parameter sizes, while stars represent unknown parameter sizes. Red dashed line denotes the top-performing model of the current month.}
    \label{fig:modelsize_perception_t}
\end{figure}

As shown in Table~\ref{tab:EmotionHallucer-PT}, we report the performance of various models on the EmotionHallucer-PT, which focuses on hallucinations in real-world emotion understanding from text. The results are further visualized in Figure~\ref{fig:modelsize_perception_t}. 
From the figure, several key observations emerge: 
(1) model performance has gradually improved over time; 
(2) unlike in the knowledge setting, the performance gap between closed-source and open-source models is relatively small; 
(3) the incorporation of reasoning capabilities into open-source models does not lead to better performance; 
(4) within the same model family, larger parameter sizes do not necessarily correlate with higher accuracy; 
(5) although both PT and EK benchmarks are based on text, models perform significantly worse on PT, suggesting that LLMs are more adept at retrieving structured emotional knowledge than interpreting nuanced emotions in real-world language;
and (6) PEP-MEK demonstrates its effectiveness, often achieving performance that matches or exceeds the best-performing models of the same period.

We hypothesize that this performance gap stems from two primary factors: 
(1) current LLMs struggle to detect subtle emotional shifts and cues in naturalistic language, making them prone to hallucination in text emotion understanding tasks; 
and (2) existing models may lack sufficient exposure to pretraining tasks that require fine-grained emotion reasoning in everyday contexts. 
These findings highlight a critical challenge for future work: enhancing MLLMs’ capacity to understand subtle emotional shifts in textual content.

\begin{table}
 \caption{Performance comparison on EmotionHallucer-PT (Perception Text).}
  \label{tab:EmotionHallucer-PT}
  \centering
  \footnotesize
  \setlength{\tabcolsep}{5pt}
  \begin{tabular*}{\textwidth}{lcccccc}
    \toprule
    \multirow{2}{*}{Methods} & Model & \multicolumn{2}{c}{Yes/No Bias} & \multicolumn{3}{c}{Accuracy on EmotionHallucer-PT}\\ 
    \cmidrule(rl){3-4} \cmidrule(rl){5-7}
    & Size & Pct. Diff ($\sim$0) & FP Ratio ($\sim$0.5)& Basic~$\uparrow$ & Hallucinated~$\uparrow$ & Overall~$\uparrow$ \\
    \midrule
    \multicolumn{7}{c}{\textit{Open-source}} \\
    Mixtral~\cite{jiang2024mixtral} & 7B & 0.37 & 0.99 & 99.00 & 25.00 & 25.00 \\

    Qwen~\cite{bai2023qwen} & 7B & 0.12 & 0.60 & 53.00 & 29.00 & 10.00 \\
    Qwen~\cite{bai2023qwen} & 14B & 0.07 & 0.41 & 53.00 & 67.00 & 34.00 \\

    Mixtral~\cite{jiang2024mixtral} & 8x7B & 0.14 & 0.33 & 46.00 & 73.00 & 36.00 \\
    
    Llama3~\cite{grattafiori2024llama} & 8B & 0.04 & 0.58 & 75.00 & 66.00 & \cellcolor{customFirst!45}50.00 \\

    Llama3.1~\cite{grattafiori2024llama} & 8B & 0.11 & 0.65 & 75.00 & 53.00 & 42.00 \\
    Llama3.1~\cite{grattafiori2024llama} & 70B & 0.40 & 1.00 & 100.00 & 21.00 & 21.00 \\

    Qwen2~\cite{yang2024qwen2technicalreport} & 7B & 0.34 & 0.95 & 96.00 & 28.00 & 28.00 \\

    Llama3.2~\cite{grattafiori2024llama} & 3B & 0.39 & 0.06 & 17.00 & 95.00 & 16.00 \\
    Llama3.3~\cite{grattafiori2024llama} & 70B & 0.40 & 1.00 & 100.00 & 21.00 & 21.00 \\

    Phi4~\cite{abdin2024phi} & 14B & 0.36 & 0.99 & 99.00 & 26.00 & 26.00 \\

    Qwen2.5~\cite{yang2024qwen2} & 3B & 0.19 & 0.19 & 50.00 & 88.00 & 46.00 \\
    Qwen2.5~\cite{yang2024qwen2} & 7B & 0.39 & 0.98 & 98.00 & 20.00 & 20.00 \\
    Qwen2.5~\cite{yang2024qwen2} & 14B & 0.38 & 1.00 & 100.00 & 25.00 & 25.00 \\
    Qwen2.5~\cite{yang2024qwen2} & 32B & 0.29 & 0.94 & 96.00 &38.00 & 37.00 \\
    Qwen2.5~\cite{yang2024qwen2} & 72B & 0.30 & 1.00 & 100.00 &40.00 & 40.00 \\

    DeepSeek-V3~\cite{liu2024deepseek} & -B & 0.28 & 0.93 & 95.00 & 38.00 & 36.00 \\

    DeepSeek-R1~\cite{guo2025deepseek} & 7B & 0.30 & 0.93 & 95.00 & 35.00 & 34.00 \\
    DeepSeek-R1~\cite{guo2025deepseek} & 8B & 0.34 & 0.97 & 98.00 & 29.00 & 28.00 \\
    DeepSeek-R1~\cite{guo2025deepseek} & 14B & 0.32 & 0.98 & 99.00 & 35.00 & 35.00 \\
    DeepSeek-R1~\cite{guo2025deepseek} & 32B & 0.23 & 0.93 & 96.00 & 49.00 & 48.00 \\
    DeepSeek-R1~\cite{guo2025deepseek} & 70B & 0.37 & 0.99 & 99.00 & 25.00 & 25.00 \\
    DeepSeek-R1~\cite{guo2025deepseek} & 671B & 0.28 & 0.94 & 96.00 & 40.00 & 40.00 \\

    QwQ~\cite{QwQ32b} & 32B & 0.31 & 1.00 & 100.00 & 38.00 & 38.00 \\

    Qwen3~\cite{qwen2024qwen3} & 4B & 0.35 & 0.99 & 99.00 & 28.00 & 28.00 \\
    Qwen3~\cite{qwen2024qwen3} & 8B & 0.29 & 0.94 & 99.00 & 38.00 & 38.00 \\
    Qwen3~\cite{qwen2024qwen3} & 14B & 0.33 & 0.95 & 96.00 & 30.00 & 28.00 \\
    Qwen3~\cite{qwen2024qwen3} & 30B & 0.26 & 0.89 & 93.00 & 41.00 & 40.00 \\
    Qwen3~\cite{qwen2024qwen3} & 32B & 0.33 & 0.97 & 98.00 & 32.00 & 32.00 \\
    Qwen3~\cite{qwen2024qwen3} & 235B & 0.23 & 0.89 & 93.00 & 45.00 & 42.00 \\

    \midrule
    LLaVA~\cite{liu2023visual} & 7B & 0.47 & 1.00 & 100.00 & 5.00 & 5.00 \\
    LLaVA~\cite{liu2023visual} & 13B & 0.21 & 0.18 & 46.00 & 88.00 & 41.00 \\
    LLaVA~\cite{liu2023visual} & 34B & 0.32 & 0.96 & 97.00 & 33.00 & 32.00 \\
    Video-ChatGPT~\cite{maaz2023video} & 7B & 0.32 & 0.17 & 19.00 & 83.00 & 18.00 \\
    Chat-UniVi~\cite{jin2024chat} & 7B & 0.47 & 0.01 & 6.00 & 99.00 & 6.00 \\
    LLaMA-VID~\cite{li2024llama} & 7B & 0.20 & 0.68 & 64.00 & 25.00 & 19.00 \\
    Video-LLaVA~\cite{lin2023video} & 7B & 0.47 & 0.98 & 98.00 & 4.00 & 4.00 \\
    Onellm~\cite{han2024onellm} & 7B & 0.50 & 1.00 & 100.00 & 0.00 & 0.00 \\
    Emotion-LLaMA~\cite{cheng2024emotion} & 7B & 0.28 & 0.22 & 21.00 & 78.00 & 21.00 \\
    \rowcolor{gray!20}
    \multicolumn{1}{r}{\tiny \textit{+PEP-MEK}} & & 0.14 & 0.32 & 46.00 & 75.00 & 39.00 \\
    Mistral-small3.1~\cite{mistral2024small31} & 24B & 0.20 & 0.84 & 91.00 & 51.00 & 47.00 \\
    Llama3.2-vision~\cite{grattafiori2024llama} &  11B & \cellcolor{customFirst!45}0.02 & \cellcolor{customFirst!45}0.53 & 69.00 & 65.00 & 47.00 \\
    Qwen2.5-VL~\cite{bai2025qwen2} & 32B & 0.39 & 1.00 & 100.00 & 23.00 & 23.00 \\
    Qwen2.5-VL~\cite{bai2025qwen2} & 72B & 0.32 & 0.98 & 99.00 & 36.00 & 36.00 \\
    Gemma3~\cite{team2025gemma} & 4B & 0.29 & 0.91 & 94.00 & 36.00 & 35.00 \\
    Gemma3~\cite{team2025gemma} & 12B & 0.35 & 0.97 & 98.00 & 28.00 & 28.00 \\
    Gemma3~\cite{team2025gemma} & 27B & 0.39 & 1.00 & 100.00 & 23.00 & 23.00 \\
    Qwen2.5-Omni~\cite{xu2025qwen2} & 7B & 0.41 & 1.00 & 100.00 & 19.00 & 19.00 \\
    \rowcolor{gray!20}
    \multicolumn{1}{r}{\tiny \textit{+PEP-MEK}} & & 0.28 & 0.91 & 94.00 & 38.00 & 36.00 \\
    
    \midrule
    \multicolumn{7}{c}{\textit{Closed-source}} \\
    GPT-4o~\cite{hurst2024GPT} & - & 0.28 & 0.94 & 96.00 & 39.00 & 39.00 \\
    Qwen-Plus~\cite{yang2024qwen2} & - & 0.31 & 0.96 & 97.00 & 35.00 & 35.00 \\
    Qwen-Max~\cite{yang2024qwen2} &- & 0.36 & 0.99 & 99.00 & 27.00 & 27.00 \\
    QwQ-Plus~\cite{QwQ32b} & - & 0.33 & 0.99 & 99.00 & 33.00 & 33.00 \\
    QvQ-Max~\cite{qwen2025QvQ72b} & - & 0.24 & 0.93 & 96.00 & 47.00 & 45.00 \\
    Gemini-2.5-Pro~\cite{gemini2025update} & - & 0.23 & 0.92 & 96.00 & 51.00 & 49.00 \\
    Gemini-2.5-Flash~\cite{gemini2025update} & - & \cellcolor{customSecond!45}0.21 & \cellcolor{customSecond!45}0.89 & 94.00 & 51.00 & \cellcolor{customSecond!45}50.00 \\
    \rowcolor{gray!20}
    \multicolumn{1}{r}{\tiny \textit{+PEP-MEK}} & & 0.12 & 0.74 & 88.00 & 54.00 & 60.00 \\
    GPT-4.1~\cite{openai2024GPT4} & - & 0.30 & 0.92 & 94.00 & 34.00 & 33.00 \\
    \bottomrule
  \end{tabular*}
\end{table}

\subsubsection{Multimodality Perception: Image}

\begin{figure}
    \centering
    \includegraphics[width=\linewidth]{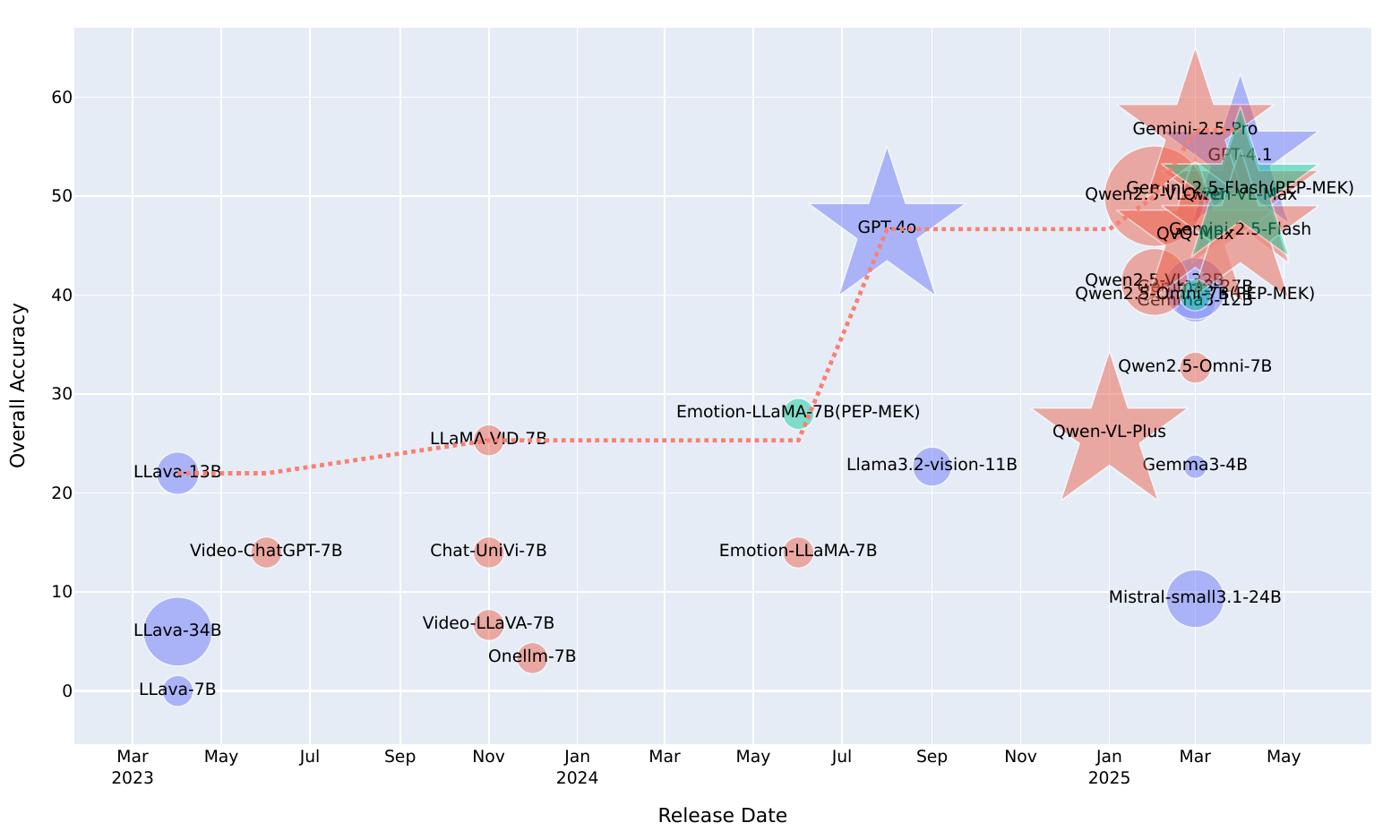}
    \caption{Performance comparison on EmotionHallucer-PI. Blue denotes MLLMs that accept only the current modality (e.g., image), red denotes models capable of handling additional modalities (e.g., video, audio), and cyan denotes the results obtained using PEP-MEK. Circles indicate known parameter sizes, while stars represent unknown sizes. The red dashed line marks the top-performing model of the current month.}
    \label{fig:modelsize_perception_i}
\end{figure}

As shown in Table~\ref{tab:EmotionHallucer-PI}, we report the performance of various models on the EmotionHallucer-PI, which focuses on hallucinations in real-world emotion understanding from images. The results are further visualized in Figure~\ref{fig:modelsize_perception_i}. 
From the figure, several key observations emerge:
(1) model performance has steadily improved over time;
(2) closed-source models generally outperform open-source models, though the performance gap is relatively small;
(3) the incorporation of reasoning capabilities into open-source models does not lead to clear performance gains;
(4) in most cases, larger parameter sizes tend to yield better results; and
(5) PEP-MEK demonstrates its effectiveness, often achieving performance that matches or exceeds the best-performing models of the same period.

Furthermore, the lack of significant improvements from reasoning-augmented models may indicate that reasoning paradigms in the visual domain are still in their early stages and require further development and refinement.
These findings highlight a critical challenge for future work: enhancing MLLMs’ capacity to understand emotional cues embedded in visual content, and to reason more effectively about complex human affect in real-world image contexts, particularly through the continued advancement and adaptation of reasoning paradigms.

\begin{table}
 \caption{Performance comparison on EmotionHallucer-PI (Perception Image).}
  \label{tab:EmotionHallucer-PI}
  \centering
  \footnotesize
  \setlength{\tabcolsep}{5pt}
  \begin{tabular*}{\textwidth}{lcccccc}
    \toprule
    \multirow{2}{*}{Methods} & Model & \multicolumn{2}{c}{Yes/No Bias} & \multicolumn{3}{c}{Accuracy on Perception-I}\\ 
    \cmidrule(rl){3-4} \cmidrule(rl){5-7}
    & Size & Pct. Diff ($\sim$0) & FP Ratio ($\sim$0.5)& Basic~$\uparrow$ & Hallucinated~$\uparrow$ & Overall~$\uparrow$ \\
    \midrule
    \multicolumn{7}{c}{\textit{Open-source}} \\
    LLaVA~\cite{liu2023visual} & 7B & 0.50 & 1.00 & 100.00 & 0.00 & 0.00 \\
    LLaVA~\cite{liu2023visual} & 13B & 0.06 & 0.44 & 47.33 & 58.67 & 22.00 \\
    LLaVA~\cite{liu2023visual} & 34B & 0.46 & 0.01 & 6.67 & 99.33 & 6.00 \\

    Video-ChatGPT~\cite{maaz2023video} & 7B & 0.04 & 0.54 & 53.33 & 45.33 & 14.00 \\

    Chat-UniVi~\cite{jin2024chat} & 7B & -0.31 & 0.18 & 19.33 & 82.00 & 14.00 \\
    LLaMA-VID~\cite{li2024llama} & 7B & -0.21 & 0.28 & 32.67 & 74.00 & 25.33\\
    Video-LLaVA~\cite{lin2023video} & 7B & 0.46 & 0.99 & 99.33 & 6.67 & 6.67 \\
    Onellm~\cite{han2024onellm} & 7B & 0.48 & 0.99 & 98.67 & 3.33 & 3.33 \\

    Emotion-LLaMA~\cite{cheng2024emotion} & 7B & 0.32 & 0.81 & 80.00 & 15.33 & 14.00 \\
    \rowcolor{gray!20}
    \multicolumn{1}{r}{\tiny \textit{+PEP-MEK}} & & 0.09 & 0.60 & 63.33 & 44.67 & 28.00 \\
    
    Llama3.2-vision~\cite{grattafiori2024llama} &  11B & 0.25 & 0.79 & 82.67 & 33.33 & 22.67 \\

    Qwen2.5-VL~\cite{bai2025qwen2} & 32B & \cellcolor{customFirst!45}-0.02 & \cellcolor{customFirst!45}0.47 & 60.67 & 65.33 & 41.33 \\
    Qwen2.5-VL~\cite{bai2025qwen2} & 72B & -0.07 & 0.39 & 64.00 & 77.33 & \cellcolor{customFirst!45}50.00 \\
    
    Gemma3~\cite{team2025gemma} & 4B & -0.14 & 0.35 & 40.00 & 68.00 & 22.67 \\
    Gemma3~\cite{team2025gemma} & 12B & -0.17 & 0.22 & 46.67 & 80.00 & 39.33 \\
    Gemma3~\cite{team2025gemma} & 27B & 0.13 & 0.66 & 74.00 & 48.67 & 40.67 \\
    
    Mistral-small3.1~\cite{mistral2024small31} & 24B & -0.44 & 0.02 & 10.67 & 98.00 & 9.33 \\
    
    Qwen2.5-Omni~\cite{xu2025qwen2} & 7B & -0.11 & 0.36 & 48.00 & 70.67 & 32.67 \\
    \rowcolor{gray!20}
    \multicolumn{1}{r}{\tiny \textit{+PEP-MEK}} & 7B & -0.11 & 0.34 & 54.00 & 76.00 & 40.00 \\
    
    \midrule
    \multicolumn{7}{c}{\textit{Closed-source}} \\
    GPT-4o~\cite{hurst2024GPT} & - & \cellcolor{customSecond!45}0.04 & \cellcolor{customSecond!45}0.56 & 72.00 & 64.00 & 46.67 \\
    QvQ-Max~\cite{qwen2025QvQ72b} & - & 0.04 & 0.57 & 72.57 & 64.00 & 46.00 \\
    Gemini-2.5-Pro~\cite{gemini2025update} & - & 0.06 & 0.62 & 82.00 & 70.00 & \cellcolor{customSecond!45}56.67 \\
    Qwen-VL-Plus~\cite{bai2025qwen2} & - & -0.24 & 0.22 & 34.67 & 82.00 & 26.00 \\
    Qwen-VL-Max~\cite{bai2025qwen2} & - & -0.07 & 0.38 & 63.33 & 78.00 & 50.00 \\
    Gemini-2.5-Flash~\cite{gemini2025update} & - & 0.05 & 0.58 & 74.67 & 64.67 & 46.67 \\
    \rowcolor{gray!20}
    \multicolumn{1}{r}{\tiny \textit{+PEP-MEK}} & - & 0.02 & 0.53 & 73.33 & 69.33 & 50.67 \\
    GPT-4.1~\cite{openai2024GPT4} & - & 0.07 & 0.61 & 77.33 & 64.00 & 54.00 \\
    \bottomrule
  \end{tabular*}
\end{table}

\subsubsection{Multimodality Perception: Audio}

\begin{figure}
    \centering
    \includegraphics[width=\linewidth]{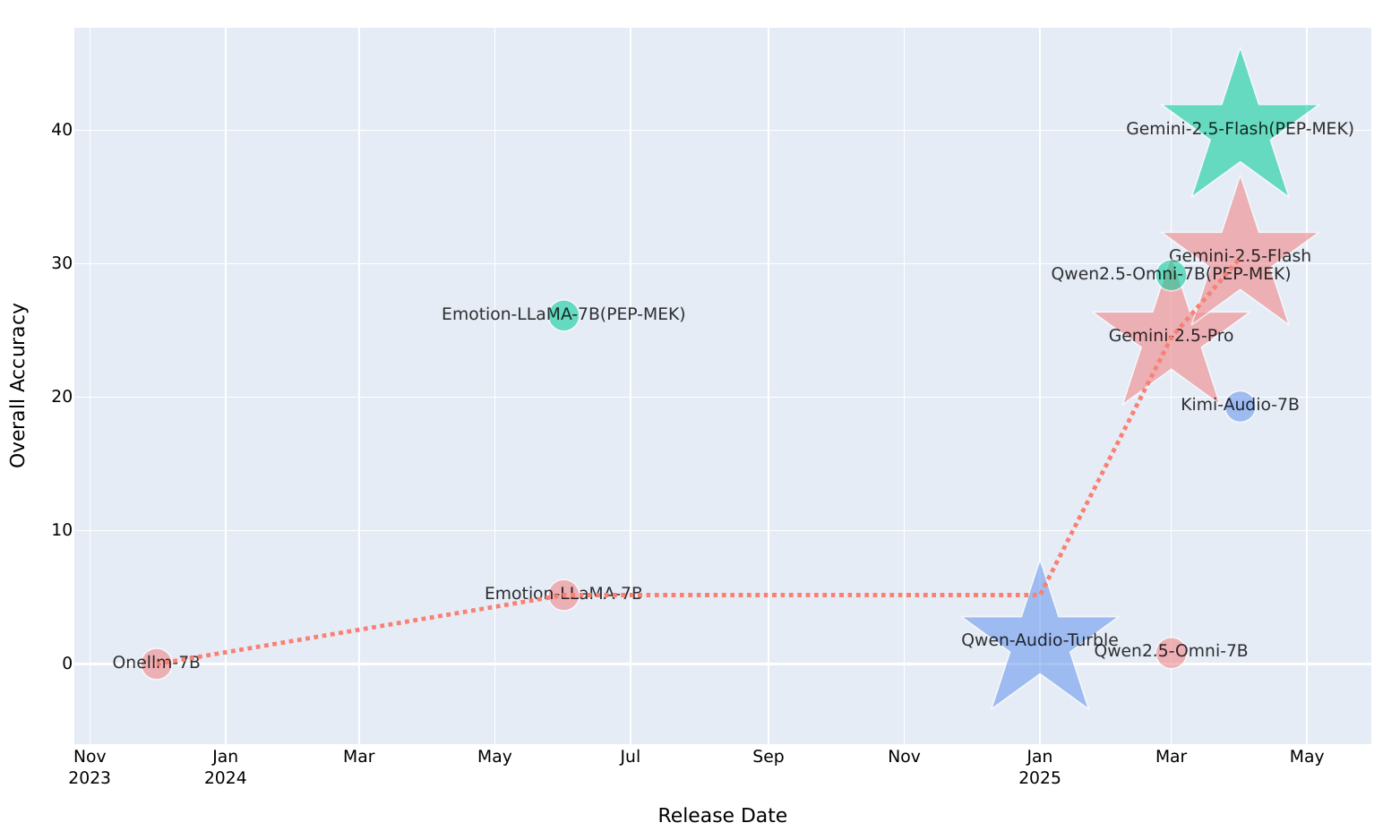}
    \caption{Performance comparison on EmotionHallucer-PA. Blue denotes MLLMs that accept only the current modality (e.g., audio), red denotes models capable of handling additional modalities (e.g., iamge, video), and cyan denotes the results obtained using PEP-MEK. Circles indicate known parameter sizes, while stars represent unknown sizes. The red dashed line marks the top-performing model of the current month.}
    \label{fig:modelsize_perception_a}
\end{figure}

As shown in Table~\ref{tab:EmotionHallucer-PA}, we report the performance of various models on the EmotionHallucer-PA benchmark, which focuses on hallucinations in real-world, perception-based emotion understanding from audio. The results are further visualized in Figure~\ref{fig:modelsize_perception_a}. From the figure, several key observations emerge:
(1) model performance has gradually improved over time, though it remains in the early stages, with most models only slightly exceeding the random guess baseline of 25\%;
(2) closed-source models significantly outperform open-source models; and
(3) PEP-MEK demonstrates strong effectiveness, often achieving performance that surpasses the best models available at the time.

We hypothesize that the relatively low performance in the audio modality is due to the current focus of audio-based MLLM research, which primarily targets tasks such as Automatic Speech Recognition (ASR) and semantic audio understanding. For instance, when refusing to answer, GPT-4o-Audio responds with: ``I'm sorry, but I can’t help with that request.''
These tasks emphasize transcribing or interpreting spoken content, but often neglect paralinguistic features, such as tone, intensity, and prosody, which are essential for emotion understanding.
These findings highlight a critical challenge for future work: enhancing MLLMs’ ability to perceive and interpret emotional signals embedded in speech, beyond lexical content. This calls for the development of audio-specific reasoning paradigms and training objectives that better capture the richness of human emotional expression in voice.

\begin{table}
 \caption{Performance comparison on EmotionHallucer-PA (Perception Audio).}
  \label{tab:EmotionHallucer-PA}
  \centering
  \footnotesize
  \setlength{\tabcolsep}{4.5pt}
  \begin{tabular*}{\textwidth}{lcccccc}
    \toprule
    \multirow{2}{*}{Methods} & Model & \multicolumn{2}{c}{Yes/No Bias} & \multicolumn{3}{c}{Accuracy on EmotionHallucer-PA}\\ 
    \cmidrule(rl){3-4} \cmidrule(rl){5-7}
    & Size & Pct. Diff ($\sim$0) & FP Ratio ($\sim$0.5)& Basic~$\uparrow$ & Hallucinated~$\uparrow$ & Overall~$\uparrow$ \\
    \midrule
    \multicolumn{7}{c}{\textit{Open-source}} \\
    Onellm~\cite{han2024onellm} & 7B & 0.50 & 1.00 & 100.00 & 0.00 & 0.00 \\
    Emotion-LLaMA~\cite{cheng2024emotion} & 7B & 0.43 & 0.92 & 91.85 & 6.52 & 5.16 \\
    \rowcolor{gray!20}
    \multicolumn{1}{r}{\tiny \textit{+PEP-MEK}} & & -0.14 & 0.36 & 33.68 & 64.67 & 22.28 \\
    Qwen2.5-Omni~\cite{xu2025qwen2} & 7B & -0.49 & 0.01 & 1.36 & 99.46 & 0.82 \\
    \rowcolor{gray!20}
    \multicolumn{1}{r}{\tiny \textit{+PEP-MEK}} & & -0.45 & 0.04 & 6.25 & 95.92 & 5.43 \\
    Kimi-Audio~\cite{ding2025kimi} & 7B & -0.23 & \cellcolor{customFirst!45}0.22 & \cellcolor{customFirst!45}36.41 & 81.79 & \cellcolor{customFirst!45}19.29 \\
    \midrule
    \multicolumn{7}{c}{\textit{Closed-source}} \\
    Qwen-Audio-Turbo~\cite{chu2024qwen2} & - & -0.49 & 0.01 & 2.17 & 99.46 & 1.63 \\
    Gemini-2.5-Pro~\cite{gemini2025update} & - & -0.14 & 0.34 & 41.03 & 69.02 & 24.46 \\
    Gemini-2.5-Flash~\cite{gemini2025update} & - & \cellcolor{customSecond!45}-0.13 & \cellcolor{customSecond!45}0.34 & 45.11 & 71.74 & \cellcolor{customSecond!45}30.43 \\
    \rowcolor{gray!20}
    \multicolumn{1}{r}{\tiny \textit{+PEP-MEK}} & & -0.12 & 0.34 & 48.10 & 72.83 & 34.51 \\
    \bottomrule
  \end{tabular*}
\end{table}

\subsubsection{Multimodality Perception: Short Video}

\begin{figure}
    \centering
    \includegraphics[width=\linewidth]{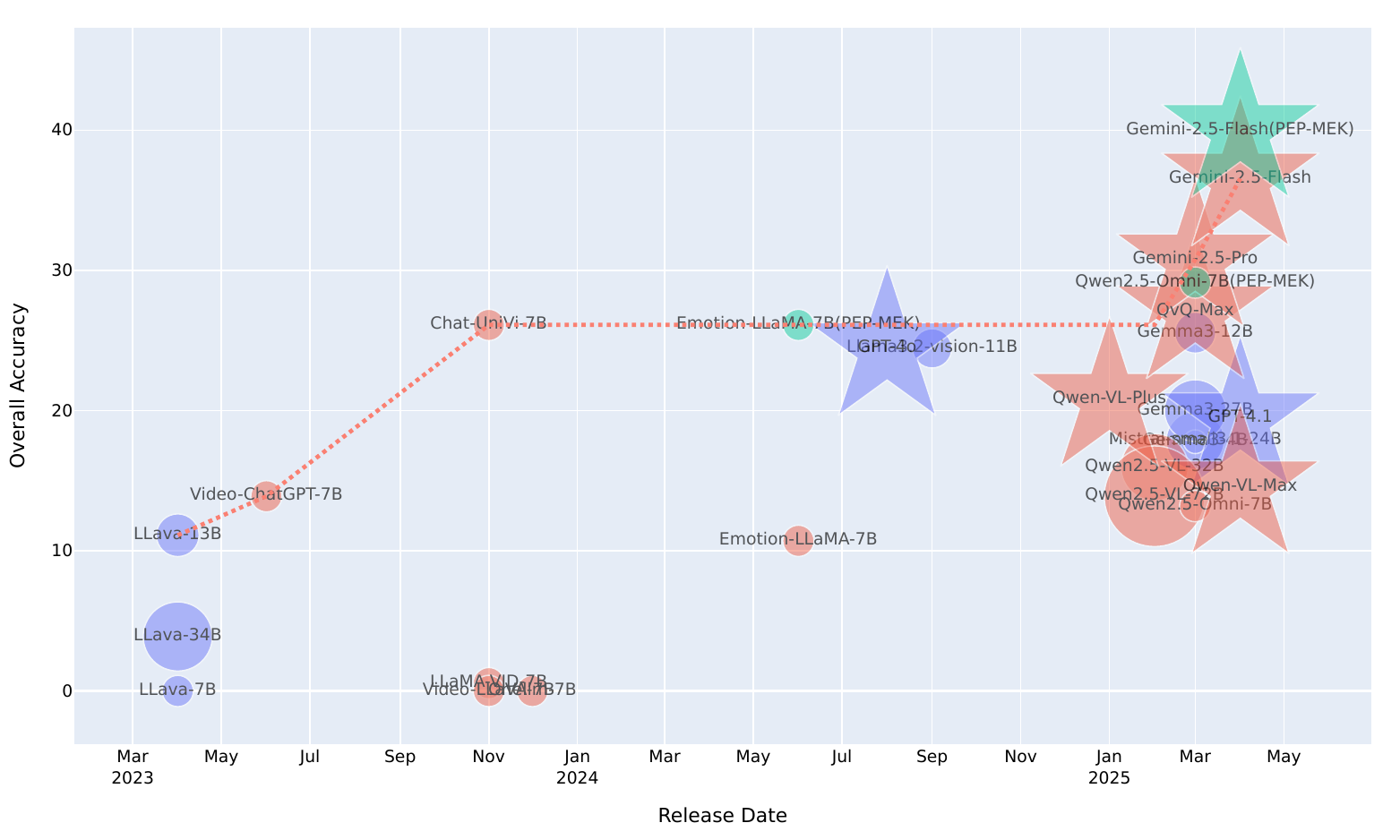}
    \caption{Performance comparison on EmotionHallucer-PV/S. Blue denotes MLLMs that accept only the current modality (e.g., video), red denotes models capable of handling additional modalities (e.g., image, audio), and cyan denotes the results obtained using PEP-MEK. Circles indicate known parameter sizes, while stars represent unknown sizes. The red dashed line marks the top-performing model of the current month.}
    \label{fig:modelsize_perception_vs}
\end{figure}

As shown in Table~\ref{tab:EmotionHallucer-PV/S}, we report the performance of various models on the EmotionHallucer-PV/S, which focuses on hallucinations in real-world, perception-based emotion understanding from short videos. The results are further visualized in Figure~\ref{fig:modelsize_perception_vs}. For models that only support image input, we adopt a key-frame sampling strategy to enable evaluation on video data.

Several key observations emerge from the figure:
(1) model performance has steadily improved over time;
(2) closed-source models generally outperform open-source models, though the performance gap remains modest;
(3) for MLLMs that only support image input, increasing sampled frames typically does not lead to performance improvements;
(4) some models, such as LLaMA-VID, Video-LLaVA, and OneLLM—perform particularly poorly on this task;
(5) Emotion LLaMA, despite being fine-tuned on emotion recognition tasks, does not outperform general-purpose models;
(6) PEP-MEK continues to demonstrate strong performance, often matching or exceeding that of the best models during the same period.

Furthermore, video-based emotion understanding and hallucination detection remains an open and underexplored challenge, with most models still in the early developmental stage. 
Our results also indicate that supervised finetuning on emotion tasks does not necessarily alleviate hallucination issues. Notably, although Emotion-LLaMA achieves excellent performance on MER, it performs poorly on EmotionHallucer-PV/S, constructed from the same data source but reframed for hallucination detection. 

These findings highlight a critical challenge for future work: to develop video-language models with stronger emotion recognition and reasoning capabilities, and to identify effective strategies for mitigating hallucinations introduced by supervised finetuning~\cite{perez2022red,achiam2023gpt}.

\begin{table}
 \caption{Performance comparison on EmotionHallucer-PV/S (Perception Short Video).}
  \label{tab:EmotionHallucer-PV/S}
  \centering
  \footnotesize
  \setlength{\tabcolsep}{4.5pt}
  \begin{tabular*}{\textwidth}{lcccccc}
    \toprule
    \multirow{2}{*}{Methods} & Model & \multicolumn{2}{c}{Yes/No Bias} & \multicolumn{3}{c}{Accuracy on EmotionHallucer-PV/S}\\ 
    \cmidrule(rl){3-4} \cmidrule(rl){5-7}
    & Size & Pct. Diff ($\sim$0) & FP Ratio ($\sim$0.5)& Basic~$\uparrow$ & Hallucinated~$\uparrow$ & Overall~$\uparrow$ \\
    \midrule
    \multicolumn{7}{c}{\textit{Open-source}} \\
    LLava~\cite{liu2023visual}/1F & 7B & 0.50 & 1.00 & 100.00 & 0.00 & 0.00 \\
    LLava~\cite{liu2023visual}/1F & 13B & -0.34 & 0.14 & 19.44 & 87.22 & 11.11 \\
    LLava~\cite{liu2023visual}/1F & 34B & -0.44 & 0.04 & 7.22 & 96.11 & 3.89 \\
    Llama3.2-vision~\cite{grattafiori2024llama}/1F & 11B & 0.18 & 0.72 & 76.67 & 40.00 & 24.44 \\
    Llama3.2-vision~\cite{grattafiori2024llama}/2F & 11B & -0.46 & 0.02 & 5.00 & 97.78 & 3.33 \\
    Llama3.2-vision~\cite{grattafiori2024llama}/4F & 11B & -0.49 & 0.01 & 1.67 & 99.44 & 1.11 \\
    Gemma3~\cite{team2025gemma}/1F & 4B & 0.21 & 0.74 & 76.11 & 33.33 & 17.78 \\
    Gemma3~\cite{team2025gemma}/2F & 4B & 0.24 & 0.74 & 75.00 & 27.22 & 14.44 \\
    Gemma3~\cite{team2025gemma}/4F & 4B & 0.22 & 0.73 & 75.00 & 31.67 & 15.00 \\
    Gemma3~\cite{team2025gemma}/1F & 12B & 0.16 & 0.69 & 73.68 & 42.78 & 25.56 \\
    Gemma3~\cite{team2025gemma}/2F & 12B & 0.15 & 0.66 & 69.44 & 39.44 & 20.56 \\
    Gemma3~\cite{team2025gemma}/4F & 12B & \cellcolor{customFirst!45}0.03 & \cellcolor{customFirst!45}0.53 & 58.33 & 52.22 & 20.56 \\
    Gemma3~\cite{team2025gemma}/1F & 27B & 0.28 & 0.82 & 83.89 & 27.78 & 17.22 \\
    Gemma3~\cite{team2025gemma}/2F & 27B & 0.22 & 0.73 & 75.00 & 31.11 & 17.22 \\
    Gemma3~\cite{team2025gemma}/4F & 27B & 0.18 & 0.70 & 72.22 & 36.67 & 20.00 \\
    Mistral-small3.1~\cite{mistral2024small31}/1F & 24B & -0.13 & 0.35 & 41.11 & 67.78 & 17.78 \\
    Mistral-small3.1~\cite{mistral2024small31}/2F & 24B & -0.23 & 0.24 & 32.78 & 78.89 & 17.22 \\
    Mistral-small3.1~\cite{mistral2024small31}/4F & 24B & -0.38 & 0.11 & 12.22 & 88.89 & 6.11 \\
    \midrule
    Video-ChatGPT~\cite{maaz2023video} & 7B & -0.09 & 0.41 & 39.44 & 58.33 & 13.89 \\
    Chat-UniVi~\cite{jin2024chat} & 7B & -0.16 & 0.32 & 39.44 & 71.67 & \cellcolor{customFirst!45}26.11 \\
    LLaMA-VID~\cite{li2024llama} & 7B & 0.49 & 0.99 & 99.44 & 0.56 & 0.56 \\
    Video-LLaVA~\cite{lin2023video} & 7B & 0.50 & 1.00 & 100.00 & 0.00 & 0.00 \\
    Onellm~\cite{han2024onellm} & 7B & 0.50 & 1.00 & 100.00 & 0.00 & 0.00 \\
    Emotion-LLaMA~\cite{cheng2024emotion} & 7B & 0.20 & 0.70 & 70.71 & 31.07 & 10.71 \\
    \rowcolor{gray!20}
    \multicolumn{1}{r}{\tiny \textit{+PEP-MEK}} & & -0.12 & 0.35 & 45.00 & 70.00 & 26.11 \\
    Qwen2.5-VL~\cite{bai2025qwen2} & 32B & 0.32 & 0.95 & 86.11 & 22.92 & 15.97 \\
    Qwen2.5-VL~\cite{bai2025qwen2} & 72B & 0.38 & 0.92 & 93.06 & 18.06 & 13.89 \\
    Qwen2.5-Omni~\cite{xu2025qwen2} & 7B & 0.37 & 0.91 & 91.67 & 17.36 & 13.19 \\
    \rowcolor{gray!20}
    \multicolumn{1}{r}{\tiny \textit{+PEP-MEK}} &  & -0.09 & 0.38 & 51.39 & 70.14 & 29.17 \\
    \midrule
    \multicolumn{7}{c}{\textit{Closed-source}} \\
    GPT-4o~\cite{hurst2024GPT}/1F & - & \cellcolor{customSecond!45}0.00 & \cellcolor{customSecond!45}0.50 & 57.78 & 57.22 & 20.56 \\
    GPT-4o~\cite{hurst2024GPT}/2F & - & 0.05 & 0.56 & 62.22 & 52.78 & 24.44 \\
    QvQ-Max~\cite{qwen2025QvQ72b} & - & 0.23 & 0.76 & 79.86 & 34.73 & 27.08 \\
    Gemini-2.5-Pro~\cite{gemini2025update} & - & 0.24 & 0.81 & 85.90 & 36.46 & 30.77 \\
    Qwen-VL-Plus~\cite{bai2025qwen2} & - & -0.14 & 0.33 & 43.75 & 72.22 & 20.83 \\
    Qwen-VL-Max~\cite{bai2025qwen2} & - & 0.38 & 0.93 & 93.75 & 17.36 & 14.58 \\
    Gemini-2.5-Flash~\cite{gemini2025update} & - & 0.19 & 0.77 & 83.33 & 45.51 & \cellcolor{customSecond!45}36.54 \\
    \rowcolor{gray!20}
    \multicolumn{1}{r}{\tiny \textit{+PEP-MEK}} & & -0.04 & 0.45 & 62.78 & 70.00 & 40.00 \\
    GPT-4.1~\cite{openai2024GPT4}/1F & - & 0.20 & 0.72 & 75.56 & 35.56 & 19.44 \\
    \bottomrule
  \end{tabular*}
\end{table}

\subsubsection{Multimodality Perception: Long Video}

\begin{figure}
    \centering
    \includegraphics[width=\linewidth]{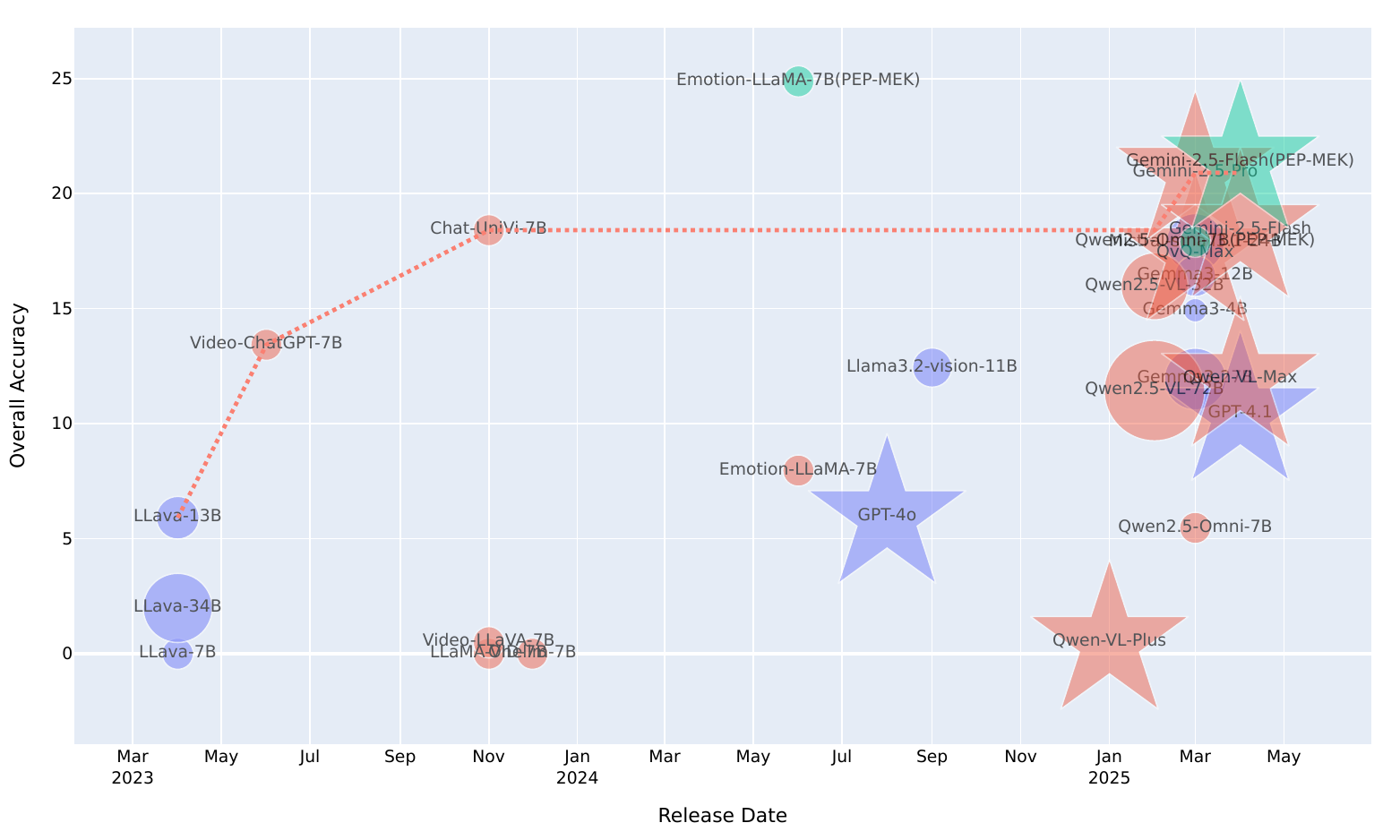}
    \caption{Performance comparison on EmotionHallucer-PV/L. Blue denotes MLLMs that accept only the current modality (e.g., video), red denotes models capable of handling additional modalities (e.g., image, audio), and cyan denotes the results obtained using PEP-MEK. Circles indicate known parameter sizes, while stars represent unknown sizes. The red dashed line marks the top-performing model of the current month.}
    \label{fig:modelsize_perception_vl}
\end{figure}

As shown in Table~\ref{tab:EmotionHallucer-PV/L}, we report the performance of various models on the EmotionHallucer-PV/L benchmark, which focuses on hallucinations in real-world, perception-based emotion understanding from long-form videos. The results are visualized in Figure~\ref{fig:modelsize_perception_vl}. For models that only support image input, we adopt a key-frame sampling strategy to approximate video-level evaluation.

From the figure, several key observations emerge:
(1) model performance has gradually improved over time, but overall results remain low, indicating that this task is still in its early stages;
(2) closed-source models generally outperform open-source ones, though the performance gap is not as large as in other benchmarks;
(3) nearly all models perform at or below random guess levels, underscoring the difficulty of emotion understanding in long video contexts;
(4) Emotion LLaMA, despite being fine-tuned on emotion recognition tasks, fails to outperform general-purpose models, suggesting limited transferability to hallucination detection;
(5) PEP-MEK demonstrates strong and consistent effectiveness, in many cases outperforming all models released in the same timeframe.

Additionally, we note that many MLLMs perform surprisingly poorly on this task, despite their strength on other modalities and tasks. 
This discrepancy suggests that current MLLMs are not yet equipped to reason about fine-grained emotion states and changes over long temporal spans, and may lack both temporal integration and emotion-specific understanding capabilities.

These findings highlight a critical challenge for future work: to develop temporal-aware, emotion-grounded MLLMs capable of robust reasoning over long-form emotional content. It will also be essential to explore new reasoning paradigms and training strategies that directly target hallucination resilience in dynamic emotional contexts.

\begin{table}
 \caption{Performance comparison on EmotionHallucer-PV/L (Perception Long Video).}
  \label{tab:EmotionHallucer-PV/L}
  \centering
  \footnotesize
  \setlength{\tabcolsep}{4.5pt}
  \begin{tabular*}{\textwidth}{lcccccc}
    \toprule
    \multirow{2}{*}{Methods} & Model & \multicolumn{2}{c}{Yes/No Bias} & \multicolumn{3}{c}{Accuracy on EmotionHallucer-PV/L}\\ 
    \cmidrule(rl){3-4} \cmidrule(rl){5-7}
    & Size & Pct. Diff ($\sim$0) & FP Ratio ($\sim$0.5)& Basic~$\uparrow$ & Hallucinated~$\uparrow$ & Overall~$\uparrow$ \\
    \midrule
    \multicolumn{7}{c}{\textit{Open-source}} \\
    LLava~\cite{liu2023visual}/1F & 7B & 0.50 & 1.00 & 100.00 & 0.00 & 0.00 \\
    LLava~\cite{liu2023visual}/1F & 13B & -0.13 & 0.37 & 38.31 & 63.68 & 5.97 \\
    LLava~\cite{liu2023visual}/1F & 34B & -0.44 & 0.05 & 6.47 & 95.02 & 1.99 \\
    Llama3.2-vision~\cite{grattafiori2024llama}/1F & 11B & 0.25 & 0.77 & 79.10 & 29.35 & 12.44 \\
    Llama3.2-vision~\cite{grattafiori2024llama}/2F & 11B & -0.38 & 0.11 & 12.44 & 89.05 & 3.48 \\
    Llama3.2-vision~\cite{grattafiori2024llama}/4F & 11B & -0.49 & 0.01 & 0.50 & 98.51 & 0.00 \\
    Gemma3~\cite{team2025gemma}/1F & 4B & -0.03 & 0.47 & 49.25 & 55.22 & 12.94 \\
    Gemma3~\cite{team2025gemma}/2F & 4B & -0.12 & 0.37 & 41.79 & 65.67 & 10.95 \\
    Gemma3~\cite{team2025gemma}/4F & 4B & -0.06 & 0.43 & 49.25 & 61.69 & 14.93 \\
    Gemma3~\cite{team2025gemma}/1F & 12B & -0.07 & 0.42 & 48.26 & 62.69 & 16.42 \\
    Gemma3~\cite{team2025gemma}/2F & 12B & -0.10 & 0.39 & 44.28 & 64.18 & 13.43 \\
    Gemma3~\cite{team2025gemma}/4F & 12B & -0.15 & 0.33 & 40.30 & 70.65 & 13.93 \\
    Gemma3~\cite{team2025gemma}/1F & 27B & -0.01 & 0.49 & 52.74 & 54.23 & 9.95 \\
    Gemma3~\cite{team2025gemma}/2F & 27B & \cellcolor{customFirst!45}0.00 & \cellcolor{customFirst!45}0.50 & 54.23 & 54.73 & 11.44 \\
    Gemma3~\cite{team2025gemma}/4F & 27B & -0.09 & 0.41 & 44.78 & 62.19 & 11.94 \\
    Mistral-small3.1~\cite{mistral2024small31}/1F & 24B & -0.40 & 0.08 & 12.44 & 92.04 & 4.98 \\
    Mistral-small3.1~\cite{mistral2024small31}/2F & 24B & -0.45 & 0.05 & 5.47 & 95.52 & 2.49 \\
    Mistral-small3.1~\cite{mistral2024small31}/4F & 24B & -0.48 & 0.02 & 2.99 & 98.51 & 1.49 \\
    \midrule
    Video-ChatGPT~\cite{maaz2023video} & 7B & 0.17 & 0.66 & 64.18 & 30.85 & 13.43 \\
    Chat-UniVi~\cite{jin2024chat} & 7B & 0.13 & 0.63 & 62.69 & 36.82 & \cellcolor{customFirst!45}18.41 \\
    LLaMA-VID~\cite{li2024llama} & 7B & 0.49 & 0.98 & 97.51 & 0.00 & 0.00 \\
    Video-LLaVA~\cite{lin2023video} & 7B & 0.50 & 1.00 & 100.00 & 0.50 & 0.50 \\
    Onellm~\cite{han2024onellm} & 7B & 0.50 & 1.00 & 100.00 & 0.00 & 0.00 \\
    Emotion-LLaMA~\cite{cheng2024emotion} & 7B & 0.36 & 0.86 & 86.57 & 14.93 & 7.96 \\
    \rowcolor{gray!20}
    \multicolumn{1}{r}{\tiny \textit{+PEP-MEK}} & & -0.04 & 0.45 & 52.74 & 60.70 & 24.88 \\
    Qwen2.5-VL~\cite{bai2025qwen2} & 32B & -0.17 & 0.31 & 36.32 & 71.14 & 9.95 \\
    Qwen2.5-VL~\cite{bai2025qwen2} & 72B & -0.16 & 0.32 & 38.81 & 70.65 & 11.44 \\
    Qwen2.5-Omni~\cite{xu2025qwen2} & 7B & -0.36 & 0.12 & 16.42 & 88.06 & 5.47 \\
    \rowcolor{gray!20}
    \multicolumn{1}{r}{\tiny \textit{+PEP-MEK}} & & -0.06 & 0.43 & 44.28 & 57.21 & 17.91 \\
    \midrule
    \multicolumn{7}{c}{\textit{Closed-source}} \\
    GPT-4o~\cite{hurst2024GPT}/1F & - & -0.41 & 0.08 & 9.95 & 92.04 & 4.48 \\
    GPT-4o~\cite{hurst2024GPT}/2F & - & -0.37 & 0.12 & 15.42 & 88.56 & 5.97 \\
    QvQ-Max~\cite{qwen2025QvQ72b} & - & -0.05 & 0.45 & 49.25 & 58.71 & 17.41 \\
    Gemini-2.5-Pro~\cite{gemini2025update} & - & \cellcolor{customSecond!45}-0.03 & \cellcolor{customSecond!45}0.46 & 55.72 & 62.19 & \cellcolor{customSecond!45}20.90 \\
    Qwen-VL-Plus~\cite{bai2025qwen2} & - & -0.49 & 0.01 & 0.50 & 99.00 & 0.00 \\
    Qwen-VL-Max~\cite{bai2025qwen2} & - & -0.05 & 0.45 & 49.25 & 59.20 & 11.94 \\
    Gemini-2.5-Flash~\cite{gemini2025update} & - & -0.06 & 0.43 & 48.76 & 60.70 & 18.41 \\
    \rowcolor{gray!20}
    \multicolumn{1}{r}{\tiny \textit{+PEP-MEK}} & & -0.19 & 0.29 & 37.81 & 75.12 & 21.39 \\
    GPT-4.1~\cite{openai2024GPT4}/1F & - & -0.29 & 0.19 & 24.88 & 82.59 & 10.45 \\
    \bottomrule
  \end{tabular*}
\end{table}

\subsection{Result of Multimodality}

\begin{figure}
    \centering
    \includegraphics[width=\linewidth]{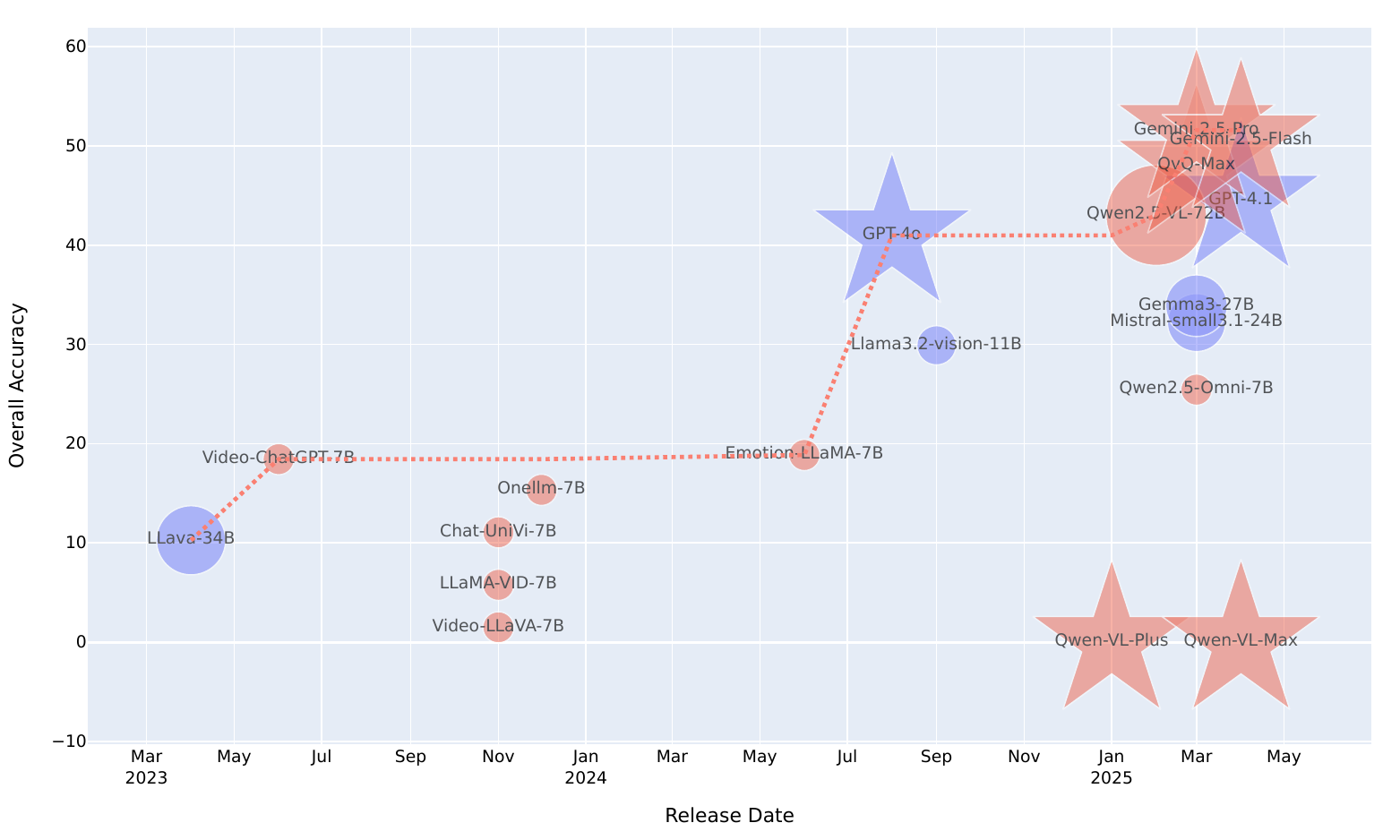}
    \caption{Performance comparison on EmotionHallucer-NoAudio. Blue denotes MLLMs that accept only the image modality, red denotes models capable of handling additional modalities (e.g., video, audio). Circles indicate known parameter sizes, while stars represent unknown sizes. The red dashed line marks the top-performing model of the current month.}
    \label{fig:modelsize_perception_woa}
\end{figure}

We further present the hallucination performance of models on EmotionHallucer-NoAudio in Table~\ref{tab:sota-ewoa-E} and Figure~\ref{fig:modelsize_perception_woa}, providing an overall view of multimodal capability. 
From the figure, several key trends can be observed:
(1) model performance has consistently improved over time, reflecting the rapid progress in multimodal learning;
(2) recent image-based MLLMs outperform earlier models with video-processing capabilities, suggesting advances in visual understanding even without temporal cues;
(3) open-source models still lag behind closed-source counterparts, though the performance gap is gradually narrowing;
(4) overall, the Gemini series stands out for both its ability to handle all input modalities and its strong overall performance across tasks.

\begin{table}
 \caption{Performance comparison on EmotionHallucer-NoAudio with additional ``Yes/No bias'' analysis.}
   \label{tab:sota-ewoa-E}
  \centering
  \footnotesize
  \setlength{\tabcolsep}{3.8pt}
  \begin{tabular*}{\textwidth}{lcccccc}
    \toprule
    \multirow{2}{*}{Methods} & Model & \multicolumn{2}{c}{Yes/No Bias} & \multicolumn{3}{c}{Accuracy on EmotionHallucer–NoAudio}\\ 
    \cmidrule(rl){3-4} \cmidrule(rl){5-7}
    & Size & Pct. Diff ($\sim$0) & FP Ratio ($\sim$0.5)& Basic~$\uparrow$ & Hallucinated~$\uparrow$ & Overall~$\uparrow$ \\
    \midrule
    \multicolumn{7}{c}{\textit{Open-source}} \\
    LLaVA~\cite{liu2023visual} & 34B & -0.05 & 0.45 & 50.25 & 59.52 & 10.27 \\
    Video-ChatGPT~\cite{maaz2023video} & 7B & \cellcolor{customFirst!45}0.09 & \cellcolor{customFirst!45}0.59 & 61.91 & 44.67 & 18.44 \\
    Chat-UniVi~\cite{jin2024chat} & 7B & \cellcolor{customFirst!45}0.09 & \cellcolor{customFirst!45}0.59 & 60.22 & 42.37 & 11.07 \\
    LLaMA-VID~\cite{li2024llama} & 7B & 0.36 & 0.86 & 85.74 & 13.66 & 5.78 \\
    Video-LLaVA~\cite{lin2023video} & 7B & 0.49 & 1.00 & 99.70 & 1.50 & 1.50 \\
    Onellm~\cite{han2024onellm} & 7B & 0.31 & 0.85 & 87.34 & 26.02 & 15.35 \\
    Emotion-LLaMA~\cite{cheng2024emotion} & 7B & 0.12 & 0.63 & 66.55 & 42.43 & 18.86 \\
    Llama3.2-vision~\cite{grattafiori2024llama} &  11B & 0.21 & 0.78 & 83.05 & 41.28 & 29.91 \\
    Gemma3~\cite{team2025gemma} & 27B & 0.15 & 0.70 & 78.66 & 49.15 & 33.90 \\
    Qwen2.5-VL~\cite{bai2025qwen2} & 72B & 0.08 & 0.63 & 78.08 & 62.15 & \cellcolor{customFirst!45}43.02 \\
    Mistral-small3.1~\cite{mistral2024small31} & 24B & -0.11 & 0.35 & 53.94 & 75.17 & 32.20 \\
    Qwen2.5-Omni~\cite{xu2025qwen2} & 7B & 0.11 & 0.65 & 72.39 & 49.74 & 25.44 \\
    \midrule
    \multicolumn{7}{c}{\textit{Closed-source}} \\
    QvQ-Max~\cite{qwen2025QvQ72b} & - & 0.07 & 0.63 & 78.18 & 63.39 & 47.98 \\
    GPT-4o~\cite{hurst2024GPT} & - & \cellcolor{customSecond!45}-0.01 & \cellcolor{customSecond!45}0.48 & 67.10 & 69.49 & 40.98 \\
    GPT-4.1~\cite{openai2024GPT4} & - & 0.05 & 0.58 & 74.58 & 64.71 & 44.47 \\
    Gemini-2.5-Flash~\cite{gemini2025update} & - & 0.06 & 0.61 & 78.55 & 66.80 & 50.56 \\
    Gemini-2.5-Pro~\cite{gemini2025update} & - & 0.07 & 0.64 & 81.31 & 67.01 & \cellcolor{customSecond!45}51.58 \\
    \bottomrule
  \end{tabular*}
\end{table}

\section{More Examples of EmotionHallucer}\label{app:examples}

\begin{figure}
    \centering
    \includegraphics[width=\linewidth]{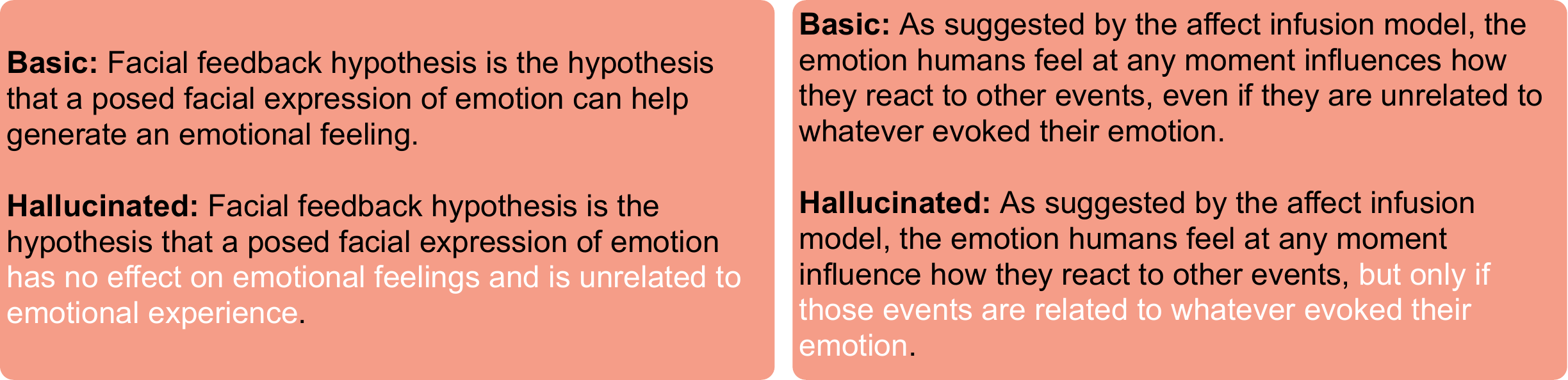}
    \caption{Two examples Basic-Hallucinated pair of Theory Hallucination.}
    \label{fig:more_example_theory}
\end{figure}

\begin{figure}
    \centering
    \includegraphics[width=\linewidth]{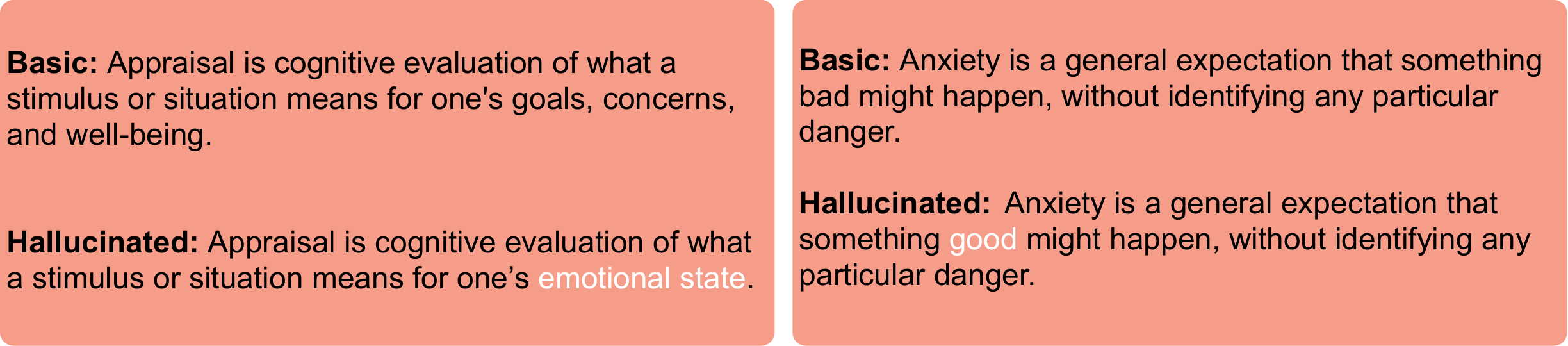}
    \caption{Two examples Basic-Hallucinated pair of Definition Hallucination.}
    \label{fig:more_example_definition}
\end{figure}

\begin{figure}
    \centering
    \includegraphics[width=\linewidth]{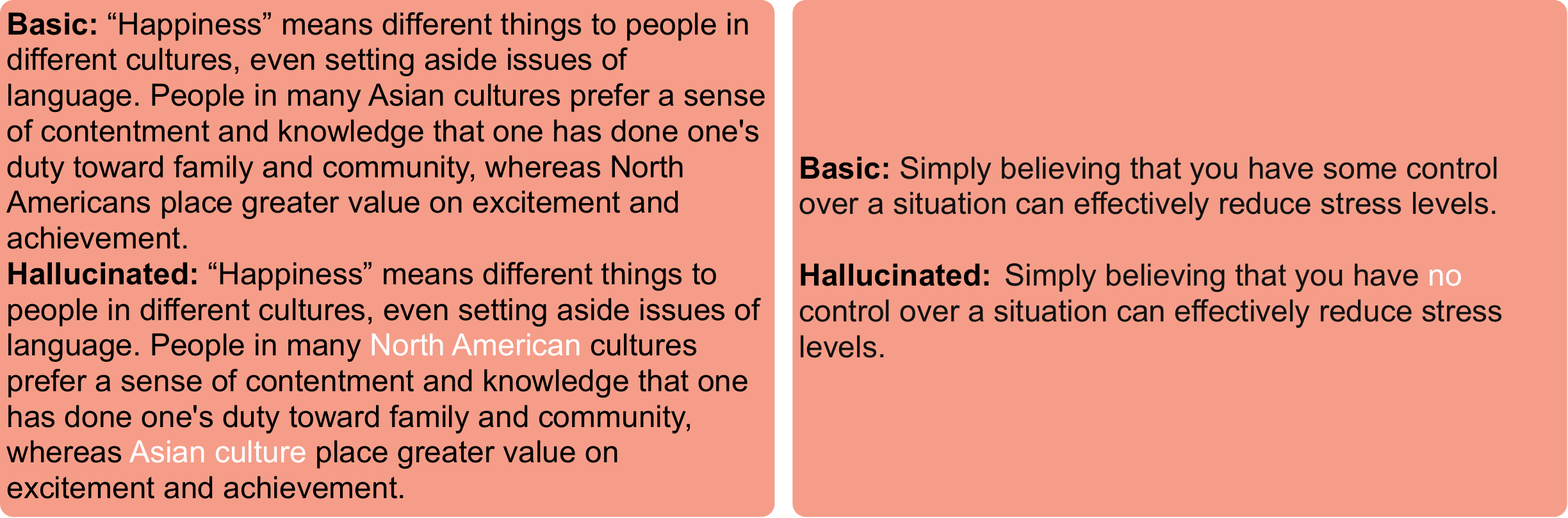}
    \caption{Two examples Basic-Hallucinated pair of Definition Hallucination.}
    \label{fig:more_example_finding}
\end{figure}

\begin{figure}
    \centering
    \includegraphics[width=\linewidth]{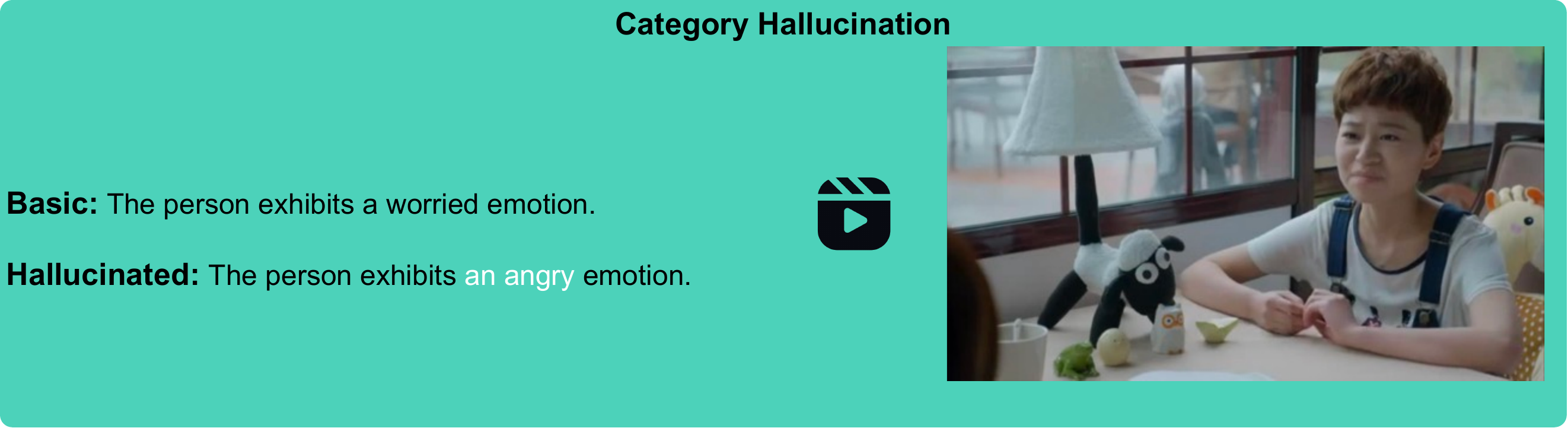}
    \caption{An example Basic-Hallucinated pair of Category Hallucination.}
    \label{fig:more_example_category}
\end{figure}

\begin{figure}
    \centering
    \includegraphics[width=\linewidth]{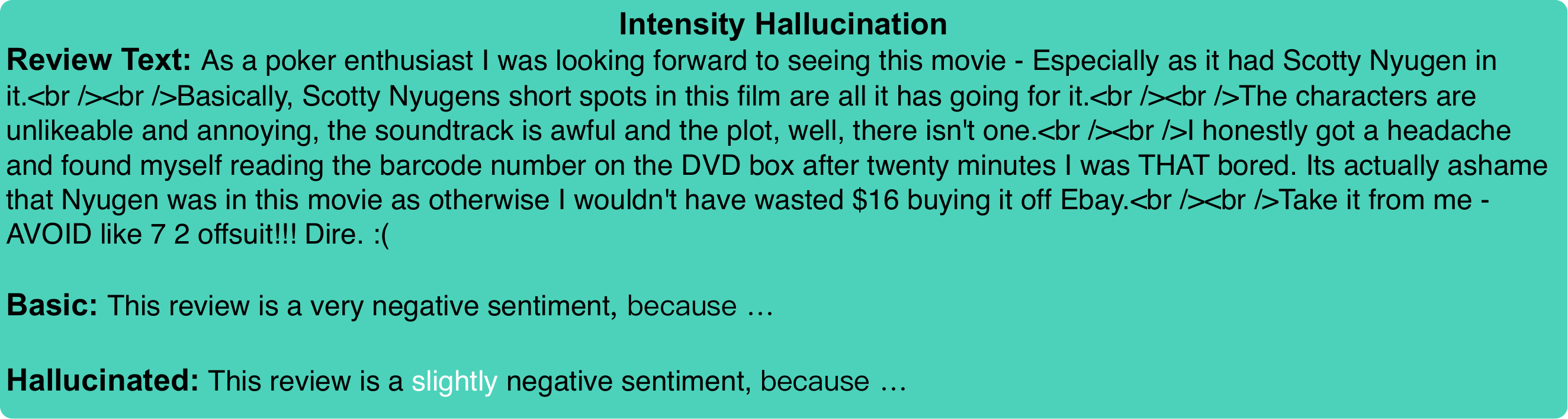}
    \caption{An example Basic-Hallucinated pair of Intensity Hallucination.}
    \label{fig:more_example_intensity}
\end{figure}

\begin{figure}
    \centering
    \includegraphics[width=\linewidth]{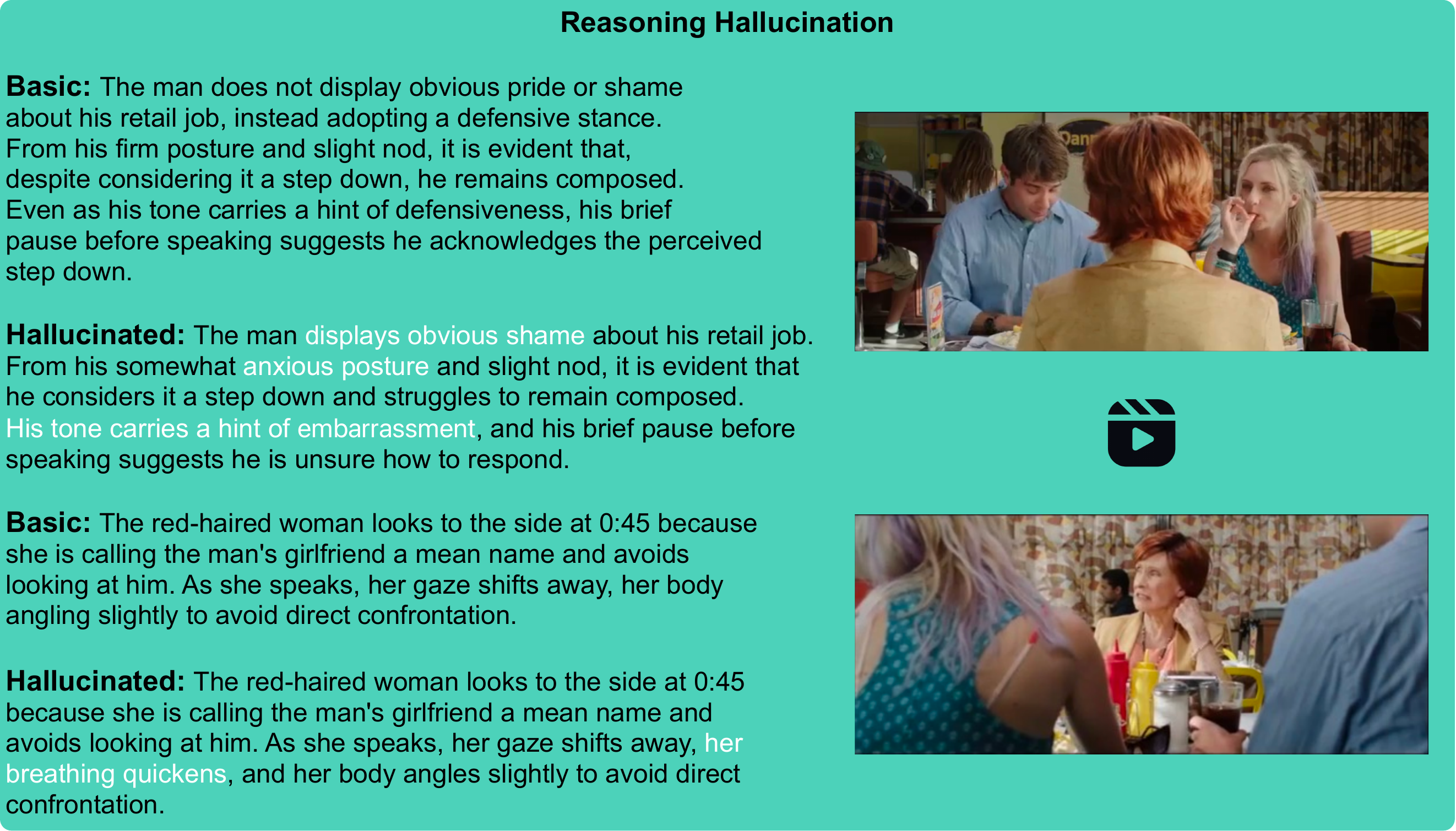}
    \caption{An example Basic-Hallucinated pair of Reasoning Result and Cue Hallucination.}
    \label{fig:more_example_reasoning}
\end{figure}

In this section, we provide more cases from EmotionHallucer, as shown Figure~\ref{fig:more_example_theory}, Figure~\ref{fig:more_example_definition}, Figure~\ref{fig:more_example_finding}, Figure~\ref{fig:more_example_category}, Figure~\ref{fig:more_example_intensity}, and Figure~\ref{fig:more_example_reasoning}. 

\section{Limitations and Discussions}\label{app:limitation}

While our benchmark provides a comprehensive evaluation of emotional hallucinations across modalities, several limitations remain. 
(1) Although we take a lot of strategies to make sure the quality of EmotionHallucer, there is noise introduced by human annotations.
(2) The benchmark currently focuses on English, without accounting for cross-lingual and cultural variations in emotional expression and model behavior. Future work should explore how language and culture influence hallucination patterns in MLLMs. 
(3) While we observe emotion hallucination phenomena, the underlying causes, such as pretraining biases, modality misalignment, or lack of emotion-specific supervision, remain underexplored. Understanding these root causes is essential for designing more robust and interpretable models. 
(4) Although we treat emotion understanding and hallucination detection separately, real-world applications often require joint emotion understanding and hallucination awareness, which calls for unified modeling strategies that integrate both capabilities.
Moreover, building a truly multimodal, emotion MLLM remains a major open challenge. 
Current MLLMs often rely on loosely coupled modality fusion and struggle with temporal and contextual integration, particularly in long-form audio and video scenarios. 

\section{Ethical Statement}

This work complies with ethical standards for AI research. 
All datasets used in this study are publicly available and were originally released under appropriate research or academic licenses. 
No private or sensitive personal data were involved. 
Our benchmark focuses on evaluating emotional hallucinations in language and multimodal models; however, we acknowledge the subjective and culturally nuanced nature of emotion, and we caution against overinterpreting model outputs in high-stakes or sensitive applications. 
Furthermore, hallucinations in emotional reasoning may lead to miscommunication or emotional misjudgment, particularly in domains such as mental health, education, or human-computer interaction. We encourage future work to incorporate robust safety checks, human oversight, and culturally inclusive evaluations when deploying such models in real-world scenarios.

\end{document}